\documentclass{article}
\usepackage[english]{babel}
\usepackage[utf8]{inputenc}
\usepackage{johd}
\usepackage{subfigure}
\usepackage{booktabs}
\usepackage{arydshln}
\usepackage{amsmath}
\usepackage{amssymb}
\newcommand{\tabincell}[2]{\begin{tabular}{@{}#1@{}}#2\end{tabular}}

\title{A Survey of Face Recognition}

\author{Xinyi Wang$^{1}$, Jianteng Peng$^{1}$, Sufang Zhang$^{1}$, Bihui Chen$^{1}$, Yi Wang, Yandong Guo$^{1}$\\
        \small $^{1}$OPPO Research Institute, Beijing, China\\

}

\date{} 

\begin{document}

\maketitle

\begin{abstract} 
\noindent Recent years witnessed the breakthrough of face recognition with deep convolutional neural networks. Dozens of papers in the field of FR are published every year. Some of them were applied in the industrial community and played an important role in human life such as device unlock, mobile payment, and so on. This paper provides an introduction to face recognition, including its history, pipeline, algorithms based on conventional manually designed features or deep learning, mainstream training, evaluation datasets, and related applications. We have analyzed and compared state-of-the-art works as many as possible, and also carefully designed a set of experiments to find the effect of backbone size and data distribution. \textcolor{blue}{This survey is a material of the tutorial named \textit{The Practical Face Recognition Technology in the Industrial World} in the FG2023.} \end{abstract}

\noindent\\

\section{Introduction}\label{sec1}
As one of the most important applications in the field of artificial intelligence and computer vision, face recognition (FR) has attracted the wide attention of researchers. Almost every year, major CV conferences or journals, including CVPR, ICCV, PAMI, etc. publish dozens of papers in the field of FR.

In this work, we give an introduction to face recognition. Chapter 2 is about the history of face recognition. Chapter 3 introduces the pipeline in the deep learning framework. Chapter 4 provides the details of face recognition algorithms including loss functions, embedding techniques, face recognition with massive IDs, cross-domain, pipeline acceleration, closed-set training, mask face recognition, and privacy-preserving. In Chapter 5, we carefully designed a set of experiments to find the effect of backbone size and data distribution. Chapter 6 includes the frequently-used training and test datasets and comparison results. Chapter 7 shows the applications. Chapter 8 introduces the competitions and open-source programs.

\section{History}\label{sec2}

Recent years witnessed the breakthrough of face recognition (FR) with deep Convolutional Neural Networks. But before 2014, FR was processed in non-deep learning ways.
In this section, we introduce conventional FR algorithms with manually designed features. 

Eigenface \cite{turk1991eigenfaces} is the first method to solve the problem of FR in computer vision. 
Eigenface reshaped a grayscale face image into a 1-D vector as its feature. 
Then, a low-dimensional subspace can be found by Principal Component Analysis (PCA), where the distributions of features from similar faces were closer. This subspace can be used to represent all face samples.
For a query face image with an unknown ID, Eigenface adopted the Euclidean distance to measure the differences between its feature and all gallery face features with known categories in the new subspace.
The category of query image can be predicted by finding the smallest distance to all features in the gallery.
Fisherface \cite{belhumeur1997eigenfaces} pointed out that PCA used in Eigenface for dimensionality reduction maximizes the variance of all samples in the new subspace. 
Eigenface did not take sample category into account. 
Therefore, Fisherface considered face labels in the training set and used Linear Discriminant Analysis (LDA) for dimensionality reduction, which maximized the ratios of inter-class and intra-class variance.

As a commonly used mathematical classification model, SVM was brought into the field of FR. 
Since SVM and face feature extractors can be de-coupled, many SVM-based FR algorithms with different extractors have been proposed. 
Deniz \emph{et al.} \cite{deniz2003face} used ICA\cite{oja2000independent} for feature extraction, and then used SVM to predict face ID. 
In order to speed up the training of ICA+SVM, 
Kong \emph{et al.} \cite{kong2011new} designed fast Least Squares SVM to accelerate the training process by modifying the Least Squares SVM algorithm.
Jianhong \emph{et al.} \cite{jianhong2008kpca} combined kernel PCA and Least Squares SVM to get a better result.

As one of the most successful image local texture extractors, Local Binary Pattern (LBP) \cite{ojala1996comparative} obtained characteristics of images through the statistical histogram. 
Many scholars have proposed FR algorithms based on the LBP feature extractor and its variants.
Ahonen \emph{et al.} \cite{ahonen2004face} adopted LBP to extract features of all regions from one face image, and then concatenated those feature vectors for histogram statistics to obtain an embedding of the whole image. 
Finally, they used Chi square as the distance metric to measure the similarity of faces. 
Wolf \emph{et al.} \cite{wolf2008descriptor} proposed an improved region descriptor based on the LBP, and got a better result with higher accuracy.
Tan \emph{et al.} \cite{tan2010enhanced} refined the LBP and proposed the Local Ternary Patterns (LTP), which had a higher tolerance of image noise.

\section{Pipelines in deep learning framework}\label{sec3}

\begin{figure}[htp]
    \centering
    \subfigure[Pipeline of training]{
    \includegraphics[width=10.2cm]{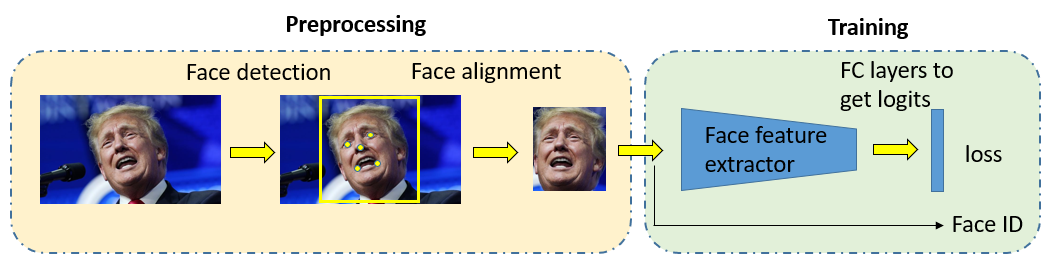}
    }
    \subfigure[Pipeline of inference]{
    \includegraphics[width=14.4cm]{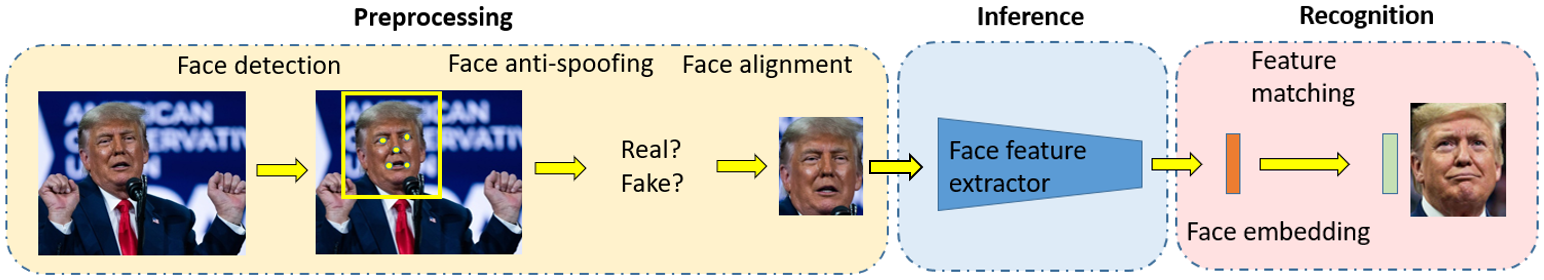}
    }
    \caption{Pipelines of training and inference in face recognition}
    \label{fig:pipe}
\end{figure}

As the computing resources were increasing, mainstream face recognition methods applied deep learning to model this problem, which replaced the aforementioned manually designed feature extraction methods. In this section, we give the pipelines of 'training' a FR deep model and 'inference' to get the ID of a face image (as shown in Fig. \ref{fig:pipe}). And we will elaborate some representative methods related to them.

In general, the pipeline of training FR model contains two steps: face images preprocessing (preprocessing) and model training (training). Getting the FR result by the trained model contains three parts: face images preprocessing (preprocessing), model inference to get face embedding (inference), and recognition by matching features between test image and images with known labels (recognition). In industry, face anti-spoofing will be added before inference in the testing pipeline since FR systems are vulnerable to presenatation attacks ranging from print, video replay, 3D mask, etc.
It needs to be mentioned that, both preprocessing steps in training and inference should be the same.

\subsection{Preprocessing}\label{subsec3.1}

In practice, faces for FR are located in a complicated image or a video. Therefore we need to apply face detection (with face landmark detection) to cut out the specific face patch first, and use face alignment to transform this face patch in a good angle or position. After these two preprocessing, a face image patch is ready for further FR procedures.
It is noticed that, face detection is necessary for FR preprocessing, while face alignment is not.

\subsubsection{Face detection}\label{subsubsec3.1.1}
Face detection (or automatic face localisation) is a long-standing problem in CV. 
Considering human face as a object, many object detection algorithms can be adopted to get a great face detection result, such as Faster-RCNN \cite{ren2015faster}, SSD \cite{liu2016ssd}, YOLO (with different versions \cite{redmon2016you,redmon2017yolo9000,redmon2018yolov3}).

However, some researchers treated human face as a special object and designed variable deep architectures for finding faces in the wild with higher accuracy.
Mtcnn \cite{zhang2016joint} is one of the most famous face detection methods in academy. Mtcnn adopted a cascaded structure with three stages of deep convolutional networks (P-net, R-net, O-net) that predict both face and landmark location in a coarse-to-fine manner.
Non-maximum suppression (NMS) is employed to merge highly overlapped candidates, which are produced by P-net, R-net and O-net.
Faceness-Net \cite{yang2017faceness} observed that facial attributes based supervision can effectively enhance the capability of a face detection network in handling severe occlusions.
As a result, Faceness-Net proposed a two-stage network, where the first stage applies several branches to generate response maps of different facial parts and the second stage refines candidate window using a multi-task CNN. At last, face attribute prediction and bounding box regression are jointly optimized.

\begin{figure}[htp]
    \centering
    \includegraphics[width=15cm]{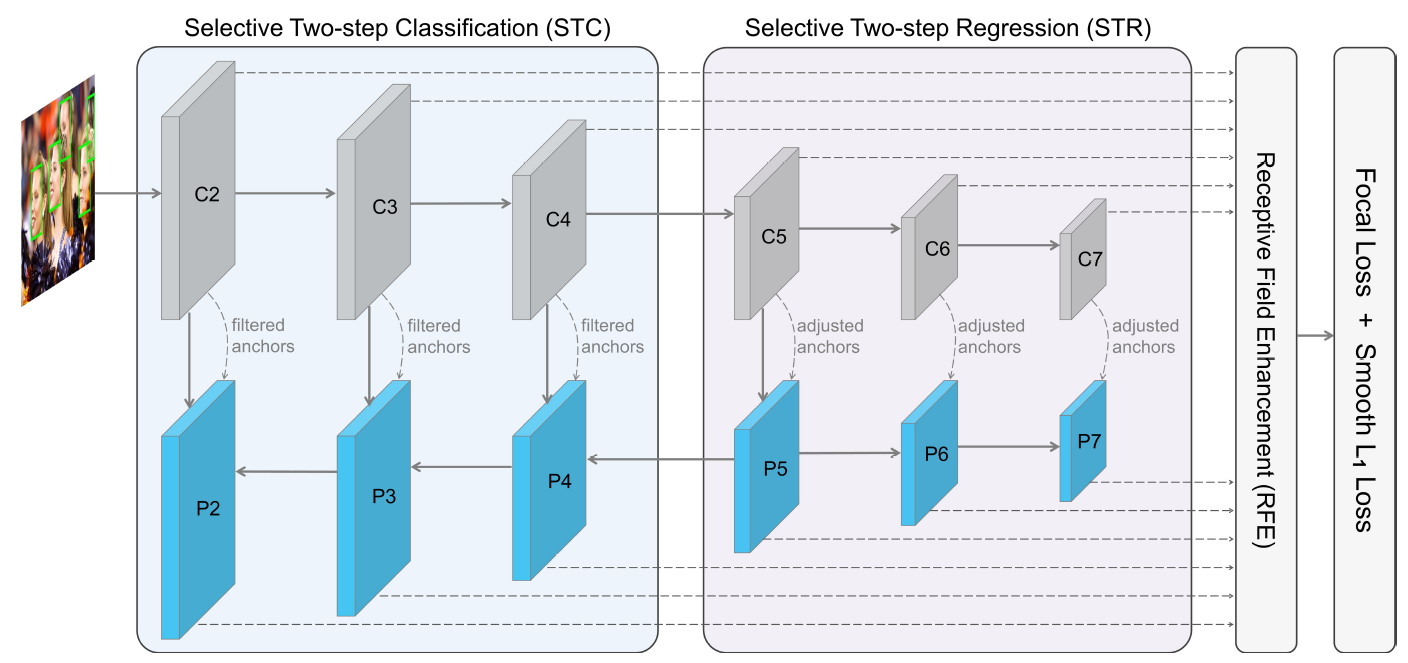}
    \caption{The Pipeline of SRN}
    \label{fig:srn}
\end{figure}

As FPN (Feature Pyramid Network) is widely used in object detection, many researchers tried to bring in FPN in face detection and obtained success in detecting tiny faces. 
SRN (Selective Refinement Network) \cite{chi2019selective} adopted FPN to extract face features, and introduced two-step classification and regression operations selectively into an anchor-based face detector to reduce false positives and improve location accuracy simultaneously. 
In particular, the SRN consists of two modules: the Selective Two-step Classification (STC) module and the Selective Two-step Regression (STR) module. 
The STC aims to filter out most simple negative anchors from low level detection layers to reduce the search space for the subsequent classifier, 
while the STR is designed to coarsely adjust the locations and sizes of anchors from high level detection layers to provide better initialization for the subsequent regressor.
The pipeline of SRN can be seen in Fig. \ref{fig:srn}

\begin{figure}[htp]
    \centering
    \includegraphics[width=16cm]{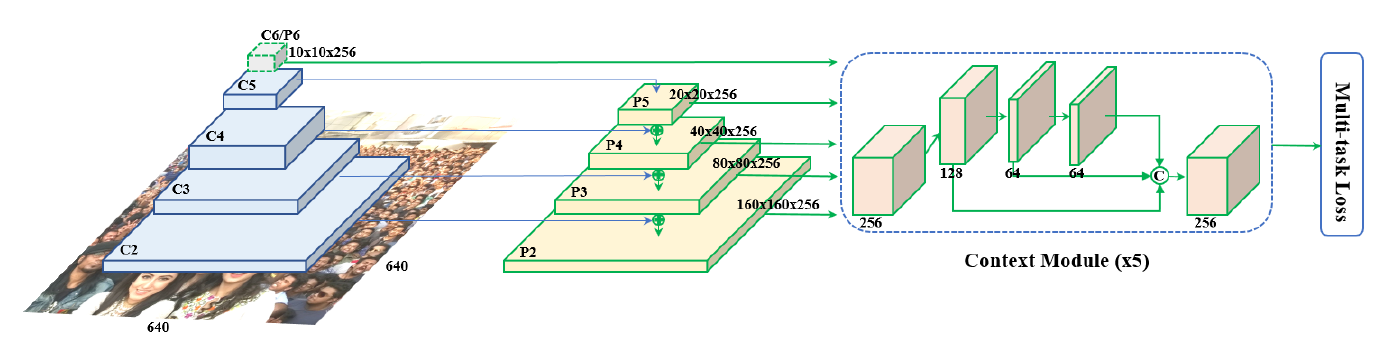}
    \caption{The Pipeline of RetinaFace}
    \label{fig:retinaface}
\end{figure}

RetinaFace \cite{deng2019retinaface} is another face detector using FPN to extract multi-level image features. However, different from two-step detecting methods, such as SRN, RetinaFace presented a single-stage face detector.
Further more, RetinaFace performed face detection in a multi-task way.
In specific, RetinaFace predicted following 4 aspects at the same time: (1) a face score, (2) a face box, (3) five facial landmarks, and (4) dense 3D face vertices projected on the image plane.
The pipeline of RetinaFace can be seen in Fig. \ref{fig:retinaface}

CRFace \cite{vesdapunt2021crface} proposed a confidence ranker to refine face detection result in high resolution images or videos.
HLA-Face \cite{wang2021hla} solved face detection problem in low light scenarios.
As the development of general 2D object detection, face detection has reached a high performance, even in real-world scenarios.
As a result, innovative papers in face detection seldom show up in recent years.

\subsubsection{Face anti-spoofing}\label{subsubsec3.1.2}
In industry, faces fed into recognition system need to be check whether they are real or not, in case someone uses printed images, videos or 3D mask to pass through the system, especially in secure scenarios.
As a result, while inference, face anti-spoofing (or presentation attack detection) is an important prepossessing step, which is located after face detection.

\begin{figure}[htp]
    \centering
    \includegraphics[width=15cm]{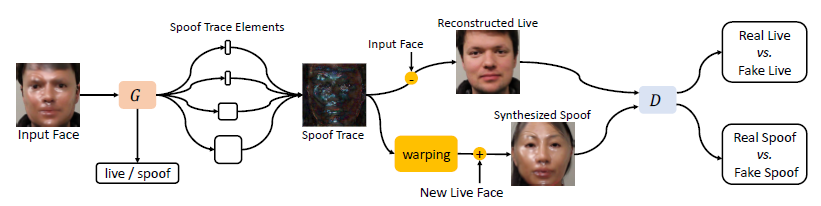}
    \caption{The Pipeline of STDN}
    \label{fig:stdn}
\end{figure}

The major methods for face anti-spoofing usually input signals from RGB cameras. 
Both hand-crafted feature based (e.g., LBP \cite{boulkenafet2015face}, SIFT \cite{patel2016secure}, SURF \cite{boulkenafet2016face} and HoG \cite{komulainen2013context}) and deep learning based methods have been proposed. CDCN \cite{yu2020searching} designed a novel convolutional operator called Central Difference Convolution to get fine-grained invariant information in diverse diverse enviroments. The output feature map $y$ can be formulated as:
\begin{equation}
y(p_0) = \sum_{p_n \in R} w(p_n) \cdot x(p_0 + p_n)
\label{eq:cdcn}
\end{equation}
where $p_0$ is the current location on both input and output feature map, and $p_n$ is the location in $R$. In addition, CDCN used NAS to search backbone for face anti-spoofing task, which is supervised by depth loss instead of binary loss. DCDN \cite{yu2021dual} decoupled the central gradient features into two cross directions (horizontal/vertical or diagonal) and achieved better performance with less computation cost when compared with CDCN. STDN \cite{liu2020disentangling} observed that there is little explanation of the classifier's decision. As a result, STDN proposed an adversarial learning network to extract the patterns differentiating a spoof and live face. The pipeline of STDN can be seen in Fig \ref{fig:stdn}. PatchNet \cite{wang2022patchnet} rephrased face anti-spoofing as a fine-grained material recognition problem. Specifically, PatchNet splitted the categories finely based on the capturing devices and presenting materials, and used patch-level inputs to learn discriminative features. In addition, the asymmetric margin-based classification loss and self-supervised similarity loss were proposed to further improve the generalization ability of the spoof feature. SSAN \cite{wang2022domain} used a two-stream structure to extract content and style features, and reassembled various content and style features to get a stylized feature space, which was used to distinguish a spoof and live face. In specific, adversarial learning was used to get a shared feature distribution for content information and a contrastive learning strategy was proposed to enhance liveness-related style information while suppress domain-specific one.

Merely using RGB signals which belong to visible spectrum has a lot of limitations in face anti-spoofing, therefore an increasing numbers of methods tried to adopt multi-channel signals to achieve higher accuracy. FaceBagNet \cite{shen2019facebagnet} used patch-level image to extract the spoof-specific discriminative information and proposed a multi-modal features fusion strategy. CMFL \cite{george2021cross} designed a cross-modal focal loss to modulate the individual channels' loss contributions, thus captured complementary information among modalities.

\subsubsection{Face alignment}\label{subsubsec3.1.3}
Face image patches from face detection are often different in shapes due to factors such as pose,
perspective transformation and so on, which will lead to a decline on recognition performance.
Under this condition, face alignment is an effective approach to alleviate this issue by transforming face patches into a similar angle and position \cite{parkhi2015deep,schroff2015facenet}.
Face image set with an identical ID will get a smaller intra-class difference after face alignment, further making the training classifier more discriminative. 

Common way to align face images is using a 2D transformation to calibrate facial landmarks to predefined frontal templates or mean face mode. This 2D transformation is normally processed by affine transformation \cite{parkhi2015deep}.
Deepface \cite{taigman2014deepface} proposed a FR pipeline which employed a 3D alignment method in order to align faces undergoing out-of-plane rotations.
Deepface used a generic 3D shape model and registered a 3D affine camera, which are used to warp the 2D aligned crop to the image plane of the 3D shape.
STN (Spatial Transform Networks) \cite{jaderberg2015spatial} introduced a learnable module which explicitly allowed the spatial manipulation of image within the network, and it can also be utilised in face alignment.
Wu \emph{et al.} \cite{wu2017recursive} merged STN based face alignment and face feature extractor together, and put forward Recursive Spatial Transformer (ReST) for end-to-end FR training with alignment. 
2D face alignment is faster than both 3D alignment and STN, and thus it is widely used in preprocessing step of FR.
APA \cite{an2019apa} proposed a more general 2D face alignment method. 
Instead of aligning all faces to near-frontal shape, 
APA adaptively learned multiple pose-specific templates, which preserved the face appearance with less artifact and information loss. 

As the mainstream FR training sets are becoming larger gradually, some FR methods choose to omit face alignment step and train (or test) with face patches directly from face detection. Technically, face alignment is a way to increase intra-class compactness. And a well-trained FR model with more IDs can also obtain a compact feature distribution for each class. Therefore, face alignment may not be necessary nowadays.

\subsection{Training and testing a FR deep model}\label{subsec3.2}
After preprocessing, face images with their ground truth IDs can be used to train a FR deep model. 
As the progressing of computational hardware, any mainstream backbone is able to build a FR network.
Different backbones with similar parameter amount have similar accuracy on FR.
As a result, you can use Resnet, ResNext, SEResnet, Inception net, Densenet, etc. to form your FR system.
If you need to design a FR system with limited calculation resource, Mobilenet with its variation will be good options.
In addition, NAS can be used to search a better network hyper-parameter. 
Network distilling technology is also welcomed while building industrial FR system.

The training algorithms will be elaborated in the following sections. For testing, a face image after preprocessing can be inputted into a trained FR model to get its face embedding. Here, we list some practical tricks which will be utilized in the model training and testing steps.
Firstly, training image augmentation is useful, especially for IDs with inadequate samples. Common augmentation ways, such as adding noise, blurring, modifying colors, are usually employed in FR training. However, randomly cropping should be avoided, since it will ruin face alignment results.
Secondly, flipping face images horizontally is a basic operation in both of training and testing FR model. In training, flipping faces can be consider as a data augmentation process. In testing, we can put image $I$ and its flipped mirror image $I'$ into the model, and get their embeddings $f$ and $f'$. Then their mean feature $\frac{f+f'}{2}$ can be used as the embedding of image $I$. This trick will further improve the accuracy of FR results.  

\subsection{Face recognition by comparing face embeddings}\label{subsec3.3}
The last part of face inference pipeline is face recognition by comparing face embeddings. According to the application, it can be divided into face verification and face identification.
To apply FR, a face gallery needs to be built. First we have a face ID set $S$, where each ID contains one (or several) face image(s). All face embeddings from the gallery images will be extracted by a trained model and saved in a database.

The protocol of face verification (FV) is: giving a face image $I$ and a specific ID in the gallery, output a judgement whether the face $I$ belongs to the ID. So FV is a 1:1 problem. We extract the feature of image $I$ and calculate its similarity $s(I,ID)$ with the ID's embedding in the database. $s(I,ID)$ is usually measured by cosine similarity. If $s(I,ID)$ is larger than a threshold $\mu$, we will output `True' which represents image $I$ belongs to the ID, and vice verse.

However, face identification (FI) is a 1:N problem. Giving a face image $I$ and a face ID set $S$, it outputs the ID related to the face image, or `not recognized'.
Similarly, we extract the feature of $I$ and calculate its similarity $s(I,ID), ID \in S$ with all IDs' embeddings in the database. 
Then, we find the maximum of all similarity values. If $\max( s(I,ID))$ is larger than a threshold $\mu$, we output its related ID $\arg_{ID} \max( s(I,ID))$ as identification target; otherwise, we output `not recognized', which shows that the person of image $I$ is not in the database.
Phan \emph{et al.} \cite{phan2022deepface} enriched the FI process by adopting an extra Earth Mover’s Distance. 
They first used cosine similarity to obtain a part of the most similar faces of the query image. Then a patch-wise Earth Mover’s Distance was employed to re-rank these similarities to get final identification results.
In the subsection \ref{subsec4.5}, we will introduce some methods to accelerate the process of face identification.

\section{Algorithms}\label{sec4}

In this section, we will introduce FR algorithms in recent years. Based on different aspects in deep FR modeling, we divided all FR methods into several categories: designing loss function, refining embedding, FR with massive IDs, FR on uncommon images, FR pipeline acceleration, and close-set training.

\subsection{Loss Function}\label{subsec4.1}

\subsubsection{Loss based on metric learning}\label{subsubsec4.1.1}
 Except from softmax based classification, FR can be also regarded as extracting face features and performing feature matching. Therefore, training FR model is a process to learn a compact euclidean feature space, where distance directly correspond to a measure of face similarity. This is the basic modeling of metric learning.
In this section, we introduce some representative FR methods based on metric learning.

Inspired by the pipeline of face verification, one intuitive loss designing is judging whether two faces in a pair have identical ID or not. 
Han \emph{et al.} \cite{han2018face} used this designing in a cross-entropy way, and the loss is:
\begin{equation}
L_{i,j}= -[ y_{ij} \log p_{ij} + (1-y_{ij}) \log (1-p_{ij}) ]
\end{equation}
where $y_{ij}$ is the binary GT of whether the two compared face images $i$ and $j$ belong to the same identity. $p_{ij}$ is the logits value.

Different from \cite{han2018face}, contrastive loss \cite{sun2015deep} was proposed to directly compare the features of two face images in a metric learning way.
If two images belong to a same ID, their features in the trained space should be closer to each other, and vice verses. 
As a result, the contrastive loss is:
\begin{equation}
L_{i,j}=
\left
\{\begin{array}{ll}
\frac{1}{2} \|f_i - f_j\|^2_2,              & if  \ y_{ij}=1 \\
\frac{1}{2} max(0, m-\|f_i - f_j\|^2_2),   & if  \ y_{ij}=-1
\end{array}\right.
\end{equation}
where $i$, $j$ are two samples of a training pair, and $f_i$ and $f_j$ are their features. $\|f_i - f_j\|^2_2$ is their Euclidean distance. $m$ is a margin for enlarging the distance of sample pairs with different IDs (negative pairs). 
Further more, Euclidean distance can be replaced by cosine distance, and contrastive loss become the following form:
\begin{equation}
L_{i,j} = \frac{1}{2} (y_{ij} - \sigma (wd+b))^2  
    \ \ , \ \  
d = \frac{f_if_j}{\|f_i\|_2\|f_j\|_2}
\end{equation}
where $d$ is the cosine similarity between feature $f_i$ and $f_j$, $w$ and $b$ are learnable scaling and shifting parameters, $\sigma$ is the sigmoid function.

Different from contrastive loss, BioMetricNet \cite{ali2020biometricnet} did not impose any specific metric on facial features. Instead, it shaped the decision space by learning a latent representation in which matching (positive) and non-matching (negative) pairs are mapped onto clearly separated and well-behaved target distributions.
BioMetricNet first extracted face features of matching and non-matching pairs, and mapped them into a new space in which a decision is made. Its loss used to measure the statistics distribution of both matching and non-matching pairs in a complicated form. 

FaceNet \cite{schroff2015facenet} proposed the triplet loss to minimize the distance between an anchor and a positive, both of which have the same identity, and maximize the distance between the anchor and a negative of a different identity.
Triplet loss was motivated in \cite{weinberger2009distance} in the context of nearest-neighbor classification.
The insight of triplet loss, is to ensure that an image $x^a$ (anchor) of a specific person is closer to all other images $x^p$ (positive) of the same person than it is to any image $x^n$ (negative) of any other person.
Thus the loss is designed as:
\begin{equation}
L = \sum_{(x^a,x^p,x^n) \in T} \max( \| f(x^a) - f(x^p)\|_2^2 - \| f(x^a) - f(x^n) \|_2^2 + \alpha , 0)
\label{eq:triplet}
\end{equation}
where $T$ is is the set of all possible triplets in the training set. $f(x)$ is the embedding of face $x$. $\alpha$ is a margin that is enforced between positive and negative pairs.
In order to ensure fast convergence, it is crucial to select triplets that violate the triplet constraint of:
\begin{equation}
\| f(x^a) - f(x^p)\|_2^2 + \alpha < \| f(x^a) - f(x^n) \|_2^2
\end{equation}
This means that, given $x^a$ , we want to select an $x^p$ (hard positive) such that $\arg\max_{x^p} \| f(x^a) - f(x^p)\|_2^2$, and similarly $x^n$ (hard negative) such that $\arg\min_{x^n} \| f(x^a) - f(x^n)\|_2^2$.
As a result, the accuracy of the triplet loss model is highly sensitive to the training triplet sample selection.

Kang \emph{et al.} \cite{kang2018pairwise} simplified both of contrastive and triplet loss and designed a new loss as:
\begin{equation}
\begin{aligned}
L   &= \lambda_1 L_t + \lambda_2 L_p + \lambda_3 L_{softmax}    \\
L_t &= \sum_{(x^a,x^p,x^n) \in T} 
      \max(0,1-\frac{\|f(x^a)-f(x_n)\|_2}{\|f(x^a)-f(x_p)\|_2+m})        \\
L_p &= \sum_{(x^a,x^p,.) \in T} \| f(x^a) - f(x^p)\|_2^2
\end{aligned}
\end{equation}
where $L_t$, $L_p$, $L_{softmax}$ are modified triplet ratio loss, pairwise loss and regular softmax loss, respectively. Factor $m$ in $L_t$ is a margin to enlarge the difference between positive pair and negative pair, which has similar use with $\alpha$ in (\ref{eq:triplet}).

Contrastive loss measures the difference of a image pair; triplet loss represented the relations in a triplet: anchor, positive and negative samples.
Some researchers have developed more metric learning based FR losses by describing more samples.

In order to enhance the discrimination of the deeply learned features, the center loss \cite{wen2016discriminative} simultaneously learned a center for deep features of each class and 
penalized the distances between the deep features and their corresponding class centers. 
Center loss can be added on softmax or other FR loss to refine feature distribution while training. 
The form of center loss is:
\begin{equation}
L = L_{softmax} + \lambda  L_c  
    \ \ , \ \   
L_c(i) = \frac{1}{2} \sum_{i=1}^m \|x_i - c_{y_i}\|^2_2
\end{equation}
where $L_{softmax}$ is softmax loss on labelled training data. $i$ is a image sample with its face label $y_i$, and $x_i$ is its deep feature. $m$ is the number of training classes.
$c_{y_i}$ denotes the $y_i$-th class center of deep features. The formulation
of center loss effectively characterizes the intra-class variations.

\subsubsection{Larger margin loss}\label{subsubsec4.1.2}
Face recognition is naturally treated as a classification problem, and thus using softmax as the loss to train a FR model is an intuitive consideration. 
Larger margin loss (also is called as angular margin based loss) is derived from softmax.
Larger margin loss is a major direction in FR, since it has largely improved the performance of FR in academy and industry. 
The intuition is that facial images with same identities are expected to be closer in the representation space, while different identities expected to be far apart.
As a result, larger margin losses encourage the intra-class compactness and penalize the similarity  of different identities.


First, we give the formulation of softmax as follows:
\begin{equation}
L_i = -\log\frac
{e^{{W_{y_i} \cdot x_i + b_{y_i}}}}
{\sum_{j} e^ { {W_{j} \cdot x_j + b_j}}}
\end{equation}
where ${W}$ is the weighting vector of the last fully connected layer, and ${W_{j}}$ represents the weight of class $j$ ; $x_{i}$ and $y_{i}$ are the face embedding of sample $i$ and its ground truth ID.
For class $y_i$, sample $i$ is a positive sample; and for class $j(j \neq y_i)$, sample $i$ is a negative sample. 
These two definitions will be used in this whole paper. 
Considering positive and negative samples, softmax can be rewritten as:
\begin{equation}
L_i = -\log\frac
{e^{{W_{y_i} \cdot x_i + b_{y_i}}}}
{e^{{W_{y_i} \cdot x_i + b_{y_i}}} + \sum_{j \neq y_i} e^ { {W_{j} \cdot x_j + b_j}} }
\end{equation}

L-Softmax \cite{liu2016large} first designed margin based loss by measuring features angles. First, it omits the bias of each class $b_{j}$, and changes the inner product of features and weights $W_j \cdot x_i$ to a new form $\|W_j \| \cdot \|x_i\| \cdot \cos(\theta_j)$, where $\theta_{j}$ is the angle between $x_{i}$ and the weight $W_{j}$ of class j.
In order to enlarge the margin of angels between each class, L-Softmax modifies $\cos(\theta_{y_i})$ to $\psi(\theta_{y_i})$ by narrowing down the space between decision boundary and the class centers. In specific, it has:
\begin{equation}
L_{i}=-\log \frac
{e^{\left\|{W}_{y_{i}}\right\|\left\|{x}_{i}\right\| \psi\left(\theta_{y_{i}}\right)}}
{e^{\left\|{W}_{y_{i}}\right\|\left\|{x}_{i}\right\| \psi\left(\theta_{y_{i}}\right)} +
\sum_{j \ne y_i} e^{\left\|{W}_{j}\right\|\left\|{x}_{i}\right\| \cos \left(\theta_{j}\right)}}
\end{equation}
\begin{equation}
\psi(\theta) = (-1)^k\cos(m\theta) - 2k
\end{equation}
where $m$ is a fix parameter which is an integer; the angle $\theta \in [0,\pi]$ has been divided in $m$ intervals: $[\frac{k\pi}{m}, \frac{(k+1)\pi}{m}]$; $k$ is an integer and $k\in[0, m-1]$. Therefore, it has $\psi(\theta) \le \cos(\theta)$.


Similar with L-Softmax, Sphereface \cite{liu2017sphereface} proposed A-Softmax, which normalized each class weight ${W}_j$ before calculating the loss. Therefore, the loss became:
\begin{equation}
L_{i}=-\log \frac
{e^{\left\|{x}_{i}\right\| \psi\left(\theta_{y_{i}}\right)}}
{e^{\left\|{x}_{i}\right\| \psi\left(\theta_{y_{i}}\right)} +
\sum_{j \ne y_i} e^{\left\|{x}_{i}\right\| \cos \left(\theta_{j}\right)}}
\end{equation}
In the training step of Sphereface, its actual loss is (1-$\alpha$)*softmax + $\alpha$*A-softmax, where $\alpha$ is a parameter in [0,1]. During training, $\alpha$ is changing gradually from 0 to 1. This design is based on two motivations. Firstly, directly training A-Softmax loss will lead to hard convergence, since it brutally pushes features from different ID apart. Secondly, training a softmax loss first will decrease the angle $\theta_{y_i}$ between feature $i$ and its related weight $W_{y_i}$, which causes the part $\cos({m \theta})$ in A-softmax loss in a monotonous area. And it is easier to get a lower loss while gradient descending.

Besides the weight of each class, NormFace \cite{wang2017normface} proposed that face embeddings need to be normalized as well. 
Also, NormFace scales up the normalized face embeddings, which will alleviate the data unbalancing problem of positive and negative samples for better convergence. However, NormFace doesn't use extra margin for positive samples as Sphereface. 
Therefore, the NormFace loss is shown as:
\begin{equation}
L_i = -\log
 \frac
{e^{s {\tilde W_{y_i} \tilde x_i}}}
{\sum_j e^{ {s\tilde W_j \tilde x_j}}}
=-\log \frac
{e^{s\cos(\theta_{y_i})}}
{\sum_j e^{s\cos(\theta_{j})}}
\label{eq:norm}
\end{equation}
where $ {\tilde W}$ and $ {\tilde x}$ are the normalized class weights and face embeddings; $s$ is a scaling up parameter and $s>1$.
To this end, both face embeddings and class weights are distributed on the hypersphere manifold, due to the normalization.

AM-Softmax \cite{wang2018additive} and CosFace \cite{wang2018cosface} enlarged softmax based loss by minus a explicit margin $m$ outside $\cos(\theta)$ as follows:
\begin{equation}
L_i = -\log
 \frac
{e^{s(\cos(\theta_{y_i})-m)}}
{e^{s(\cos(\theta_{y_i})-m)} + \sum_{j \ne y_i} e^{s\cos(\theta_j)}}
\label{eq:CosFace}
\end{equation}

ArcFace \cite{deng2019arcface} putted the angular margin $m$ inside $\cos(\theta)$, and made the margin more explainable. The ArcFace loss is shown as:
\begin{equation}
L_i = -\log \frac{e^{s(cos(\theta_{y_i}+m))}}{e^{s(\cos(\theta_{y_i}+m))} + \sum_{j \ne {y_i}} e^{s\cos(\theta_j)}}
\label{eq:ArcFace}
\end{equation}

The aforementioned angular margin based losses (such as CosFace and ArcFace) includes sensitive hyper-parameters which can make training process unstable. 
P2SGrad \cite{zhang2019p2sgrad} was proposed to address this challenge by directly designing the gradients for training in an adaptive manner.

AM-Softmax \cite{wang2018additive}, CosFace \cite{wang2018cosface} and ArcFace \cite{deng2019arcface} only added angular margins of positive samples. Therefore, their decision boundaries only enforced positive samples getting closer to its centers.
On the contrary, SV-Softmax \cite{wang2018support} got the intra-class compactness by pushing the hard negative samples away from positive class centers, which was chosen by hard example mining. 
In SV-Softmax, a binary label $I_j$ adaptively indicates whether a sample $j$ is hard negative or not, which was defined as:
\begin{equation}
I_{j}=
\left
\{\begin{array}{ll}
0, & \cos \left(\theta_{y_i}\right)-\cos \left(\theta_{j}\right) \geq 0 \\
1, & \cos \left(\theta_{y_i}\right)-\cos \left(\theta_{j}\right)<0
\end{array}\right.
\end{equation}
Therefore, the loss of SV-Softmax is:
\begin{equation}
L_i = -\log \frac{e^{s(\cos(\theta_{y_i}))}}{e^{s(\cos(\theta_{y_i}))} + h(t, \theta_{j}, I_j) \sum_{j \ne y_i} e^{s\cos(\theta_j)}}
\label{eq:sv}
\end{equation}
\begin{equation}
h(t, \theta_j,I_j) = e^{s(t-1)(\cos(\theta_j)+1)I_k}
\end{equation}

One step forward, SV-softmax evolve to SV-X-Softmax by adding large margin loss on positive samples:
\begin{equation}
L_i = -\log \frac
{e^{sf(m, \theta_{y_i})}}
{e^{sf(m, \theta_{y_i})} +  \sum_{j \ne y_i} h(t, \theta_{j}, I_j) e^{s\cos(\theta_j)}}
\label{eq:svx}
\end{equation}
where $f(m, \theta_{y_i})$ can be any type of large margin loss above, such as CosFace or ArcFace.
At the same time, $I_j$ is changed to:
\begin{equation}
I_{j}=
\left
\{\begin{array}{ll}
0, & f(m, \theta_{y_i})-\cos \left(\theta_{j}\right) \geq 0 \\
1, & f(m, \theta_{y_i})-\cos \left(\theta_{j}\right)<0
\end{array}\right.
\end{equation}
SV-Softmax has also evolved to MV-Softmax \cite{wang2020mis} by re-defining $h(t, \theta_j,I_j)$.

In aforementioned methods such as CosFace \cite{wang2018cosface} and ArcFace \cite{deng2019arcface}, features have been normalized in the loss. 
Ring loss \cite{zheng2018ring} explicitly measured the length of features and forced them to a same length $R$, which is shown as:
\begin{equation}
L_R = \frac{\lambda}{2m} \sum_{i=1}^{m} (\| f(x_i)\|_2 - R )^2
\end{equation}
where $f(x_i)$ is the feature of the sample $x_i$. $R$ is the target norm value which is also learned and $m$ is the batch-size.
$\lambda$ is a weight for trade-off between the primary loss function. 
In \cite{zheng2018ring}, the primary loss function was set to softmax and SphereFace \cite{liu2017sphereface}.

Considering a training set with noise, Hu \emph{et al.} \cite{hu2019noise} concluded that the smaller angle $\theta$ between a sample and its related class center, the greater probability that this sample is clean. 
Based on this fact, a paradigm of noisy learning is proposed, which calculates the weights of samples according to $\theta$, where samples with less noise will be distributed higher weights during training.
In a training set with noisy samples, the distribution of $\theta$ normally consists of one or two Gaussian distributions. 
These two Gaussian distributions correspond to noisy and clean samples. 
For the entire distribution, left and right ends of $\theta$ distribution are defined as $\delta_l$ and $\delta_r$. And the peaks of the two Gaussian distributions are $\mu_l$ and $\mu_r$ ($\mu_l=\mu_r$ if the distribution consists of only one Gaussian). 
During training, $\theta$ tends to decreasing, so $\delta_r$ can indicate the progress of the training.
In the initial training stage, the $\theta$ distribution of clean and noisy samples will be mixed together. At this stage, the weights of all samples are set to the same, that is, $w_{1,i}=1$. 
As the training progressing, $\theta$ of the clean sample is gradually smaller than the noisy sample. 
At this time, clean sample should be given a higher weight, that is, $w_{2_i} = \frac{softmax(\lambda z)}{softmax(\lambda)}$, where $z=\frac{\cos\theta-\mu_l}{\delta_r-\mu_l}$, $softmax(x)=\log(1+e^x)$, $\lambda$ is the regularization coefficient. 
At later training stage, the weight of semi-hard samples should be higher than that of easy samples and hard samples, that is, $w_{3,i}=e^{-(\cos\theta-\mu_r)^2/2\sigma^2}$.

In conclusion, the final loss based on AM-Softmax is
\begin{equation}
L_i = -w_i\log\frac
{e^{s(\cos(\theta_{y_i})-m)}}
{e^{s(\cos(\theta_{y_i})-m)} + \sum_{j \ne y_i} e^{s\cos(\theta_j)}}
\end{equation}
\begin{equation}
w_i=\alpha(\delta_r)w_{1,i}+\beta(\delta_r)w_{2,i}+\gamma(\delta_r)w_{3,i}
\end{equation}
where $\alpha(\delta_r)$, $\beta(\delta_r)$ and $\gamma(\delta_r)$ are parameters to reflect the training stage, which are designed empirically.

Similar with \cite{hu2019noise}, the sub-center ArcFace \cite{deng2020sub} also solved the problem of noisy sample learning on the basis of ArcFace. 
The proposed sub-centers encourage one dominant sub-class that contains the majority of clean faces and non-dominant sub-classes that include hard or noisy faces aggregate, and thus they relax the intra-class constraint of ArcFace to improve the robustness to label noise.

In this method, Deng \emph{et al.} modified the fully connected layer classifier with dimension $d\times n$ to a sub-classes classifier with dimension $d\times n\times K$, 
where $d$ is the dimension of embedding, $n$ is the number of identities in the training set, and $K$ is the number of sub-centers for each identity. 
The sub-center ArcFace loss is shown as
\begin{equation}
L_i = -\log \frac{e^{s(\cos(\theta_{i,y_i}+m))}}{e^{s(\cos(\theta_{i,y_i}+m))} + \sum_{j \ne {y_i}} e^{s\cos(\theta_{i,j})}}
\label{eq:subcenterArcFace}
\end{equation}
\begin{equation}
\theta_{i,j} = \arccos \max_k ( W_{j_k}^T  x_i), k={1, \cdots, K}
\end{equation}
where $\theta_{i,j}$ is the smallest angular between embedding $x_i$ and all sub-class weights in class $j$.

CurricularFace \cite{huang2020curricularface} adopted the idea of curriculum learning and weighted the class score of negative samples. So in the earlier stage of training CurricularFace, the loss fits easy samples. And at the later training stage, the loss fits hard samples. In specific, CurricularFace loss is :
\begin{equation}
L_i = -\log \frac{e^{sf(m, \theta_{y_i})}}{e^{sf(m, \theta_{y_i})} +  \sum_{j \ne y_i}  e^{s N(t, m,\theta_j, \theta_{y_i})}}
\label{eq:curri}
\end{equation}
and 
\begin{equation}
N(t, m,\theta_j, \theta_{y_i})=\left\{\begin{array}{ll}\cos\theta_j,                 & \cos(\theta_{y_i}+m)-\cos \left(\theta_{j}\right) \geq 0 \\\cos\theta_j(t+\cos\theta_j), & \cos(\theta_{y_i}+m)-\cos \left(\theta_{j}\right)<0\end{array}\right.
\end{equation}

Different from SV-Softmax, $t$ in $N(t, \theta_j)$ from CurricularFace is dynamically updated by Exponential Moving Average (EMA), which is shown as follows:
\begin{equation}
t^{(k)} = \alpha r^{(k)} + (1-\alpha)r^{(k-1)}
\end{equation}
where $t_0 = 0$, $\alpha$ is the momentum parameter and set to 0.99.

NPCface \cite{zeng2020npcface} observed that high possibility of co-occurrence of hard positive and hard negative appeared in large-scale dataset. It means that if one sample with ground truth class $i$ is hard positive, it has a larger chance to be a hard negative sample of class $j$. Therefore, NPCface emphasizes the training on both negative and positive hard cases via the collaborative margin mechanism in the softmax logits, which is shown as:

\begin{equation}
L_i = -\log\frac{e^{s\cos(\theta_{y_i}+\tilde {m_i})}}{e^{s\cos(\theta_{y_i}+\tilde {m_i})} + \sum_{j\ne y_i} e^{s\cos(\theta_{j}+\tilde {m_j})}}
\label{eq:npc}
\end{equation}
\begin{equation}
\tilde m_i=\left\{\begin{array}{ll}m_0+\frac{\sum_{j\ne y_i}(M_{i,j}\cos(\theta_j))}{\sum_{j\ne y_i}M_{i,j}} m_1, & \text{if} \ \sum_{j\ne y_i}M_{i,j} \neq 0 \\
m_0, & \text{if} \ \sum_{j\ne y_i}M_{i,j} = 0\end{array}\right.
\end{equation}
where $M_{i,j}$ is a binary indicator function, which represents whether sample $i$ is a hard negative sample of class $j$.

UniformFace \cite{duan2019uniformface} proposed that above large margin losses did not consider the distribution of all training classes.
With the prior that faces lie on a hypersphere manifold,
UniformFace imposed an equidistributed constraint by uniformly
spreading the class centers on the manifold, so that the
minimum distance between class centers can be maximized
through completely exploitation of the feature space. 
Therefore, the loss is :
\begin{equation}
L = L_{lml} + L_u
    \ \ , \ \ 
L_u = \frac{\lambda }{M(M-1)}
\sum^M_{j_1=1} \sum_{j_2\ne j_1} \frac
{1}{d(\boldsymbol c_{j_1}, \boldsymbol c_{j_2})}  
\end{equation}
where $L_{lml}$ can be any large margin loss above, such as CosFace or ArcFace; and $L_u$ is a uniform loss, which enforces the distribution of all training classes more uniform. 
$M$ is the number of classes in $L_u$, and $\boldsymbol c_i$ is the center of class $i$, $\lambda$ is a weight.
As the class centers $c_j$ are continuously changing during
the training, the entire training set is require to update $c_j$ in each iteration, which is not applicable in practice. Therefore, UniformFace updated the centers on each mini-batch by a newly designed step as follows: 
\begin{equation}
\Delta \boldsymbol  c_j = \frac{\sum_{i=1}^n \delta(y_i=j)\cdot(\boldsymbol  c_j - \boldsymbol  x_j )}{1 + \sum_{i=1}^n \delta(y_i=j)}
\end{equation}
where $n$ is the number of samples in a mini-batch, $x_j$ is a embedding in this batch, $\delta(.) = 1$
if the condition is true and $\delta(.) = 0$ otherwise.

Similar with \cite{duan2019uniformface}, Zhao \emph{et al.} \cite{Zhao_2019_CVPR} also considered the distribution of each face class in the feature space. They brought in the concept of `inter-class separability' in large-margin based loss and put forward RegularFace.  

The inter-class separability of the $i$-th identity class: $Sep_i$ is defined as the largest cosine similarity between class $i$ and other centers with different IDs, where:
\begin{equation}
Sep_i = \max_{j\neq i} \cos (\varphi_{i,j }) =\max_{j\neq i}\frac{W_i\cdot W_j}{\|W_i\|\cdot\|W_j\|}
\end{equation}
where $W_i$ is the $i$-th column of $W$ which represents the weight vector for ID $i$.
RegularFace pointed out that different ids should be uniformly distributed in the feature space, as a result, the mean and variance of $ Sep_i $ will be small. Therefore RegularFace jointly supervised the FR model with angular softmax loss
and newly designed exclusive regularization. The overall loss function is:
\begin{equation}
L = L_s + \lambda \mathcal L_r(W)
    \ \ , \ \ 
\mathcal L_r(W) = \frac{1}{C} \sum_i \max_{j\ne i} \frac{W_i\cdot W_j}{\|W_i\| \cdot \| W_j\|}
\end{equation}
where $C$ is the number of classes. $\lambda$ is the weight coefficient. $L_s$ is the softmax-based classification loss function or other large margin loss \cite{deng2019arcface,wang2018cosface}, and $\mathcal L_r(\theta, W) $ is an exclusive regularization, which provides extra inter-class separability and encourages uniform distribution of each class. 

Above large margin loss based classification framework only consider the distance representation between samples and class centers (prototypes). 
VPL \cite{deng2021variational} proposed a sample-to-sample distance representation and embedded it into the original sample-to-prototype classification framework.
In VPL, both sample-to-sample and sample-to-prototype distance is replaced by the distance between sample to its variational prototype.
In specific, the loss of VPL is:
\begin{equation}
L_i = -\log\frac{ e^{s\cos(\tilde \theta_{y_i} + m)} }{ e^{s\cos(\tilde \theta_{y_i} + m)} + \sum_{j \ne y_i} e^{s\cos(\tilde \theta_{y_i})}}
\end{equation}
where $\tilde \theta_{j}$ is the angle between the feature $x_i$ and the variational prototype $\tilde W_{j}$. The definition of $m$ and $s$ are similar with ArcFace \cite{deng2019arcface}: $m$ is the additive angular margin set as 0.5, and $s$ is the feature scale parameter set as 64.
Variational prototype $\tilde W_{j}$ is iteratively updated in training step as:
\begin{equation}
\tilde W_{j} =  \lambda_1 W_{j} + \lambda_2 M_{j} 
\end{equation}
where $W_{j}$ is the prototype of class $j$, namely the weight of class $j$. $M_{j}$ is the feature centers of all embeddings in the mini batch which are belongs to class $j$. $\lambda_1$ and $\lambda_2$ are their weights.

UIR \cite{yu2019unknown} trained the face feature extractor in semi-supervised way and modified original softmax loss by adding more unlabelled data. 
UIR design a loss $\lambda L_{uir}$ to measure the penalty on unlabelled training data, which dose not belong to any ID of labelled data. UIR assumed that, inference the unlabelled data on a well-trained face classifier, their logits scores of each class should be as equal as possible. Therefore, the assumption can be abstracted as the following optimization problem:
\begin{equation}
\max p_1\cdot p_2 \cdot ... \cdot p_n,  \ s.t. \sum_1^n p_i = 1
\end{equation}
where $n$ is the number of IDs in the training set. As a result, changing this optimization in a loss way, we have the UIR loss follows:
\begin{equation}
L_i = L_{softmax} + \lambda L_{uir}
    \ \ , \ \
L_{uir} = -\sum_{i=1}^n \log (p_i)
\end{equation}
where $L_{softmax}$ is softmax loss on labelled training data, and it can be replaced by other large margin loss. 

AdaCos \cite{zhang2019adacos} took CosFace \cite{wang2018cosface} as a benchmark and proposed new method to automatically modify the value of margin and scale hyper-parameter: $m$ and $s$ as the training iteration goes.

AdaCos observed that when $s$ is too small, the positive probability of one sample $i$ after margin based loss is also relatively small, even if the angular distance $\theta(i,y_i)$ between $i$ and its class center is small enough. On the contrary, when $s$ is too large, the positive probability of sample $i$ is reaching 1, even if the angular distance $\theta(i,y_i)$ is still large.

As a result, changing the value of $s$ during for a proper value will be good at FR model convergence.
The loss of AdaCos is similar with CosFace. But its scale parameter $s$ is changing as follows:
\begin{equation}
s^{(t)}= \left\{\begin{array}{ll}  \sqrt{2} \log (N-1),   &  t=0 
\\ \frac{\log B_{avg}^{(t)}}{\cos(\min(\pi/4 , \theta_{med}^{(t)}))} , 
& t>0 \end{array}\right.
\end{equation}
where $N$ is the number of training IDs; $t$ is the iteration times.
$\theta_{med}^{(t)}$ is the median of all corresponding classes' angles,
$\theta^{(t)}_{i,y_i}$ from the mini-batch at the $t$-th iteration.
$B_{avg}^{(t)}$ is the average of $B_i^{(t)}$, which is:
\begin{equation}
B_{avg}^{(t)} = \frac{1}{\|N^{(t)}\|} \sum_{i \in N^{(t)}} B_{i}^{(t)}
    \ \ , \ \
B_{i}^{(t)} = \sum_{k \neq y_i} e^{s^{(t-1)} \cos(\theta_{i,k}) }
\end{equation}
where $B_{i}^{(t)}$ is the total loss of all negative samples related to ID $i$. $N^{(t)}$ is the set of IDs in the mini-batch at $t$-th iteration, and $\|N^{(t)}\|$ is its number. The analysis of $s^{(t)}$ in detail can be check in the article of AdaCos \cite{zhang2019adacos}, which will not be elaborated here.

Similar with AdaCos \cite{zhang2019adacos}, Fair loss \cite{liu2019fair} also chose to adaptively modify the value of $m$ in each iteration, however it is controlled by reinforcement learning.
In detail, Fair loss is shown as follows:
\begin{equation}
L_i = -\log\frac{  P^*_{y_i}(m_i(t) , x_i)    }{  P^*_{y_i}(m_i(t) , x_i)   + \sum_{j \ne y_i}  P_j(x_i) }
\end{equation}
where $P_j(x_i) = e^{s \cos((\theta_j)}$ is the same with (\ref{eq:ArcFace}) and (\ref{eq:CosFace}).
The specific formulation of $P^*_{y_i}(m_i(t) , x_i)$ can be different according to different large margin loss.
Based on CosFace \cite{wang2018cosface} or ArcFace \cite{deng2019arcface}, $P^*$ can be formulated as follows:
\begin{equation}
P^*_{y_i}(m_i(t) , x_i)   = e^{s(\cos(\theta_{y_i})-m_i(t))}
    \ \ , or \ \
P^*_{y_i}(m_i(t) , x_i)   = e^{s(\cos(\theta_{y_i} + m_i(t)))}
\end{equation}
Then the problem of finding an appropriate margin adaptive strategy is formulated as a Markov Decision Process (MDP),
described by $(S,A, T ,R)$ as the states, actions, transitions and rewards. 
An agent is trained to adjust the margin in every state based on enforcement learning. 

Representative large margin methods such as CosFace \cite{wang2018cosface} and ArcFace 
have an implicit assumption that all the classes possess sufficient samples to describe its distribution, so that a manually set margin is enough to equally squeeze each intra-class variations.
Therefore, they set same values of margin $m$ and scale $s$ for all classes. 
In practice, data imbalance of different face IDs widely exists in

mainstream training datasets. 
For those IDs with rich samples and large intra-class variations, the space spanned by existing training samples can represent the real distribution.

But for the IDs with less samples, its features will be pushed to a smaller hyper space if we give it a same margin as ID with more samples.
AdaptiveFace \cite{liu2019adaptiveface} proposed a modified AdaM-Softmax loss.

by adjusting the margins for different classes adaptively. AdaM-Softmax modified from CosFace is shown as follows:
\begin{equation}
L_{ada} = -\log \frac{e^{s(\cos(\theta_{y_i}+m_{y_i}))}}{e^{s(\cos(\theta_{y_i}+m_{y_i}))} + \sum_{j \ne {y_i}} e^{s\cos(\theta_j)}}
\end{equation}
where the $m_{y_i}$ is the margin corresponding to class $y_i$ and it is learnable.
Intuitively, a larger $m$ is preferred to reduce the intra-class variations. Therefore, the final loss function of AdaptiveFace is:
\begin{equation}
L = L_{ada} + \lambda (-\frac{1}{N} \sum_{i} m_i)
\end{equation}
where $N$ is the number of IDs in the training set; and $\lambda$ is positive and controls the strength of the margin constraint. As the training goes by, AdaM-Softmax can adaptively allocate large margins to
poor classes and allocate small margins to rich classes.

Huang \emph{et al.} \cite{huang2020improving} thought the aforementioned large margin loss is usually fail on hard samples. 
As a result, they adopted ArcFace to construct a teacher distribution from easy samples and a student distribution from hard samples. 
Then a Distribution Distillation Loss (DDL) is proposed to constrain the student distribution to approximate the teacher distribution. DDL can lead to smaller overlap between the positive and negative pairs in the student distribution, which is similar with histogram loss \cite{ustinova2016learning}.
Ustinova \emph{et al.} \cite{ustinova2016learning} first obtained the histogram of similarity set of positive pairs $\mathcal S^+$ and that of negative pairs $\mathcal S^-$ in a batch as their distribution. Then histogram loss is formulated by calculating the probability that the similarity of $\mathcal S^-$ is greater than the one of $ \mathcal S^+$ through a discretized integral.
In DDL, Huang \emph{et al.} divided the training dataset into hard samples $\mathcal H$ and easy samples $\mathcal E$ 
\textcolor{red}{,} and constructed positive and negative pairs in sets $\mathcal H$ and $\mathcal E$\textcolor{red}{,} respectively. Then calculated the distribution of positive pairs similarity $H_r^+$ and the distribution of negative pairs similarity $H_r^-$, and calculated the KL divergence of $\mathcal E$ (as teacher) and $\mathcal H$ (as student) on the positive/negative pairs similarity distribution as loss, namely
\begin{equation}
\begin{aligned}
    \mathcal L_{KL} 
&= \lambda_1 \mathbb D_{KL}(P^+||Q^+) + \lambda_2 \mathbb D_{KL}(P^-||Q^-) \\
&=\lambda_1\sum_s P^+(s)\log\frac{P^+(s)}{Q^+(s)} + \lambda_2\sum_s P^-(s)\log\frac{P^-(s)}{Q^-(s)}
\end{aligned}
\end{equation}
where $\lambda_1$ and $\lambda_2$ are the weight coefficients. 
$P^+$ and $P^-$ are the similarity distributions of positive and negative pairs in $\mathcal E$.
$Q^+$ and $Q^-$ are the similarity distributions of positive and negative pairs in $\mathcal H$.
$\mathcal L_{KL}$ may make the distribution of teachers close to the distribution of students, so the author proposes order loss to add a constraint, namely
\begin{equation}
\mathcal L_{order} = -\lambda_3\sum_{(i,j)\in(p,q)}(\mathbb E[\mathcal S_i^+] - \mathbb E[\mathcal S_j^-])
\end{equation}
where $\mathcal S^+_p$ and $\mathcal S^-_p$ represent the positive pairs and negative pairs of the teacher, respectively, and $\mathcal S^+_q$ and $\mathcal S^-_q$ represent the student’s positive pairs and negative pairs. 
The final form of DDL is the sum of $\mathcal L_{KL}$, $\mathcal L_{order}$ and $L_{ArcFace}$.

In equation (\ref{eq:norm}), (\ref{eq:CosFace}), (\ref{eq:ArcFace}), (\ref{eq:sv}), (\ref{eq:svx}), (\ref{eq:curri}) and (\ref{eq:npc}), all face embeddings $x$ and class weights ${W}$ are normalized and thus distributed on the hypersphere manifold.
And Ring loss \cite{zheng2018ring} explicitly set 
a constrain where the length of embeddings should be as same as possible.
However, many papers have demonstrated that the magnitude of face embedding can measure the quality of the given face. 
It can be proven that the magnitude of the feature embedding monotonically increases if the subject is more likely to be recognized. As a result, MagFace \cite{meng2021magface} introduced an adaptive mechanism to learn a well-structured within-class feature distributions by pulling easy
samples to class centers with larger magnitudes while pushing hard samples away and shirking their magnitudes. In specific, the loss of MagFace is:
\begin{equation}
L_i = -\log \frac{e^{s\cos(\theta_{y_i}+m(a_i))}}{e^{s\cos(\theta_{y_i}+m(a_i))} + \sum_{j \ne {y_i}} e^{s\cos(\theta_j)}}+ \lambda_g g(a_i)
\end{equation}
where $a_i$ is the magnitude of face feature of sample $i$ without normalization. $m(a_i)$ represents a magnitude-aware angular
margin of positive sample $i$, which is monotonically increasing.
$g(a_i)$ is a regularizer and designed as a monotonically decreasing convex function.
$m(a_i)$ and $g(a_i)$ simultaneously enforce direction and magnitude of face embedding, and $\lambda_g$ is a parameter balancing these two factors.
In detail, magnitude $a_i$ of a high quality face image is large, and $m(a_i)$ enforces the embedding $x_i$ closer to class center $W_i$ by giving a larger margin; and $g(a_i)$ gives a smaller penalty if $a_i$ is larger. 

Circle loss \cite{sun2020circle} analyzed that, the traditional loss functions (such as triplet and softmax loss) are all optimizing $(s_n-s_p)$ distance, where $s_n$ is inter-class similarity and $s_p$ is intra-class similarity. 
This symmetrical optimization has two problems: inflexible optimization and fuzzy convergence state.
Based on this two facts, Sun \emph{et al.} proposed Circle loss, where greater penalties are given to similarity scores that are far from the optimal results. 

Similar with MagFace, Kim \emph{et al.} \cite{kim2022adaface} also emphasize misclassified samples should be adjusted according to their image quality.
They proposed AdaFace to adaptively control gradient changing during back propagation. Kim \emph{et al.} assume that hard samples should be emphasized when the image quality is high, and vice versa.
As a result, AdaFace is designed as:
\begin{equation}
L_i = -\log
 \frac
{   e^{s  \cos(\theta_{y_i} + g_{\textmd{angle}})-g_{\textmd{add}}  }  }
{ e^{s  \cos(\theta_{y_i} + g_{\textmd{angle}})-g_{\textmd{add}}  }  + \sum_{j \ne y_i} e^{s\cos(\theta_j)}}
\label{eq:adaface}
\end{equation}
\begin{equation}
g_{\textmd{angle}}= -m \cdot \widehat{\|z_i\|} , \ \   g_{\textmd{add}}= m \cdot \widehat{\|z_i\|} + m , \ \ 
\widehat{\|z_i\|} = \lfloor \frac{ \|z_i\| - \mu_z}{\sigma_z / h}\rceil^1_{-1}
\end{equation}
where $\|z_i\|$ measures the quality of face $i$, and $\widehat{\|z_i\|}$ is normalized quality by using batch statistics $\mu_z$ and $\sigma_z$ with a factor $h$. Similar with Arcface and CosFace, $m$ represents the angular margin.
AdaFace can be treated as the generalization of Arcface and CosFace: when $\widehat{\|z_i\|}=-1$, function (\ref{eq:adaface}) becomes ArcFace; when $\widehat{\|z_i\|}=0$, it becomes CosFace.

At the end of this subsection, we introduce a FR model quantization method, which was specially designed for large margin based loss.
Traditionally, the quantization error for feature $\boldsymbol f_i$ is defined as follows:
\begin{equation}
{\rm QE}(\boldsymbol f_i) = \frac{1}{d}\sum_{l=1}^d(\boldsymbol f_i^l- Q(\boldsymbol f_i^l))^2
\end{equation}
where $\boldsymbol f_i$ and $Q(\boldsymbol f_i)$ denote a full precision (FP) feature and its quantization. $d$ and the superscript $l$ represent the length of features and the $l$-th dimension.
Wu \emph{et al.} \cite{wu2020rotation} redefined the quantization error (QE) of face feature as the angle between its FP feature and its quantized feature :
\begin{equation}
{\rm AQE}(\boldsymbol f_i) = \arccos\left(\left<
\frac{\boldsymbol f_i}{\left\| \boldsymbol f_i \right\|_2} , \frac{{\tilde {Q}}( \boldsymbol f_i)}{\left\| {\tilde {Q}}( \boldsymbol f_i) \right\|_2}
\right>\right)
\end{equation}

Wu \emph{et al.} \cite{wu2020rotation} believed that for each sample, quantization error $\rm AQE$ can be divided into two parts: error caused by the category center of the sample after quantization (class error), and error caused by the sample deviating from the category center due to quantization (individual error), namely
\begin{equation}
    {\rm AQE}(\boldsymbol f_i)={\rm AQE}(\boldsymbol c_{y_i}) + \mathcal{I} (\boldsymbol f_i)
\end{equation}
The former term (class error of class $y_i$) will not affect the degree of compactness within the class, while the latter  term (individual error) will. Therefore, the individual error should be mainly optimized. They introduced the individual error into CosFace as an additive angular margin, named rotation consistent margin (RCM):
\begin{equation}
\mathcal L_i = -\log\frac{\exp(s\cdot\cos(\theta_{i,j} + \delta(j=y_i) \cdot m+ \delta(j=y_i) \cdot\lambda\theta_Q)}{\sum_j\exp(s\cdot\cos(\theta_{i,j} + \delta(j=y_i) \cdot m+ \delta(j=y_i) \cdot\lambda\theta_Q)}
\end{equation}
\begin{equation}
\theta_Q = \| \mathcal{I}(\boldsymbol f_i) \| = \| {\rm AQE}(\boldsymbol f_i)-{\rm AQE}(\boldsymbol c_{y_i}) \|
\end{equation}
where $\delta(j=y_i)$ is the indicative function, where $j=y_i$ gives its value 1, otherwise 0.

\subsubsection{FR in unbalanced training data}\label{subsubsec4.1.3}

Large-scale face datasets usually exhibit a massive number of classes with unbalanced distribution.
Features with non-dominate IDs are compressed into a small area in the hypersphere, leading to training problems. 
Therefore, for different data unbalance phenomena, different methods were proposed.

The first data unbalance phenomenon is long tail distributed, which widely exists in the mainstream training set, such as MS-Celeb-1M. 
In MS-Celeb-1M dataset, the number of face images per person falls drastically, and only a small part of persons have large number of images.
Zhang \emph{et al.} \cite{zhang2017range} set an experiment to show that including all tail data in training can not help to obtained a better FR model by contrastive loss \cite{sun2015deep}, triplet loss \cite{schroff2015facenet}, and center loss \cite{wen2016discriminative}. Therefore, the loss needs to be delicately designed.

Inspired by contrastive loss, range loss \cite{zhang2017range} was designed to penalize intra-personal variations especially for
infrequent extreme deviated value, while enlarge the inter-personal differences simultaneously. 
The range loss is shown as:
\begin{equation}
L = L_{softmax} + \alpha L_{R_{intra}} + \beta L_{R_{inter}} 
\end{equation}
where $\alpha$ and $\beta$ are two weights, $L_{R_{intra}}$ denotes the intra-class loss and $L_{R_{inter}}$ represents the inter-class loss.
$L_{R_{intra}}$ penalizes the maximum harmonic range within each class:
\begin{equation}
L_{R_{intra}} = \sum_{i \in I} L^i_{R_{intra}} = \sum_{i \in I} \frac{k}{\sum_{j=1}^k \frac{1}{D_j}} 
\end{equation}
where $I$ denotes the complete set of identities in current mini-batch, and $D_j$ is the $j$-th largest Euclidean distance between all samples with ID $i$ in this mini-batch. 
Equivalently, the overall cost is the harmonic mean of the first k-largest ranges within each class, and $k$ is set to 2 in the experiment.
$L_{R_{inter}}$ represents the inter-class loss that
\begin{equation}
L_{R_{inter}} = \max(M-D_{center} , 0) = \max(M-\|\overline{x}_{Q}- \overline{x}_{R}\|^2_2 , 0)
\end{equation}
where $D_{Center}$ is the shortest distance between the centers of two classes, 
and $M$ is the max optimization margin of $D_{Center}$. 
$Q$ and $R$ are the two nearest classes within the current mini-batch, while $\overline{x}_{Q}$ and $\overline{x}_{R}$ represents their centers.

Zhong \emph{et al.} \cite{zhong2019unequal} first adopted a noise resistance (NR) loss based on large margin loss to train on head data, which is shown as follows:
\begin{equation}
L_{NASB}(i) = - [\alpha(P_{y_{ip}}) \log(P_{y_i}) + \beta(P_{y_i}) \log (P_{y_{ip}})] 
\label{eq:nasb}
\end{equation}
where $y_i$ is the training label and $y_{ip}$ is the current predict
label.
$P_{y_{ip}}$ is the predict probability of training label class and 
$P_{y_i}$ is that of the current predict class. Namely:
\begin{equation}
y_{ip} = \arg \max_{y_i} \frac{e^{W_j^Tx_i+b_j}}{\sum_{k} e^{W_k^Tx_i+b_k}}
\end{equation}
\begin{equation}
P_{y_{i}}  = \frac{e^{W_{y_i}^Tx_i+b_{y_i}}}{\sum_{k} e^{W_k^Tx_i+b_k}}    \ , \
P_{y_{ip}} = \frac{e^{W_{y_{ip}}^Tx_i+b_{y_{ip}}}}{\sum_{k} e^{W_k^Tx_i+b_k}} 
\end{equation}
$\alpha(P)$ and $\beta(P)$ control the degree of combination:
\begin{equation}
\alpha(P)= 
\left\{
\begin{array}{ll}  
\rho,   &  P>t 
\\ 
0 , & P \le 0 
\end{array}\right.
 \ , \
\beta(P)= 
\left\{
\begin{array}{ll}  
1-\rho,   &  P>t 
\\ 
0 ,       & P \le 0 
\end{array}\right.
\end{equation}
The NR loss (\ref{eq:nasb}) can be further modified to CosFace \cite{wang2018cosface} or ArcFace \cite{deng2019arcface} forms.
After a relatively discriminative model have been learned on the head data by $L_{NASB}$, center-dispersed loss is employed to deal with the tail data. It extracts features of tail identities using the base model supervised by head data; 
then add the tail data gradually in an iterative way and disperse these identities in the feature space so that we can take full advantage of their modest but indispensable information.
To be more specifically, Center-dispersed Loss can be formulated as:
\begin{equation}
L_{CD} = \min \frac{1}{m(m-1)} \sum_{1 \le i < j \le m} S_{i,j}^2
    \ \ , \ \ 
S_{i,j} = (\frac{C_i}{\|C_i\|})^T(\frac{C_i}{\|C_i\|})  
\end{equation}
where $S_{i,j}$ is the similarity between identity $i$ and $j$ in mini-batch,
and the most hard $m$ identities are mined from a candidate bag to construct a tail data mini-batch for efficiency.
$C_i$ and $C_j$ represent normalized features centers of identity $i$ and $j$, which can be relatively robust even to moderate noise.

The second data unbalance phenomenon is shallow data. 
In many real-world scenarios of FR, the training dataset is limited in depth, i.e. only small number of samples are available for most IDs. 
By applying softmax loss or its angular modification loss (such as CosFace \cite{wang2018cosface}) on shallow training data, results are damaged by the model degeneration and over-fitting issues. 
Its essential reason consists in feature space collapse \cite{du2020semi}. 

Li \emph{et al.} \cite{li2021virface} proposed a concept of virtual class to give the unlabeled data a virtual identity in mini-batch, and treated these virtual classes as negative classes.
Since the unlabeled data is shallow such that it is hard to find samples from the same identity in a mini-batch, each unlabeled feature can be a substitute to represent the centroid of its virtual class.
As a result, by adding a virtual class term into the large margin based loss (such as ArcFace \cite{deng2019arcface}), the loss has: 
\begin{equation}
L_i = -\log \frac{e^{s(cos(\theta_{y_i}+m))}}{   e^{s(\cos(\theta_{y_i}+m))} + 
       \sum_{j \ne {y_i}} e^{s\cos(\theta_j)} + 
       \sum_{u =1}^U e^{s\cos(\theta_u)}}
\end{equation}
where $U$ is the number of unlabeled shallow data in mini-batch, and $\theta_u$ is the angular between embedding $x_i$ and $x_u$, which is also the centroid of virtual class $u$.
For the purpose of exploiting more potential of the unlabeled shallow data, a feature generator was designed to output more enhanced feature from unlabeled data. 
More details about the generator can be check in \cite{li2021virface}. 


The above methods solve data unbalance problem in class-level therefore treated the images from the same person with equal importance. However, Liu et al. \cite{liu2022learning} thought that they have different importance and utilized meta-learning to re-weighted each sample based on multiple variation factors. Specifically, it updated four learnable margins, each corresponding to a variation factor during training.

The factors included ethnicity, pose, blur, and occlusion.

\subsection{Embedding}\label{subsec4.2}
Different from designing delicate losses in the last subsection, embedding refinement is another way to enhance FR results. 
The first idea of embedding refinement set a explicit constraint on face embeddings with a face generator.
The second idea changed the face embedding with auxiliary information from training images, such as occlusion and resolution.
The third idea models FR in a multi-task way. Extra tasks such as age and pose prediction were added in the network. 

\subsubsection{Embedding refinement by face generator}\label{subsubsec4.2.1}
FR methods based on face generator usually focus on age or pose invariant FR problems. DR-GAN \cite{tran2017disentangled} solved pose-invariant FR by synthesizing faces with different poses. DR-GAN learned an identity representation for a face image by an encoder-decoder structured generator.
And the decoder output synthesized various faces of the same ID with different poses.
Given identity label $y^d$ and pose label $y^p$ of an face $x$, the encoder $G_{enc}$ first extract its pose-invariant identity representation $f(x)= G_{enc}(x)$. Then $f(x)$ is concatenated with a pose code $c$ and a random noise $z$. The decoder $G_{dec}$ generates the synthesized face image $\hat{x} = G_{dec}(f(x),c,z)$ with the same identity $y^d$ but a different pose specified by a pose code $c$. 
Given a synthetic face image from the generator, the discriminator $D$ attempts to estimate its identity and pose, which classifies $\hat{x}$ as fake.

\begin{figure}[htp]
    \centering
    \includegraphics[width=14cm]{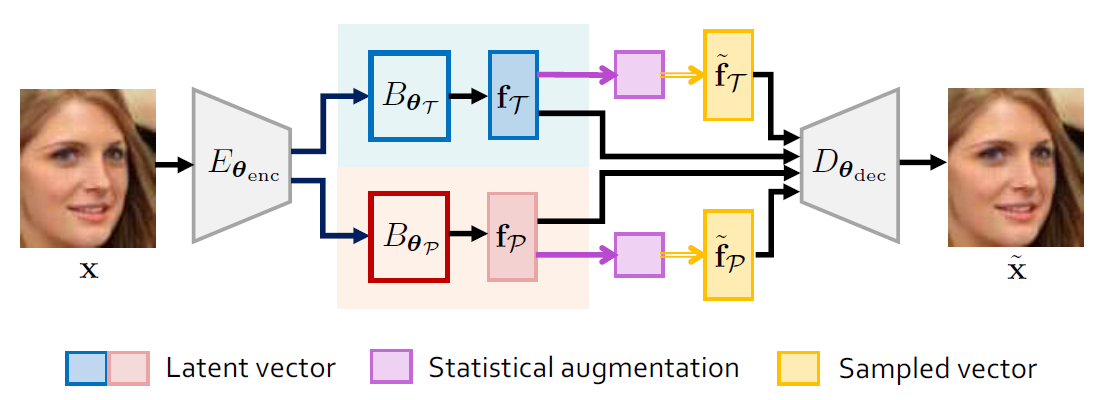}
    \caption{The architectures of R3AN.}
    \label{fig:d2ae2018}
\end{figure}

Liu \emph{et al.} \cite{liu2018exploring} proposed an identity Distilling and Dispelling Auto-encoder (D2AE) framework that adversarially learnt the identity-distilled features for identity verification. The structure of D2AE is shown in Fig. \ref{fig:d2ae2018}.
The encoder $E_{\theta_{enc}}$ in D2AE extracted a feature of an input image $x$, which was followed by
the parallel identity distilling branch $B_{\theta_{T}}$ and identity dispelling branch $B_{\theta_{P}}$. 
The output $f_T= B_{\theta_T}(E_{\theta_{enc}}(x))$ and $f_P= B_{\theta_P}(E_{\theta_{enc}}(x))$ are identity-distilled feature and identity-dispelled feature.
$f_T$ predicted the face ID of $x$ by optimizing a softmax loss.
On the contrary, $B_{\theta_{P}}$ needs to fool the identity classifier, where the so-called
``ground truth'' identity distribution is required to be constant over all identities and equal to $\frac{1}{N_{ID}}$ 
($N_{ID}$ is the number of IDs in training set). 
Thus, $f_P$ has a loss:
\begin{equation}
L_H = \frac{1}{N_{ID}} \sum_{j=1}^{N_{ID}} \log y_P^{j}
\end{equation}
where $y_P^{j}$ is the logits of this classifier with index $j$. The gradients for $L_H$ are back-propagated to $B_{\theta_{P}}$ and $E_{\theta_{enc}}$ with this identity classifier fixed.
At last, an decoder $D_{\theta_{dec}}$ is used to further enhance $f_T$ and $f_P$ by imposing a bijective mapping between an input image $x$ and its semantic features, with a reconstruction loss:
\begin{equation}
L_X = \frac{1}{2} \|x - D_{\theta_{dec}}(f_T,f_P)\|^2_2
\end{equation}
As shown in Fig. \ref{fig:d2ae2018}, $\tilde{f}_T$ and $\tilde{f}_P$ are the augmented feature of $f_T$ and $f_P$ by adding Gaussian noise on them, which can also be employed to train the decoder. The generated image with $\tilde{f}_T$ and $\tilde{f}_P$ needed to preserve the ID of input image $x$.

\begin{figure}[htp]
    \centering
    \includegraphics[width=14cm]{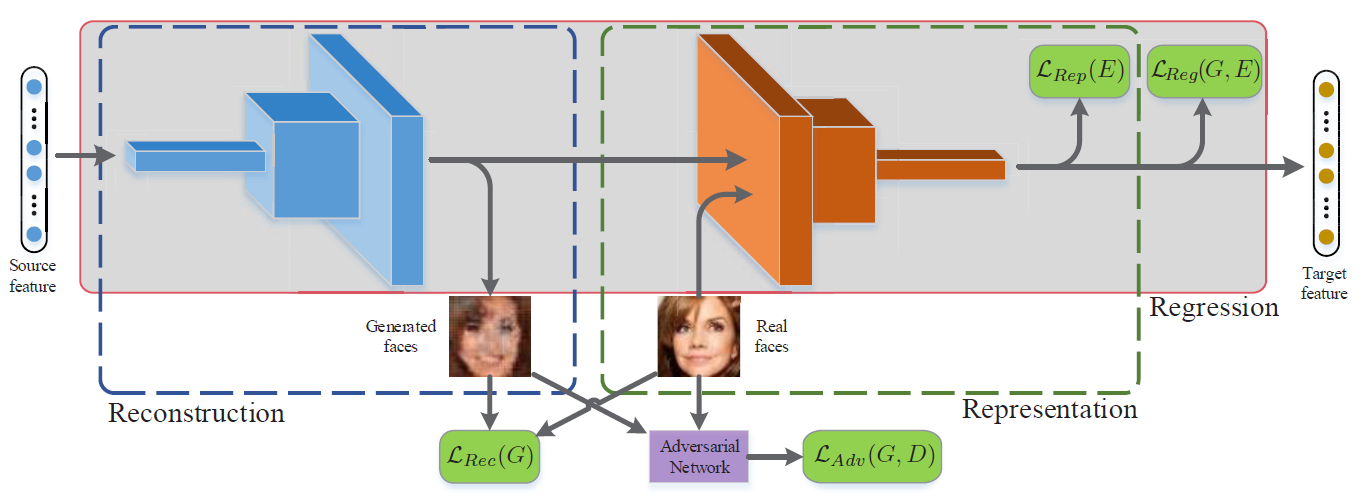}
    \caption{The architectures of R3AN.}
    \label{fig:r3an2019}
\end{figure}

Chen \emph{et al.} \cite{chen2019r3} brought forth cross model FR (CMFR) problem, and proposed R3AN to solve it. CMFR is defined as recognizing feature extracted from one model with another model’s gallery. Chen \emph{et al.} built a encoder to generate a face image of the source feature (from FR model 1), and trained a encoder to match with target feature (in FR model 2). 
The architecture in R3AN is shown in Fig. \ref{fig:r3an2019}.
The training of R3AN has 3 parts: reconstruction, representation and regression. 
In reconstruction, a generator $G$ is trained by source feature $X$ and its related face image $I$. 
The reconstruction loss is formulated as follows:
\begin{equation}
L_{Rec}(G) = \mathbb{E}_{X,I} [\|I - G(X)\|_2]
\end{equation}
Similar with GAN, the generated face image is needed adversarial learning $L_{Adv}$ to become as real as possible.
In representation, a encoder $E$ is trained, which takes the original face image $I$ as input and learns representation $Y$ of the target model.
An L2 loss is adopted to supervise the training of representation module as follows:
\begin{equation}
L_{Rep}(E) = \mathbb{E}_{I,Y} [\|Y - E(I)\|_2]
\end{equation}
Finally, regression module synchronizes the $G$ and $E$ in our feature-to-feature learning framework, and it maps source $X$ to target $Y$. As a result, the regression loss exists is expressed as:
\begin{equation}
L_{Reg}(G,E) = \mathbb{E}_{X,Y} [\|Y - E(G(X))\|_2]
\end{equation}

\begin{figure}[htpb]
    \centering
    \includegraphics[width=16cm]{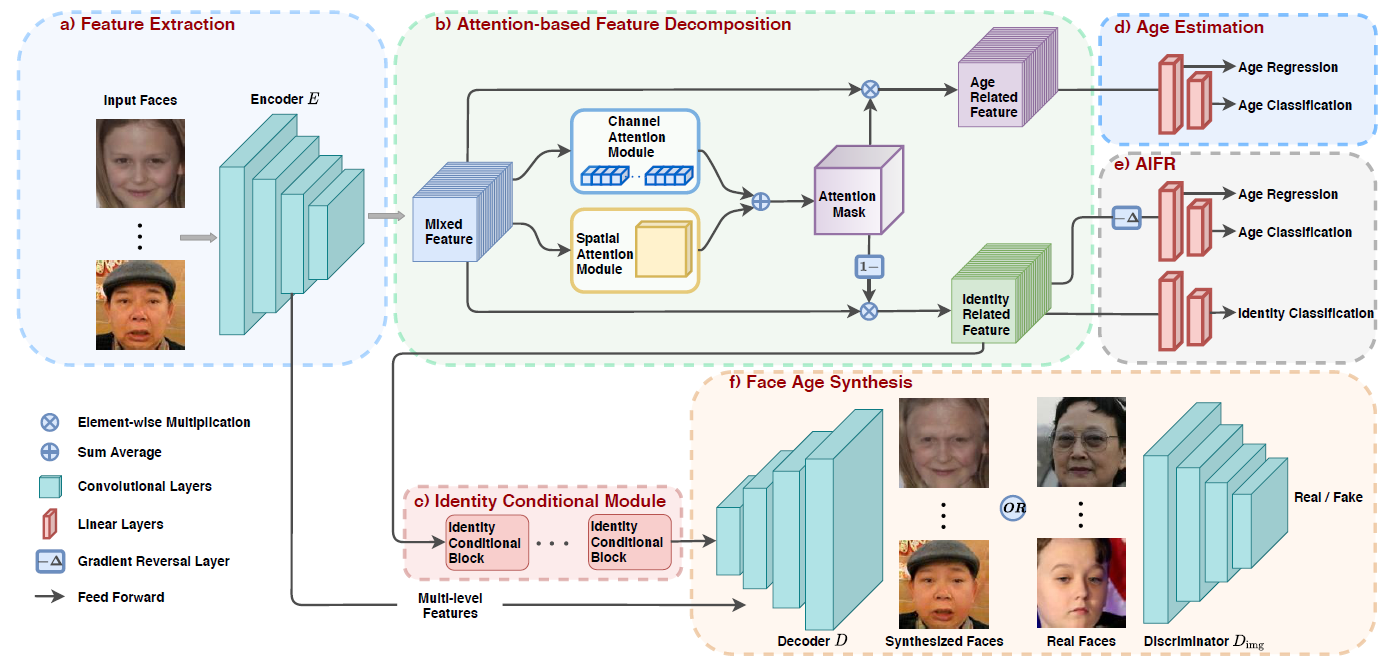}
    \caption{An overview of MTLFace. }
    \label{fig:age2021}
\end{figure}

Huang \emph{et al.} \cite{huang2021age} proposed MTLFace to jointly learn age-invariant FR and face age synthesis.
Fig. \ref{fig:age2021} depicts the pipeline of this method. 
First, an encoder $E$ extracts face feature $X$ of a image $I$, which is further fed into attention-based feature decomposition (AFD) module. AFD decomposes the feature $X$ into age related feature $X_{age}$ and identity related feature $X_{id}$ by attention mask, namely:
\begin{equation}
X = X_{age} + X_{id} = X \circ \sigma(X) + X \circ (1-\sigma(X))
\end{equation}
where $\circ$ denotes element-wise multiplication and $\sigma$ represents an attention module which is composed of channel and spacial attention modules.
Then $X_{age}$ and $X_{id}$ are used for age estimation and age-invariant FR (AIFR). 
In AIFR, CosFace \cite{wang2018cosface} supervises the learning of $X_{id}$.
In addition, a cross-age domain adversarial learning is proposed to encourage $X_{id}$ to be age-invariant with a gradient reversal layer (GRL) \cite{ganin2016domain}. The loss for AIFR is formulated as:
\begin{equation}
L^{AIFR} = L_{CosFace}(X_{id}) + \lambda_1 L_{AE}(X_{age}) + \lambda_2 L_{AE}(GRL(X_{id}))
\end{equation}
where $L_{AE}$ is the age estimation loss, which contains age value regression and age group classification. 
In order to achieve a better performance of AIFR, MTLFace designed a decoder to generate synthesized faces to modify $X_{id}$ explicitly.
In face age synthesis (FAS), $X_{id}$ is first fed into identity condition module to get a new feature with age information (age group $t$). Then a decoder is utilized to generate synthesized face $I_t$.
In order to make sure $I_t$ get correct ID and age information, $I_t$ is also fed into the encoder and AFD. The final loss of the generator becomes:
\begin{equation}
X_{age}^{t} , X_{id}^t = AFD(E(I_t))
    \ , \ \ 
L^{FAS}_{age} = L_{CE}(X_{age}^t,t)
    \ , \ \ 
L^{FAS}_{id} =  \mathbb{E} \|X_{id}^t - X_{id}\|^2_F 
\end{equation}
where $L^{FAS}_{age}$ constrains $I_t$ has age $t$, where $L_{CE}$ is cross-entropy loss. $L^{FAS}_{id}$ encourages the identity related features of $I$ and $I_t$ to get closer, where $\|\dot\|_F$ denotes the Frobenius norm.
Finally, MTLFace builds a discriminator to optimize $I_t$ to get a real looking appearance.

Uppal \emph{et al.} \cite{uppal2021teacher} proposed the Teacher-Student Generative Adversarial Network (TS-GAN) to generate depth images from single RGB images in order to boost the performance of FR systems.
TS-GAN includes a teacher and a student component. 
The teacher, which consists of a generator and a discriminator, aims to learn a latent mapping between RGB channel and
depth from RGB-D images. 
The student refines the learned mapping for RGB images by further training the generator.
While training FR, the model inputs a RGB image and its generates depth image, and extracts their features independently.
Then the final face embedding by fusing RGB and depth features is used to predict face ID.

\begin{figure}[htpb]
    \centering
    \includegraphics[width=15cm]{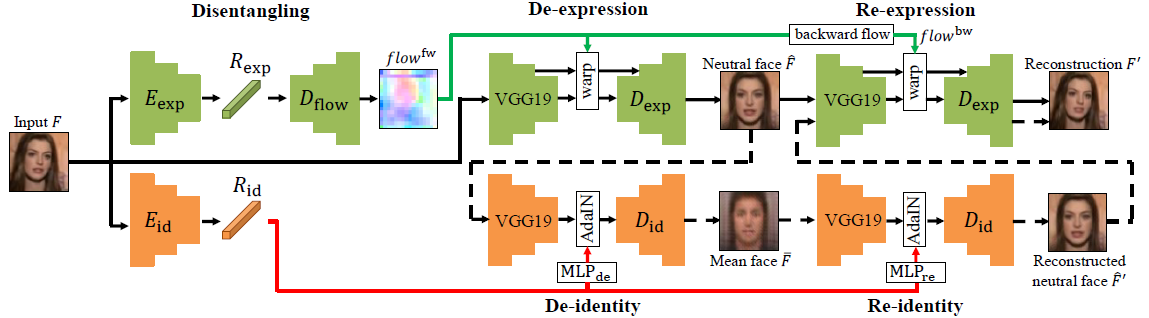}
    \caption{The architecture of \cite{chang2021learning}. }
    \label{fig:cycle2021}
\end{figure}

\cite{chang2021learning} learned facial representations from unlabeled facial images by generating face with de-expression. The architecture of the proposed model is shown in Fig. \ref{fig:cycle2021}.
In this method, a face image $F$ can be decomposed as $F = \tilde{F} + \textmd{id} + \textmd{exp} = \hat{F} + \textmd{exp}$, 
where $\tilde{F}$ is the global mean face shared among all the faces, and $\hat{F}$ is the neutral face of a particular identity specified by id.
id and exp are the identity and expression factors respectively.
In order to get $\tilde{F}$ and $\hat{F}$, a image $F$ is first used to extract expression and identity representations by using a unsupervised disentangling method. 
By exploring the disentangled representations, networks $D_{flow}$, $D_{exp}$ , MLPs and $D_{id}$
are trained to generate the representation-removed images $\tilde{F}$ and $\hat{F}$, and to reconstruct the representation-added images, the input face $F^{'}$and the neutral face $\hat{F}^{'}$.

\subsubsection{Embedding refinement by extra representations}\label{subsubsec4.2.2}

Both \cite{iliadis2017robust} and \cite{dong2019low} considered face embedding as a low-rank representation problem.
Their frameworks aim at adding noise into images, which can be divided into face features linear reconstruction (from a dictionary) and sparsity constraints.

Neural Aggregation Network (NAN) \cite{yang2017neural} is a typical video FR method by manipulating face embeddings. Yang \emph{et al.} proposed that, face images (in a video) with a same ID should be merged to build one robust embedding. Given a set of features with a same ID from one video $\{ f_i \lvert i=1,2,\dots,K \}$, it will be merged into an embedding $r$ by weighted summation as:
\begin{equation}
r = \sum_{i=1}^{K} a_i f_i \ , \ \sum_i a_i = 1
\end{equation}
where $a_i$ is weight for feature $f_i$. And $a_i$ is learned by a neural aggregation network, which is based on stacking attention blocks.

He \emph{et al.} \cite{he2018dynamic} proposed the Dynamic Feature Matching method to address partial face images due to  occlusion or large pose.
First, a fully convolutional network is adopted to get features from probe and gallery face image with arbitrary size, which are denoted as $p$ and $g_c$ ($c$ is the label of the gallery image).
Normally, it fails to compute the similarity of $p$ and $g_c$ on account of feature dimension-inconsistent.
As a result, a sliding window of the same size as $p$ is used to decompose $g_c$ into $k$ sub-feature maps $G_c=[g_{c_1},g_{c_2},\dots,g_{c_k}]$.
Then the coefficients $w_c$ of $p$ with respect to $G_c$ is computed by following loss:
\begin{equation}
L(w_c) = y_c (\| p- G_c w_c\|^2_2 - \alpha p^T G_c w_c )+ \beta \|w_c\|_1 
\end{equation}
where term $p^T G_c w_c$ is the similarity-guided constraint, and term $\beta \|w_c\|_1$ is a l1 regularizer. $\alpha$ and $\beta$ are constants that control the strength of these two constraints. $yc = \{1,-1\}$ means that $p$ and $G_c$ are from the same identity or not. 

\begin{figure}[htp]
    \centering
    \includegraphics[width=15cm]{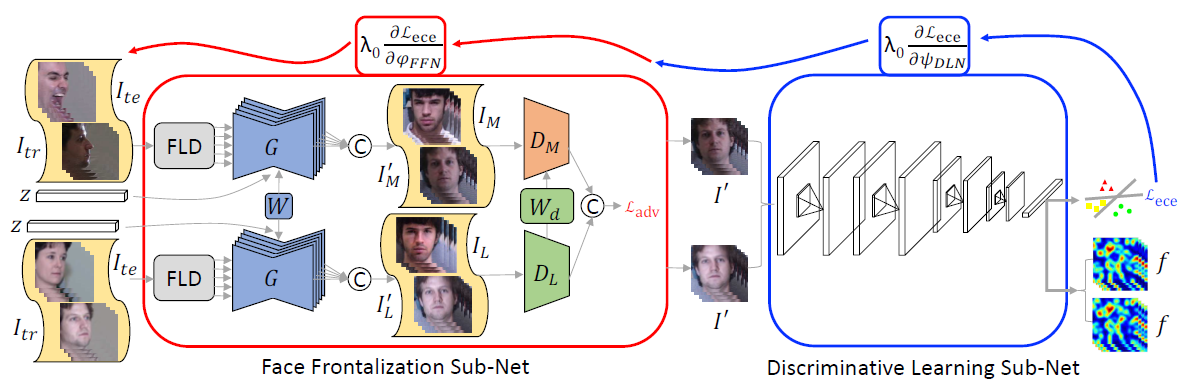}
    \caption{An overview of the PIM framework.}
    \label{fig:pose2018}
\end{figure}

Zhao \emph{et al.} \cite{zhao2018towards} proposed the Pose Invariant Model (PIM) for FR in the wild, by the Face Frontalization Sub-Net (FFSN).
First, a face image is input into a face landmark detector to get its landmark patches. The input of PIM are profile face images with four landmark patches, which are collectively denoted as $I_{tr}$. Then the recovered frontal face is $I^{\prime} =G(I_{tr})$, where $G$ is a encoder-decoder structure.
Similar with traditional GAN, a discriminative learning sub-net is further connected to the FFSN, to make sure $I^{\prime}$ visually resemble a real face with identity information.
As a result, features from a profile face with its generated frontal face will be used to get a better face representation. 
An overview of the PIM framework is shown in Fig. \ref{fig:pose2018}

Yin \emph{et al.} \cite{yin2019towards} proposed a Feature Activation Diversity (FAD) loss to enforce face representations to be insensitive to local changes by occlusions.
A siamese network is first constructed to learn face representations from two faces: one with synthetic occlusion $I$ and one without $I$.

\begin{figure}[htp]
    \centering
    \includegraphics[width=15cm]{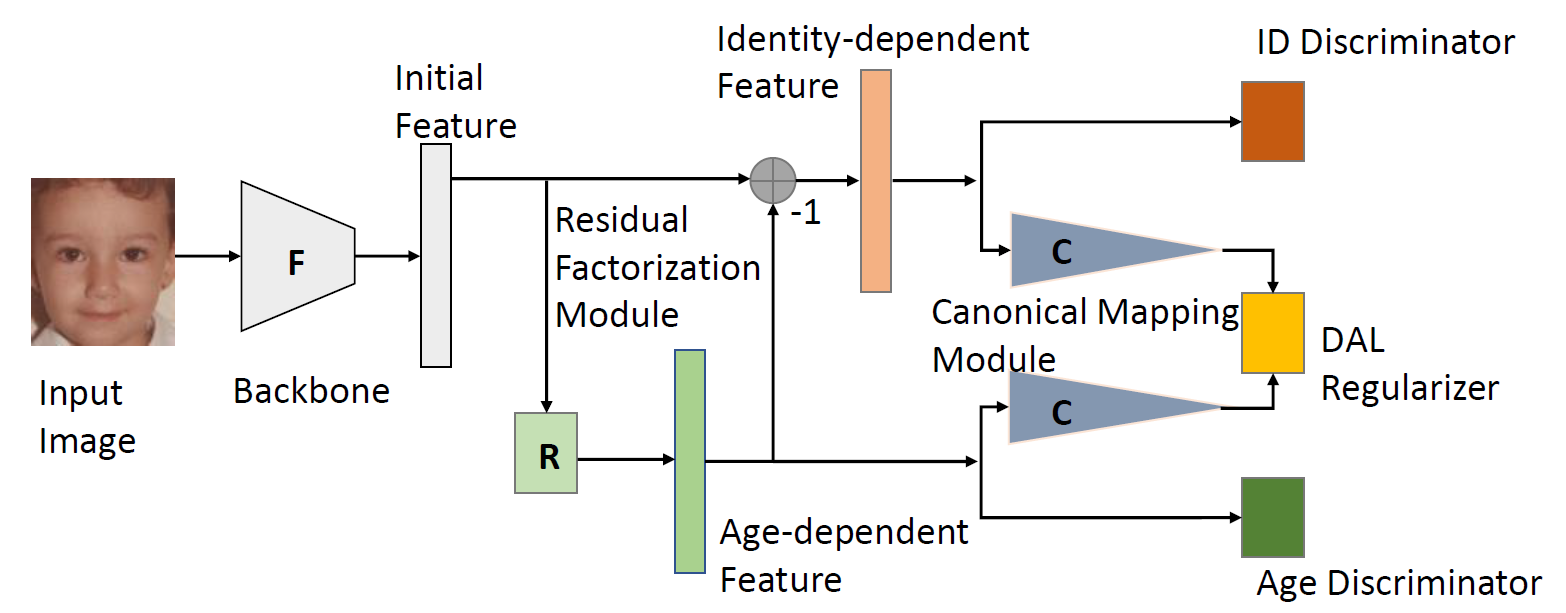}
    \caption{An overview of the proposed method of \cite{wang2019decorrelated}.}
    \label{fig:deco2019}
\end{figure}

Wang \emph{et al.} \cite{wang2019decorrelated} solved age-invariant FR problem by factorizing a mixed face feature into
two uncorrelated components: identity-dependent component and age-dependent component. The identity dependent component includes information that is useful for FR, and age-dependent component is treated as a distractor in the problem of FR.
The network proposed in \cite{wang2019decorrelated} is shown in Fig. \ref{fig:deco2019}.
The initial feature $x$ of a face image is extracted by a backbone net $F$, followed by the residual factorization module. 
The two factorized components $x_{id}$ (identity-dependent component) and $x_{age}$ (age-dependent component), where $x = x_{id} + x_{age}$.
$x_{age}$ is obtained through a mapping function $R$ ($x_{age} = R(x)$), 
and the residual part is regarded as $x_{id}$ ($x_{id} = x-R(x)$)
Then, $x_{id}$ and $x_{age}$ are used for face ID classification and age prediction.
In addition, a Decorrelated Adversarial Learning (DAL) regulizer is designed to reduce the correlation between the decomposed features, namely:
\begin{equation}
L_{DAL} = \lvert \rho  \rvert = \lvert \frac{Cov(v_{id},v_{age})}{\sqrt{Var(v_{id})Var(v_{age})}} \rvert
\end{equation}
\begin{equation}
v_{id} = c(x_{id}) = w_{id}^T x_{id} \ , \\  v_{age} = c(x_{age}) = w_{age}^T x_{age}
\end{equation}
where variables $v_{id}$ and $v_{age}$ are mappings from $x_{id}$ and $x_{age}$ by a linear Canonical Mapping Module, and $w_{id}$ and $x_{age}$ are the learning parameters for canonical mapping.
$\rho$ is the correlation coefficient. A smaller value of $\lvert \rho \rvert$ represents the irrelevance between $v_{id}$ and $v_{age}$, which means $x_{age}$ is decoupled with $x_{id}$ successfully.

\begin{figure}[htp]
    \centering
    \includegraphics[width=15cm]{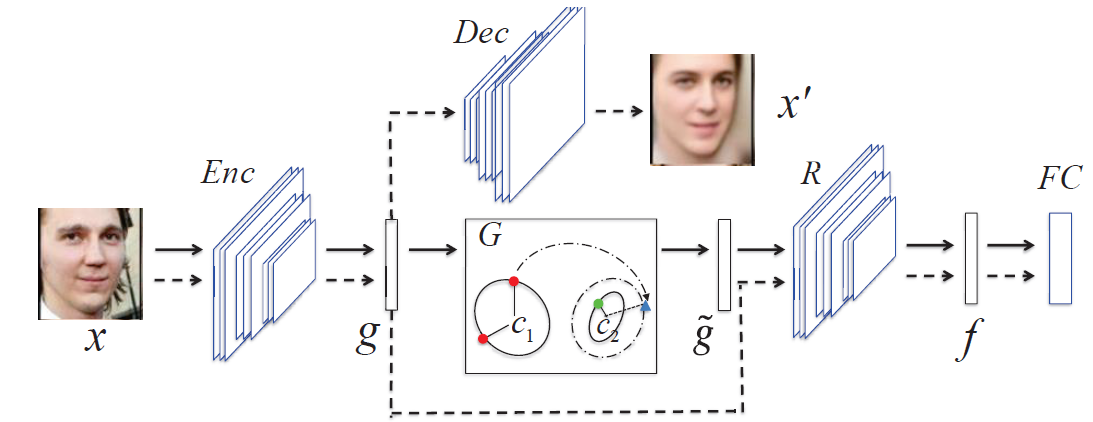}
    \caption{An overview of the proposed method of \cite{wang2019decorrelated}.}
    \label{fig:trans2019}
\end{figure}

Yin \emph{et al.} \cite{yin2019feature} proposed a feature transfer framework to augment the feature space of under-represented subjects with less samples from the regular subjects with sufficiently diverse samples.
The network (which can be viewed in Fig. \ref{fig:trans2019}) is trained with an alternating bi-stage strategy. 
At first stage, an encoder $Enc$ is fixed and generates feature $g$ of a image $x$. Then the feature transfer $G$ of under-represented subjects is applied to generate new feature samples $\tilde g$ that are more diverse. 
These original and new features $g$ and $\tilde g$ of under-represented subjects will be used to reshape the decision boundary. 
Then a filtering network $R$ is applied to generate discriminative identity features f = R(g) that are fed to a linear
classifier $FC$ with softmax as its loss.
In stage two, we fix the FC, and update all the other models. 
As a result, the samples that are originally on or across the boundary are pushed towards their center.
Also, while training the encoder, a decoder $Dec$ is added after feature $g$ to recover the image $x$ with L2 loss. 

PFE \cite{shi2019probabilistic} proposed that, an image $x_i$ should have an ideal embedding $f(x_i)$ representing its identity and less unaffected by any identity irrelevant information. The $n(x_i)$ is the uncertainty information of $x_i$ in the embedding space.
So, the embedding predicted by DNNs can reformulated as $z_i = f(x_i)+n(x_i)$, where $z_i$ can be defined as a Gaussian distribution: $p(z_i \lvert x_i) = N(z_i ; \mu_i , \sigma^2_i I)$.
Inspired by \cite{shi2019probabilistic}, \cite{chang2020data} proposed Data Uncertainty Learning (DUL) to extract face feature (mean) and its uncertainty (variance) simultaneously.
Specifically, DUL first sample a random noise $\epsilon$ from a normal distribution, 
and then generate $s_i$ as the equivalent sampling representation: 
$s_i = \mu_i + \epsilon \sigma_i , \epsilon \sim N(0,I) $.
As a result, a classification loss $L_{softmax}$ can be used to optimize representation $s_i$.
In addition, an regularization term $L_{kl}$ explicitly constrains $N(\mu_i , \epsilon_i)$ to be close to a normal distribution, $N(0, I)$, measured by Kullback-Leibler divergence (KLD). 
Therefore, the final loss became $L = L_{softmax}+ \lambda L_{kl}$, where:
\begin{equation}
L_{kl} = KL[ N(z_i \lvert \mu_i , \sigma_i^2) \lvert \rvert N(\epsilon_i \lvert 0 ,I)] = -\frac{1}{2}(1+ \log \sigma^2 - \mu^2 - \sigma^2 )
\end{equation}

\cite{li2021spherical} pointed out the failure of PFE theoretically, and addressed its issue by extending the von Mises Fisher density to its r-radius counterpart and deriving a new optimization objective in closed form.

\begin{figure}[htp]
    \centering
    \includegraphics[width=15cm]{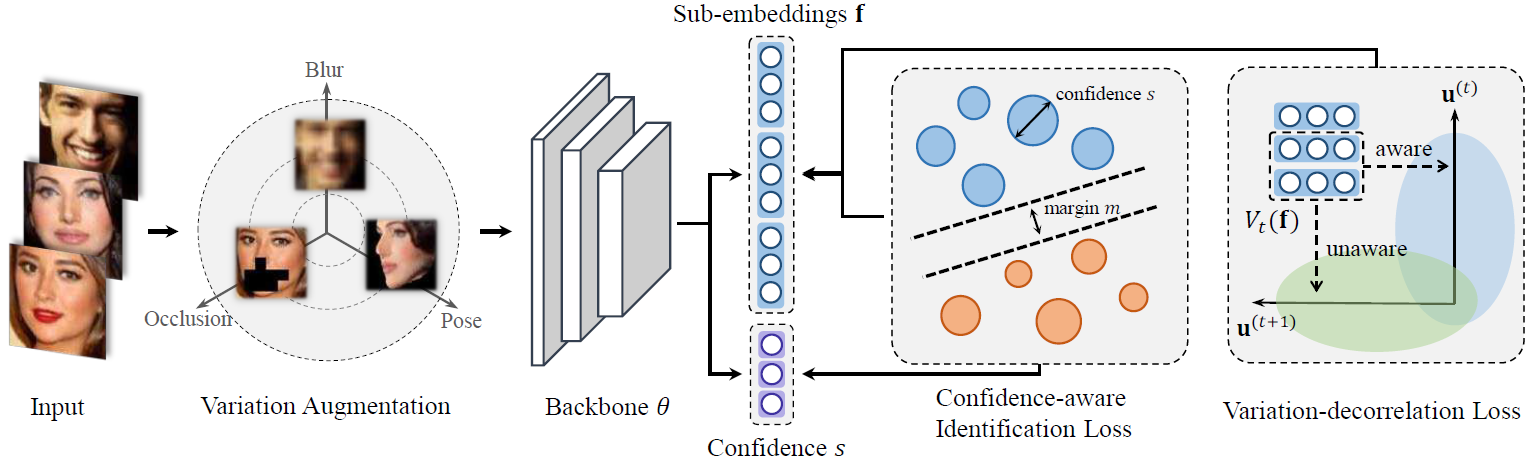}
    \caption{The framework of the proposed method.}
    \label{fig:univer2020}
\end{figure}

Shi \emph{et al.} \cite{shi2020towards} proposed a universal representation learning method for FR.
In detail, a high-quality data augumentation method was used according to pre-defined variations such as blur, occlusion and pose. 
The feature representation extracted by a backbone is then split into sub-embeddings associated with sample-specific confidences.
Confidence-aware identification loss and variation decorrelation loss are developed to learn the sub-embeddings.
In specific, confidence-aware identification loss is similar with the CosFace, where the confidences of each image are used as scale parameter in the Equation (\ref{eq:CosFace}). The framework is shown in Fig. \ref{fig:univer2020}.

\begin{figure}[htp]
    \centering
    \includegraphics[width=15cm]{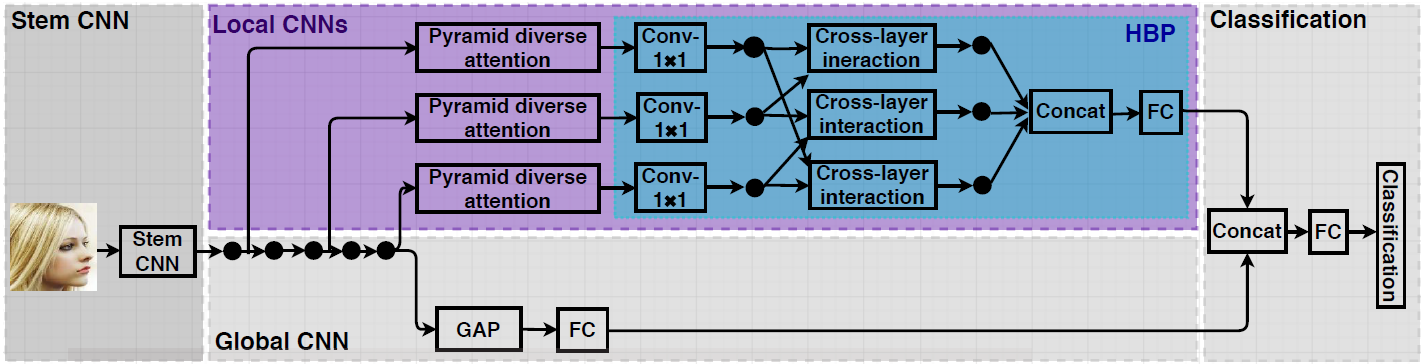}
    \caption{The framework of the proposed hierarchical pyramid diverse attention (HPDA) model.}
    \label{fig:hpda2020}
\end{figure}

\cite{wang2020hierarchical} proposed a hierarchical pyramid diverse attention (HPDA) network to learn multi-scale diverse local representations adaptively. 
Wang \emph{et al.} observed that face local patches played important roles in FR when the global face appearance changed dramatically. 
A face feature is first extracted by a stem CNN, then fed into a global CNN with global average pooling and fully connected layer to get global feature. 
At the same time, Local CNNs are developed to extract multi-scale diverse local features hierarchically.
Local CNNs mainly consist of a pyramid diverse attention (PDA) and a hierarchical bilinear pooling (HBP).
The PDA aims at learning local features at different scales by the Local Attention Network \cite{wang2019ls}. 
The HBP aggregates local information from hierarchical layers to obtain a more comprehensive local representation.
At last, local and global features are concatenated together to get the final rich feature for classification.  
The framework of HPDA is shown in Fig. \ref{fig:hpda2020}.

\begin{figure}[htp]
    \centering
    \includegraphics[width=15cm]{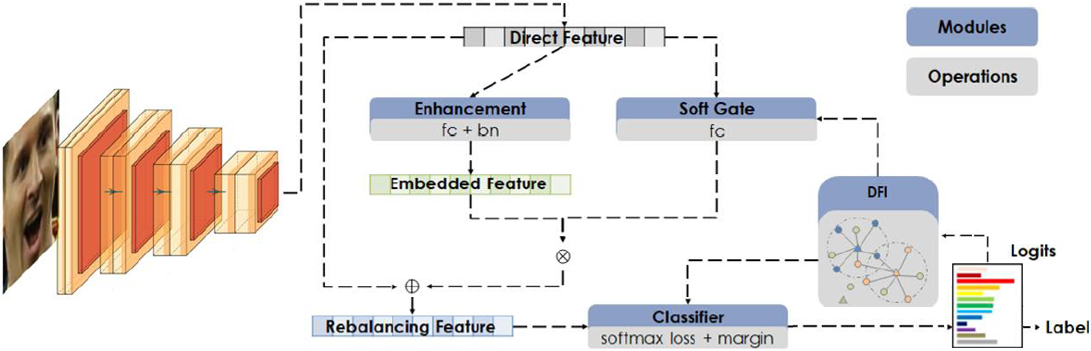}
    \caption{An overview of DB mechanism with three main modules: DFI, RBM and DBM.}
    \label{fig:long2020}
\end{figure}

In order to handle long-tail problem \cite{zhang2017range,zhong2019unequal} in FR, \cite{cao2020domain} proposed the Domain Balancing (DB) mechanism to obtain more discriminative features of long-tail samples, which contains three main modules: the Domain Frequency Indicator (DFI),  the Residual Balancing Mapping (RBM) and the Domain Balancing Margin (DBM).
The DFI is designed to judge whether a sample is from head domains or tail domains, based on the inter-class compactness. 
The classes with smaller compactness (larger DFI value) are more likely to come from a tail domain and should be relatively upweighted. DFI value is calculated by the weights in the final classifier.
The light-weighted RBM block is applied to balance the domain distribution. 
RBM block contains a soft gate $f(x)$ and a feature enhancement module $R(x)$. $f(x)$ is used to measure the feature $x$ depending on DFI value and $R(x)$ is a boost for feature $x$ when it comes from tail samples. 
In RBM block, if feature $x$ probably belongs to a tail class, a large enhancement is assigned to the output rebalancing feature $x_{re}$, namely $x_{re} = x + f(x)R(x)$.
Finally, the DBM in the loss function to further optimize the feature space of the tail domains to improve generalization, by embedding the DFI value into CosFace:
\begin{equation}
L_{dbm} = - \log \frac{e^{s(\cos(\theta_{y_i})- \beta_{y_i} m)}}
                      {e^{s(\cos(\theta_{y_i})- \beta_{y_i} m)} + \sum_{j \ne y_i} e^{s\cos(\theta_j)}}
\end{equation}
where $\beta_{y_i}$ is the DFI value of class $y_i$. A large DFI value of $y_i$ shows that class $y_i$ is in tail domain and it gets a larger margin while training.
An overview of this network is shown in Fig. \ref{fig:long2020}.

\begin{figure}[htp]
    \centering
    \includegraphics[width=15cm]{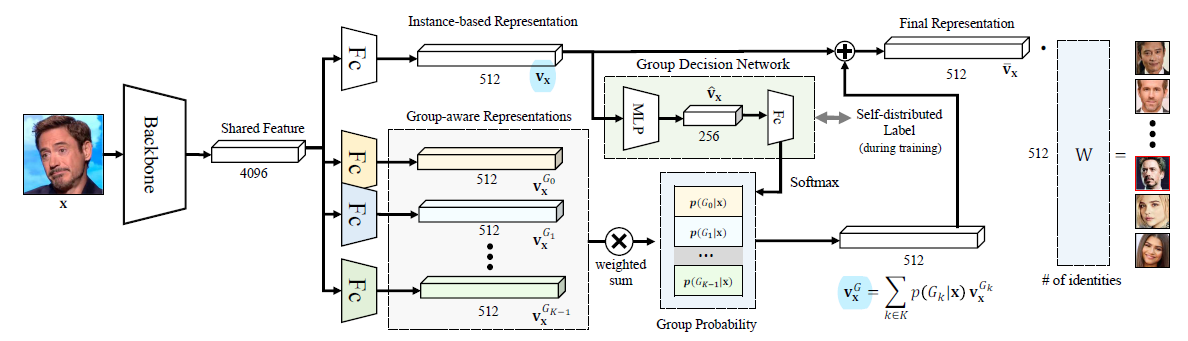}
    \caption{An overview of GroupFace.}
    \label{fig:group2020}
\end{figure}

GroupFace \cite{kim2020groupface} learned the group-aware representations by providing self-distributed labels that balance the number of samples belonging to each group without additional annotations, which can narrow down the search space of the target identity. 
In specific, given a face sample $x$, GroupFace first extracts a shared feature and deploys a FC layer to get an instance-based representation $v_x$ and 
K FC layers for group-aware representations $v_x^G$ (K is set to 32 in their experiments with best performance).
Here, a group is a set of samples that share any common visual-or-non-visual features that are used for FR.
Then, a group decision network, which is supervised by the self-distributed labeling, outputs a set of group probabilities 
$\{ p(G_0 \lvert x), p(G1\lvert x), \dots , p(G_{K-1}\lvert x) \}$ 
from the instance-based representation. 
The final representation $\bar{v}_x$ is an aggregation of the instance-based representation and 
the weighted sum $v_x^G$ of the group-aware representations with the group probabilities.
At last, a FC with weights $W$ is adopted to predict face ID, where ArcFace is used to train the network.
In training, in order to construct the optimal group-space, a self-grouping loss, which reduces
the difference between the prediction and the self-generated label, is defined as:
\begin{equation}
L_{sg} = -\frac{1}{N} \sum_{i=1}^N \textmd{CrossEntropy}( \textmd{softmax}(f(x_i)), G^*(x_i))
\end{equation}
\begin{equation}
G^*(x) = \arg \max_{k} \tilde{p}(G_k \lvert x)
    \ \ , \ \
\tilde{p}(G_k \lvert x) = \frac{1}{K}[ p(G_k \lvert x)-\frac{1}{K}] + \frac{1}{K}
\end{equation}
where $N$ is the number of samples in a minibatch. $f(x_i)$ is the output logits of the final FC (prediction). $G^*(x_i)$ represents self-generated label, which is the optimal self-distributed label with largest group probability.
Therefore, the final loss is $L_{ArcFace} + L_{sg}$.
The structure of GroupFace is shown in Fig. \ref{fig:group2020}.

\begin{figure}[htp]
    \centering
    \includegraphics[width=15cm]{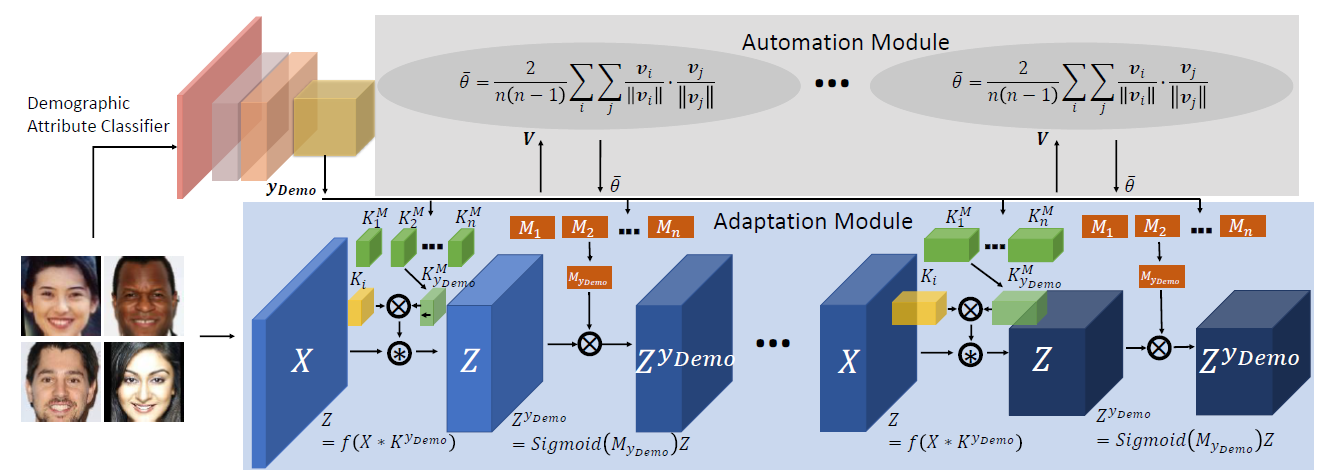}
    \caption{An overview of the GAC for mitigating FR bias.}
    \label{fig:gac2021}
\end{figure}

Gong \emph{et al.} \cite{gong2021mitigating} thought the faces of every demographic group should be more equally represented. Thus they proposed an unbiased FR system which can obtain equally salient features for faces across demographic groups. 
The propose FR network is based on a group adaptive classifier (GCA) which utilizes dynamic kernels and attention maps to boost FR performance in all demographic groups.
GAC consists of two main modules, an adaptive layer and an automation module.
In an adaptive layer, face features are convolved with a unique kernel for each demographic group, and multiplied with adaptive attention maps to obtain demographic-differential features.
The automation module determines in which layers of the network adaptive kernels and attention maps should be applied.
The framework of GAC is shown in Fig. \ref{fig:gac2021}. 

Similar with \cite{huang2021age}, Hou \emph{et al.} \cite{hou2021disentangled} solved the AIFR problem by factorizing identity-related and age-related representations $x_{id}$ and $x_{age}$.
Then $x_{id}$ and $x_{age}$ were optimized by age and identity discriminators. 
In addition, a MI (mutual information) Estimator is designed as a disentanglement constraint to reduce the mutual information
between $x_{id}$ and $x_{age}$.

\subsubsection{Multi-task modeling with FR}\label{subsubsec4.2.3}

Besides face ID, many methods chose to bring in more supervised information while training a FR model.
In this subsection, we introduce multi-task modeling. 

\begin{figure}[htp]
    \centering
    \includegraphics[width=15cm]{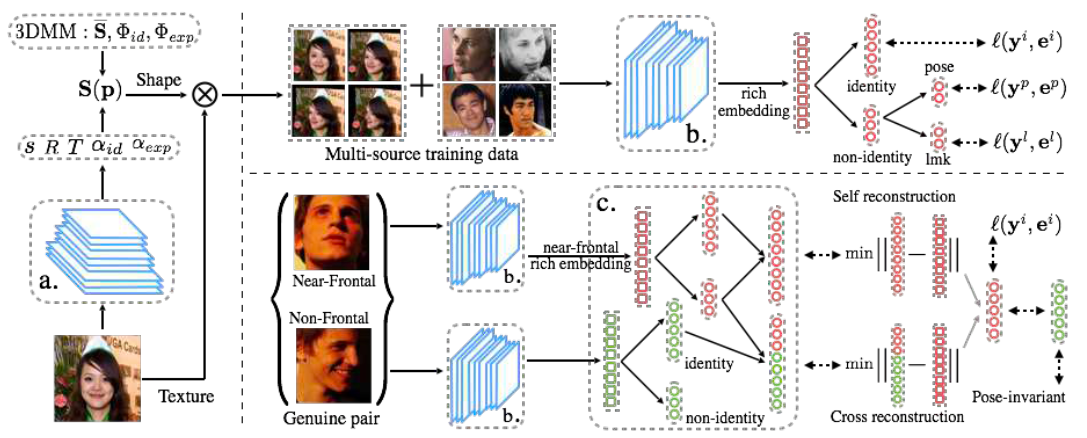}
    \caption{The architectures of R3AN.}
    \label{fig:posei2017}
\end{figure}

Peng \emph{et al.} \cite{peng2017reconstruction} presented a method for learning pose-invariant feature representations.
First, a 3D facial model is applied to synthesize new viewpoints from near-frontal faces. 
Besides ID labels $e^i$, face pose $e^p$ and landmarks $e^l$ are also used as supervisions, and rich embedding is then achieved by jointly learning the identity and non-identity features with extractor $\theta^r$. 
In training, the rich embedding is split into identity, pose and landmark features, which will be fed into different losses, softmax loss for ID estimation, and L2 regression loss for pose and landmark prediction.
Finally,  a genuine pair, a near-frontal face $x_1$ and a non-frontal face $x_2$, is fed into the recognition network $\theta^r$ to obtain the embedding $e^r_1$ and $e^r_2$. The $x_1$ and the $x_2$ share the same identity.
Disentangling based on reconstruction is applied to distill the identity feature from the non-identity one for robust and pose-invariant representation. 
The framework of \cite{peng2017reconstruction} is presented in Fig. \ref{fig:posei2017}

Wang \emph{et al.} \cite{wang2018orthogonal} solved age-invariant FR by adding age prediction task.
To reduce the intra-class discrepancy caused by the aging, Wang \emph{et al.} proposed an approach named Orthogonal Embedding CNNs (OE-Cnns) to learn the age-invariant deep face features.
OE-Cnns first trains a face feature extractor to get the feature $x_i$ of sample $i$. Then $x_i$ is decomposed into two components. 
One is identity-related component $x_{id}$, which will be optimized by identity
classification task by SphereFace \cite{liu2017sphereface}; 
and the other is age-related component $x_{age}$, which is used to estimate age and will be optimized by regression loss formulated as follows:
\begin{equation}
L_{age} = \frac{1}{2M} \sum_{i=1}^{M} \| f(\frac{x_i}{\|x_i\|_2}) - z_i \|^2_2
\end{equation}
where $\|x_i\|_2$ is the length of embedding $x_i$, $z_i$ is the corresponding
age label. $M$ is the batchsize. $f(.)$ is a mapping function aimed to associate $\frac{x_i}{\|x_i\|_2}$ and $z_i$.
While inference, after removing $x_{age}$ from $x$, $x_{id}$ will be obtained that is supposed to be age-invariant.

\begin{figure}[htp]
    \centering
    \includegraphics[width=15cm]{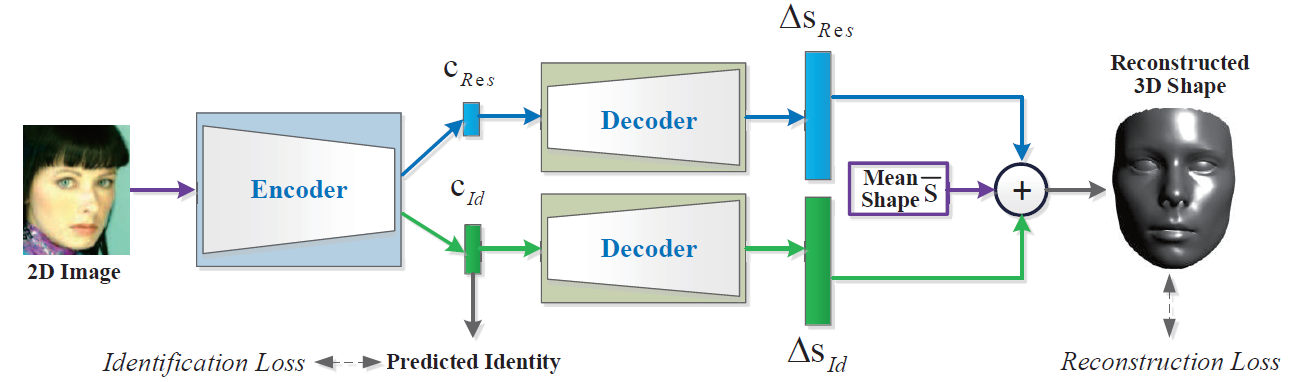}
    \caption{Overview of method \cite{liu2018disentangling} with encoder-decoder based joint learning pipeline for face recognition and 3D shape reconstruction}
    \label{fig:disent2018}
\end{figure}

Liu \emph{et al.} \cite{liu2018disentangling} merged 3D face reconstruction and recognition.
In \cite{liu2018disentangling} each 3D face shape $\boldsymbol s$ is represented by the concatenation of its vertex coordinates
$\boldsymbol s = [x_1,y_1,z_1 , x_2,y_2,z_2 , \dots , x_n,y_n,z_n]^T$,
where $n$ is the number of vertices in the point cloud of the 3D face.
Based on the assumption that 3D face shapes are composed by identity-sensitive and identity-irrelevant parts, 
$\boldsymbol s$ of a subject is rewritten as:
\begin{equation}
\boldsymbol s = \overline{ \boldsymbol s } + \Delta  \boldsymbol s_{id} + \Delta  \boldsymbol s_{res} 
\end{equation}
where $\overline{ \boldsymbol s }$ is the mean 3D face shape
, $\Delta  \boldsymbol s_{id}$ is the
identity-sensitive difference between $\boldsymbol s$ and $\overline{ \boldsymbol s }$, and $\Delta  \boldsymbol s_{res} $
denotes the residual difference.
A encoder is built to extract the face feature of a 2D image, which will be divided into two parts: $\boldsymbol c_{id}$ and $\boldsymbol c_{res}$.
$\boldsymbol c_{id}$ is the latent representation employed as features for face recognition 
and $\boldsymbol c_{res}$ is the representation face shape.
$\boldsymbol c_{id}$ and $\boldsymbol c_{res}$ are further input into two decoders to generate $\Delta  \boldsymbol s_{id}$ and $\Delta \boldsymbol s_{res}$.
Finally, identification loss $L_C$ and reconstruction loss $L_{R}$ will be designed to predict face ID and reconstruct 3D face shape. $L_C$ is softmax loss, which directly optimizes $\boldsymbol c_{id}$. 
The overall network of this method can be seen in \ref{fig:disent2018}

Wang \emph{et al.} \cite{wang2020fm2u} proposed a unified
Face Morphological Multi-branch Network (FM2u-Net) to generate face with makeup for makeup-invariant face verification.
FM2u-Net has two parts: FM-Net and AttM-Net.
FM-Net can synthesize realistic makeup face images by transferring specific regions of cosmetics via cycle consistent loss.
Because of the lack of sufficient and diverse makeup/non-makeup training pairs, 
FM-Net uses the sets of original images and facial patches as supervision information, and employ cycle consistent loss \cite{zhu2017unpaired} to guide realistic makeup face generation.
A softmax for ID prediction and a ID-preserving loss are added on FM-Net to constrain the generated face consistent with original face ID.
AttM-Net, consisting of one global and three local (task-driven on two eyes and mouth) branches, can capture the complementary holistic and detailed information of each face part, and focuses on generating makeup-invariant facial representations by fusing features of those parts. In specific, the total loss for AttM-Net is performed on the four part features and the fused one $f_{cls}$:
\begin{equation}
L_{AttM} = \gamma_1 L_{l-eye} + \gamma_2 L_{r-eye} + \gamma_3 L_{mouth} + \gamma_4 L_{global} + \gamma_5 L_{cls}
\end{equation}
where $\gamma_*$ are weights. $L_{l-eye}$, $L_{r-eye}$ and $L_{mouth}$ are softmax loss of left, right eye and mouth patch. $L_{global}$ and $L_{cls}$ are losses based on the whole face feature and fused feature $f_{cls}$, which are also softmax pattern. 

\begin{figure}[htp]
    \centering
    \includegraphics[width=15cm]{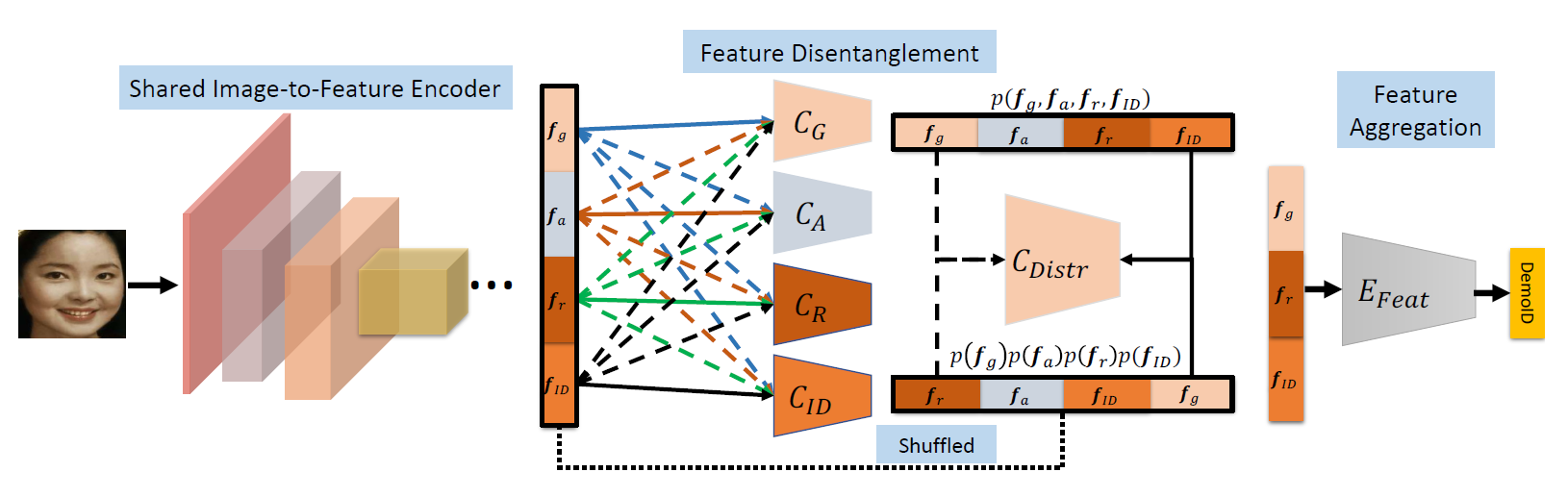}
    \caption{Overview of the DebFace network.}
    \label{fig:attri2020}
\end{figure}

Gong \emph{et al.} \cite{gong2020jointly} proposed a de-biasing adversarial network (DebFace) to jointly learn FR and demographic attribute estimation (gender, age and race).
DebFace network consists of four components: 
the shared image-to-feature encoder $E_{Img}$, 
the four attribute classifiers (including gender $C_G$, age $C_A$, race $C_R$, and identity $C_{ID}$), 
the distribution classifier $C_{Distr}$, 
and the feature aggregation network $E_{Feat}$. 
DebFace first projects an image $x_i$ to its feature representation $E_{Img}(x_i)$ by the encoder $E_{Img}$. 
Then the representation is decoupled into gender, age, race and identity vectors.
Next, each attribute classifier operates the corresponding vector to classify the target attribute
by optimizing parameters of $E_{Img}$ and the respective classifier $C_*$. 
The learning objective $L_{C_{Demo}}$ ($C_{Demo} = \{ C_G,C_A,C_R\}$) is cross entropy loss.
For the identity classification, AM-Softmax \cite{wang2018additive} is adopted in $L_{C_{ID}}$. 
To de-bias all of the representations, adversarial loss $L_{Adv}$ is applied to
the above four classifiers such that each of them will NOT be able to predict
correct labels when operating irrelevant feature vectors. 
To further improve the disentanglement, the mutual
information among the attribute features is reduced by a distribution classifier $C_{Distr}$. 
At last, a factorization objective function $L_{Fact}$
is utilized to minimize the mutual information of the four attribute representations.
Altogether, DebFace endeavors to minimize the joint loss:
\begin{equation}
L =  L_{C_{Demo}} + L_{C_{ID}} +  L_{C_{Distr}} + \lambda_1 L_{Adv} + \lambda_2 L_{Fact}
\end{equation}
where $\lambda_*$ are hyper-parameters determining how much the representation is decomposed and decorrelated in each training iteration.
The pipeline of DebFace is shown in fig \ref{fig:attri2020}

Similar with DebFace, PASS (Protected Attribute Suppression System) \cite{dhar2021pass} proposed by Dhar \emph{et al.} also learned face features which are insensitive to face attribute.
In PASS, a feature $f_{in}$ is firstly extracted by a pretrained model.
Then a generator M inputs $f_{in}$ and generates a new feature $f_{out}$ that are agnostic to face attributes (such as gender and skintone).
A classifier is applied to optimized $f_{out}$ to identify face ID.
In addition, $f_{out}$ is fed to ensemble E, and each of the attribute discriminators in E is used to compute attribute classification. 
In order to force the feature $f_{out}$ insensitive to face attributes, an adversarial loss for model M with respect to all the models in E is calculated by constraining a posterior probability of $\frac{1}{N_{att}}$ for all categories in the attribute, where $N_{att}$ denotes the number of classes in the considered attribute.

\begin{figure}[htp]
    \centering
    \includegraphics[width=15cm]{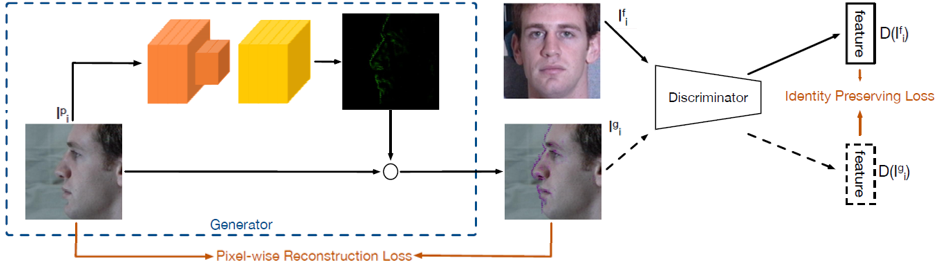}
    \caption{The pipeline of pseudo facial generation framework. }
    \label{fig:pseudo2021}
\end{figure}

Wang \emph{et al.} \cite{wang2021pseudo} relieved the problem of FR with extreme poses by
lightweight pseudo facial generation.
This method can depict the facial contour and make appropriate modifications to preserve the critical identity information without generating any frontal facial image.
The proposed method includes a generator and a discriminator.
Different from traditional GAN, the generator does not use encoder-decoder style. Instead, the lightweight pseudo profile facial generator is designed as a residual network, whose computational costs are much lower.
To preserve the identity consistent information, 
the embedding distances between the generated pseudo ones and their corresponding frontal facial images are minimized, which is similar with the ID-preserving loss in \cite{wang2020fm2u}.
Suppose $D(.)$ denotes a face embedding from a facial
discriminator, the identity preserving loss is formulated as
$L_{id} = \| D(I^g) - D(I^f)\|_2^2$,
where $I^f$ and $I^g$ are real frontal and generated pseudo face, which share a same ID.  
Fig \ref{fig:pseudo2021} depicts the pipeline of this method. 

Besides big poses, the performance of FR algorithm also degenerates under the uncontrolled illumination. To solve it, He et al. \cite{he2022enhancing} introduced 3D face reconstruction into the FR training process as an auxiliary task. The reconstruction was performed based on imaging principles. Four important imaging parameters were learned by two auxiliary networks in training. In the parameters, view matrix and illumination were identity irrelevant and extracted from shallow layers, while depth and albedo were identity relevant and extracted from the intermediate layer. Based on the parameters extraction and reconstruction loss, the FR network can focus on identity relevant features.

\subsection{FR with massive IDs}\label{subsec4.3}
Larger number of IDs (width) in training set can usually achieve a greater FR result. Thus in the real-world FR applications, it is crucial to adopt large scale face datasets in the wild. 
However, the computing and memory cost linearly scales up to the number of classes. 
Therefore, some methods aim at enlarging the throughput of IDs while training.

Larger number of IDs leads to a larger classifier, which may exceed the memory of GPU(s). An intuitive solution is to split the classifier along the class dimension and evenly distribute it to each card. 
On this basis, Partial FC \cite{an2020partial}\cite{an2022killing} further proposed that, for each sample, it is not necessary to use all the negative class centers when calculating logits. A relatively high accuracy can also be achieved by sampling only a part of the negative class centers.

BroadFace \cite{kim2020broadface} thought that only a small number of samples are included in the minibatch, and each parameter update of the minibatch may cause bias, which will fail to converge to the optimal solution.
As a result, BroadFace saved the embedding of the previous iteration $e_i^-$ into a queue, and optimized the classifier and model parameters together with the embedding  of the current iteration $e_i$. Compared with the traditional training method, BroadFace involved more samples in each iteration.
However, as the training progressing, the feature space of the model will gradually drift. There will be a gap between the embedding in the queue and the current feature space. Directly using the embedding of the queue for optimization may result in a decrease in model performance. In this case, the classifier parameters of the previous iteration are used to compensate the embedding in the queue, that is
\begin{equation}
e^*_i=e_i^- + \rho(y) \approx  e_i^- + \frac{\left\|e_i^-\right\|}{\left\| W_{y_i}^-\right\|} (W_y-W_y^-)
\end{equation}
where $y$ is the label of $e_i^-$, $W_y$ and $W_y^-$ are the classifier weights of current and previous iteration.
Compared with traditional methods whose batchsize is usually 256 or 512, BroadFace can adopt thousands of embeddings to optimize the model in each iteration.


Partial FC proposed that only 10\% of the class number can be used to train a high-precision recognition model. 
Inspired by Partial FC, Li \emph{et al.} \cite{li2021dynamic} proposed a method to train FR model with massive IDs, and also improve the performance on long-tail distributed data. 
Li \emph{et al.} tried to discard the FC layer in the training process, and used a queue of size $K$ to store the category center ($K \ll C$, $C$ is the number of IDs in the training set), which is treated as the negative class center. 
And the dynamic class generation method is used to generate the positive class center. 
Then the positive class center is pushed into the queue, and the oldest class centers of the queue are popped out.
This training process has two problems: 
1, there is no guarantee that the positive class center of the current batch is not included in the queue; 2, as training progressing, the feature space will drift, and the previous negative class center cannot match the current feature space. 
For the first problem, the author simply setted the duplicate logits of all positive class centers in the queue to negative infinity, so that the response to softmax is 0. 
For the second problem, the author referred to the idea of MOCO and used the model composed of EMA with feature extractor parameters to generate the positive center.
This article used a queue of size $B+K$ to store the category center of the previous iteration, where $B$ is the batchsize and $K$ is the capacity of the queue. In each iteration, only $B$ elements in the queue are updated. 

In \cite{li2021virtual}, a new layer called Virtual FC layer was proposed to reduce the computational consumption of the classification paradigm in training.
The algorithm splits $N$ training IDs into $M$ groups randomly. 
The identities from group $l$ share the $l$-th column in the projection matrix $W \in \mathbb{R}^{D \times M}$, where $D$ is the dimension of face embedding. 
The $l$-th column of $W$ is called $anchor_l$. 
If the mini-batch contains identities from group $l$, $anchor_l$ is of type $anchor_{corr}$. 
Otherwise, it is of type $anchor_{free}$. The anchor type is adaptive in every training iteration.
Each column in the projection matrix $W$ of the final FC layer indicates the
centroid of a category representation, thus the corresponding
anchors is 
$anchor_{corr,l} = \frac{1}{K} \sum_{i=1}^K f_{i,l}$,
where $f_{i,l}$ the feature of the $i$-th image that belongs to group l, and $K$ is the number of features in group $l$. 
In training, $anchor_l$ will be updated by the above equation if it belongs to $anchor_{corr}$, and it stays the same if it is of type $anchor_{free}$. 
As mentioned above, several IDs are related in group $l$. Therefore, \cite{li2021virtual} further came up with a regrouping strategy to avoid sampling identities that are from the same group into a mini-batch.

Faster Face Classification ($F^{2}C$) \cite{wang2022efficient}, adopted Dynamic Class Pool (DCP) for storing and updating the identities’ features dynamically, which could be regarded as a substitute for the FC layer.

\subsection{Cross domain in FR}\label{subsec4.4}

Generally, when we utilized the FR algorithms, the training and testing data should have similar distributions. 
However, face images from different races, or in different scenes (mobile photo albums, online videos, ID cards) have obvious domain bias. 
Unsatisfactory performance will occur when training and testing domain gap exits, due to the poor generalization ability of neural networks to handle unseen data in practice. 
Therefore, domain adaptation is proposed to solve this problem.
In this subsection, we first proposed general domain adaptation methods in FR.
And then we list some FR methods with uncommon training images.

Model-agnostic meta-learning (MAML) \cite{finn2017model} is one of the most representative methods in domain adaptation. It aims to learn a good weights initialization of a model, so that it can get good results on new tasks with a small amount of data and training iterations. 
The input of the algorithm is a series of tasks with its corresponding support set and query set.
The output is a pre-trained model. 
A special optimization process was proposed in this work, that is, for each iteration, the initial parameter of the model is $\theta$, for each task $\mathcal T_i$, a gradient descent is performed on the support set with a larger learning rate to get special parameters $\theta'_i$ for each task's model, and then use the model under parameter $\theta_i'$ on the query set to find the gradient of each task to $\theta$ , and then perform gradient descent for $\theta$ with a smaller learning rate.
The whole process is divided into two gradient calculations. 
The first time is to calculate the gradient that can improve performance for each task. 
The second time is to update the model parameters under the guidance of the first gradient descent result.

Many researchers brought MAML in FR algorithms to alleviate cross domain problem. 
Guo \emph{et al.} \cite{guo2020learning} proposed meta face recognition (MFR) to solve the domain adaptation problem in FR through meta-learning.
In each iteration of training, only one of N domains in the training set will be chosen as meta-test data, corresponding to the query set of MAML; and the rest of (N-1) domains in the training data are used as meta-train data, corresponding to the support set of MAML. All these data constitutes a meta-batch.
Meta-test data is utilized to simulate the domain shift phenomenon in the application scenario. 
Then, the hard-pair attention loss $\mathcal L_{hp}$, the soft-classification loss $\mathcal L_{cls}$, and the domain alignment loss $\mathcal L_{da}$ are proposed. The $\mathcal L_{hp}$ optimizes hard positive and negative pairs by shrinking the Euclidean distance in hard positive pairs and pushing hard negative pairs away. 
The $\mathcal L_{cls}$ is for face ID classification, which is modified from cross-entropy loss. 
To perform domain alignment, the $\mathcal L_{da}$ is designed to make the mean embeddings of multiple meta-train domains close to each other. 
In the optimization process, this article also followed a similar method to MAML: for model parameters $\theta$, MFR first uses meta-train data to update the model parameters to obtain the $\theta'$ by $\mathcal L_{hp}$, $\mathcal L_{cls}$, and $\mathcal L_{da}$. 
Then MFR uses the updated $\theta'$ to calculate meta-test loss on $\mathcal L_{hp}$, $\mathcal L_{cls}$. After that, the gradient is utilized to further update the model parameters. 
The difference is that when updating the $\theta$, not only the meta-test loss is used to calculate the gradient, but also the meta-train loss is used, which is for the balance of meta-train and meta-test. 

\begin{figure}[htp]
    \centering
    \includegraphics[width=15cm]{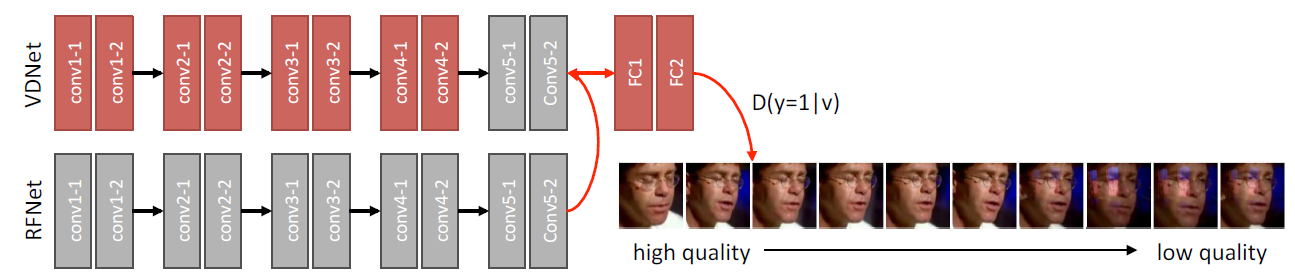}
    \caption{The network architecture for RFNet and VDNet.}
    \label{fig:domain2017}
\end{figure}

Faraki \emph{et al.} \cite{faraki2021cross} pointed out that the domain alignment loss $\mathcal L_{da}$ in MFR may lead to a decrease in model performance. The reason is that while the mean value of the domain is pulled closer, samples belonging to different IDs may be pulled closer, resulting in a decrease in accuracy. 
Therefore, they proposed cross domain triplet loss based on triplet loss, which is shown as follows:
\begin{equation}
\begin{aligned}
& l_{cdt}(^i\mathbb T, ^j\mathbb T; \theta_r) = \\
& \frac{1}{B} \sum_{b=1}^B  \left[   \frac{1}{HW} \sum_{h=1}^H\sum_{w=1}^W  d_{j\Sigma^+}^2([f_r(^ia_b)]_{h,w},[f_r(^ip_b)]_{h,w})- \right. \\
& \left.  \frac{1}{HW} \sum_{h=1}^H\sum_{w=1}^W d_{j\Sigma^-}^2([f_r(^ia_b)]_{h,w},[f_r(^in_b)]_{h,w})+ \tau   \right]_{+} \\
\end{aligned}
\end{equation}
where $^j\mathbb T$ represents the triplets of domain $j$, $\theta_r$ represents the parameters of the representation model, $f_r$ represents the representation model, the output of $f_r$ is a tensor with dimensions $(H, W, D)$. The $(a, p, n)$ denotes the anchor, positive and negative sample in triplets.  $d^2_{j \Sigma^+}$ represents the Mahalanobis distance of the covariance matrix based on the distance of the positive pairs in domain $j$.
The cross domain triplet loss can align the distribution between different domains.
The proposed optimization process is also similar to MAML and MFR. The training data is divided into meta-train and meta-test. 
The initial parameter is $\theta$. Then in each iteration, meta-train data is firstly optimized by CosFace \cite{wang2018cosface} and triplet loss \cite{schroff2015facenet}, to obtain the updated parameters $\theta'$. The $\theta'$ is used to calculate the meta-test loss on the meta-test. Finally the meta-test loss is calculated to update the model parameter. The optimization on meta-test data uses cross entropy loss, triplet loss, and cross domain triplet loss. 
Sohn \emph{et al.} \cite{sohn2017unsupervised} proposed an unsupervised domain adaptation method for video FR using large-scale unlabeled videos and labeled still images. 
It designed a reference net named RFNet with supervised images as a reference, and a video net named VDNet based on the output of RFNet with labeled and synthesized still images and unlabeled videos. 
The network architecture can be seen in Fig. \ref{fig:domain2017}. 
Based on RFNet and VDNet, four losses are proposed, namely the feature match loss $\mathcal L_{FM}$, the feature restoration loss $\mathcal L_{FR}$, the image classification loss $\mathcal L_{IC}$ and the adversarial loss $\mathcal L_{Adv}$.
The $\mathcal L_{FM}$ is for shortening the gap between the embeddings extracted from RFNet and VDNet on labeled images $\mathcal I$, whose formulation is as follows:
\begin{equation}
    \mathcal L_{FM} = \frac{1}{ \lvert \mathcal I \rvert} \sum_{x \in \mathcal I}  \left\|  \phi(x) - \psi(x) \right\|_2^2
\end{equation}
where $\phi(x)$ and $\psi(x)$ are the output features of VDNet and RFNet for a same input image $x$.
Then the authors set a feature restoration constrain on VDNet. They add a series of data argumentation operations $B(\cdot)$ to image set $\mathcal I$, including linear motion blur, scale variation, JPEG compression, etc., 
and use the $\mathcal L_{FR}$ to optimize VDNet to “restore” the original RFNet representation of an image without data
augmentation. The $\mathcal L_{FR}$ is designed as follows
\begin{equation}
{\mathcal L_{FR}} = \frac{1}{\lvert \mathcal I \rvert} 
\sum_{x\in\mathcal I}\mathbb E_{B(\cdot)} \left[\left\| \phi(B(x)) - \psi(x)\right\|_2^2\right]
\end{equation}
where $\mathbb E_{B(\cdot)}$ is the expectation over the distribution of the image transformation kernel $B(\cdot)$.
The $\mathcal{L}_{IC}$ is designed in metric learning formulation to reduce the gap between features of images from the RFNet and their related synthesis from the VDNet. Given $N$ pairs of examples from $N$ different classes
$\{(x_i, x_i^+)\}^N_{i=1}$, the $\mathcal{L}_{IC}$ is shown as follows:
\begin{equation}
    \mathcal{L}_{IC}=-\frac{1}{N} \sum_{i=1}^{N} \log \frac{\exp \left(\phi\left(B_{i}\left(x_{i}^{+}\right)\right)^{\top} \psi\left(x_{i}\right)\right)}{\sum_{n=1}^{N} \exp \left(\phi\left(B_{i}\left(x_{i}^{+}\right)\right)^{\top} \psi\left(x_{n}\right)\right)}
\end{equation}
The $\mathcal L_{Adv}$ is adopted to fool the discriminator and refine the gaps between the domain of image (y=1), synthesized images (y=2), and videos (y=3), which is shown as follows:
\begin{equation}
    \mathcal L_{Adv} = -\mathbb E_{x\in\mathcal B(\mathcal I) \cup \mathcal V} [\log\mathcal D(y=1  \lvert \phi(x))]
\end{equation}
where $\mathcal V$ represents video images domain.


Some methods try to solve FR problems of uncommon images (such as NIR images or radial lens distortion images), which has huge domain gap with conventional RGB images in mainstream FR datasets.

\begin{figure}[htp]
    \centering
    \includegraphics[width=\textwidth]{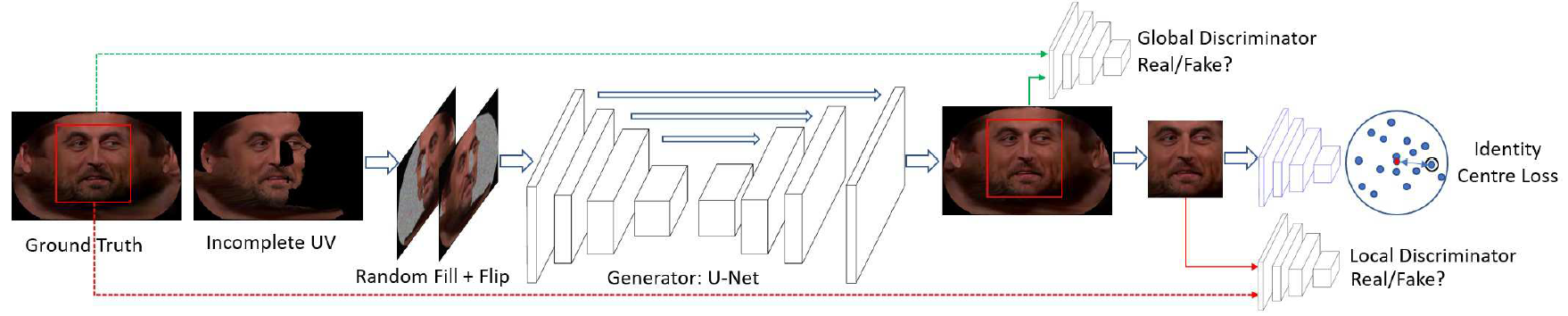}
    \caption{The pipeline of the UV-GAN. \cite{deng2018uv}}
    \label{fig:UVGAN}
\end{figure}

To solve pose-invariant FR problem, Sengupta \emph{et al.} \cite{deng2018uv} adopted facial UV map in their algorithm.
An adversarial generative network named UV-GAN was proposed to generate the completed UV from the incompleted UV that came from the 3D Morphable Model \cite{booth20173d}. 
Then face images with different poses can be synthesized from the UV, which will be utilized to get pose-invariant face embeddings. The UV-GAN is composed of the generator, the global and local discriminator, and the pre-trained identity classification network (Fig. \ref{fig:UVGAN}).
Two discriminators ensure that the generated images are consistent with their surrounding contexts with vivid details. 
The pre-trained identity classification network is an identity preserving module and fixed during the training process. 

In addition to pose variation, two-dimensional (2D) FR is also faced with other challenges such as illumination, scale and makeup. 
One solution to these problems is the three-dimensional (3D) FR. 
Gilani et al. \cite{gilani2018learning} proposed the Deep 3D FR Network (FR3DNet) to solve the shortage of 3D face data. Inspired by \cite{gilani2017dense}, they presented a synthesized method. 
The synthesized method firstly conducted dense correspondence. 
After obtaining synthesized images, commercial software was utilized to synthesis varying facial shapes, ethnicities, and expressions.
The architecture of the FR3DNet was similar to \cite{parkhi2015deep} and it adopted a large kernel size to process the point cloud information. 
To feed the 3D point cloud data into the network, the authors changed the data to three channel images. 
The first channel was the depth information obtained from the $gridfit$ algorithm \cite{d2005surface}. 
The second and third channels were generated based on the azimuth angles and the elevation angle in spherical coordinates.

\begin{figure}[htp]
    \centering
    \includegraphics[width=0.8\textwidth]{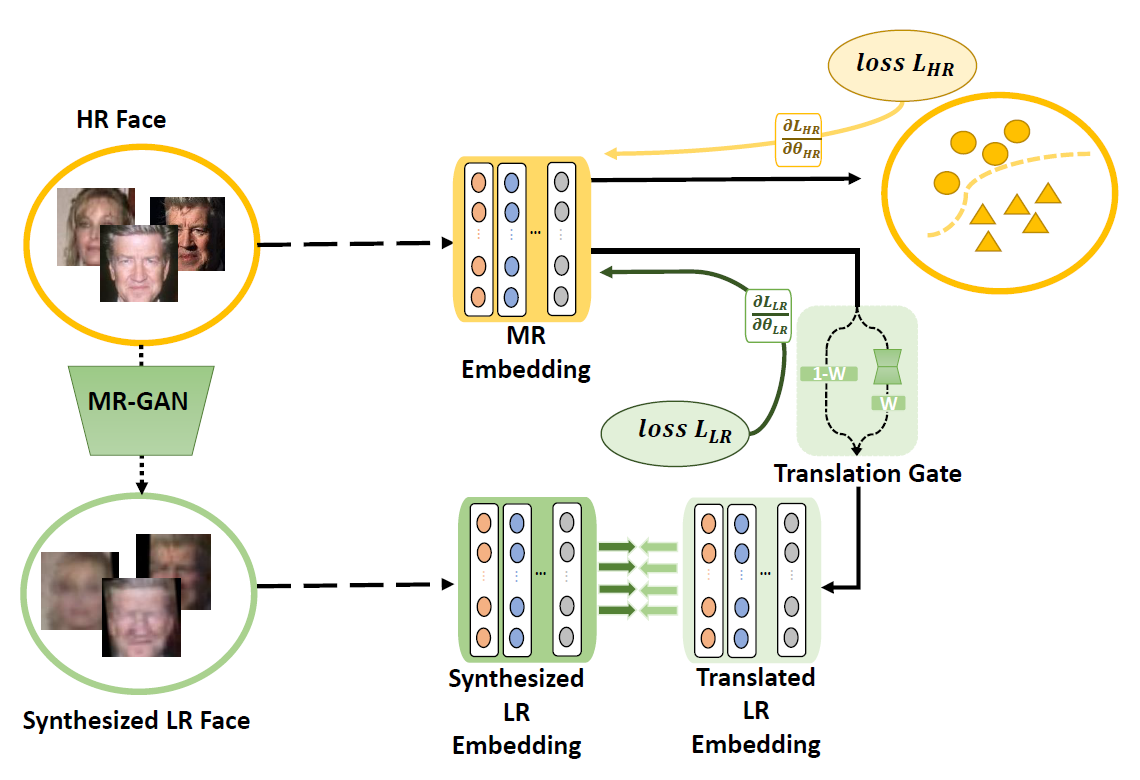}
    \caption{The pipeline of the RAN \cite{fang2020generate}.}
    \label{fig:RAN}
\end{figure}

FR in surveillance scenario is another challenge domain, because most of faces in this situation are poor quality images with low-resolution (LR). While most images in academic training sets are high-resolution (HR).
Fang et al. \cite{fang2020generate} built an resolution adaption network (RAN) (Fig. \ref{fig:RAN}) to alleviate low-resolution problem.
It contained three steps. 
First, the multi-resolution generative adversarial network was proposed to generate LR images. 
It input three images ($x_{r1}$, $x_{r2}$, and $x_{r3}$) with different resolutions, which were processed by the parallel sub-networks \cite{sun2019deep}. 
In the second step, HR and LR FR model were trained separately to obtain face representations. 
At last, a feature adaption network was designed to allow the model have high recognition ability in both HR and LR domains. 
The loss $L_{HR}$ using ArcFace \cite{deng2019arcface} was applied to the HR face representations($f_{HR}$) to directly make the model applicable to HR faces. 
At the same time, the $f_{HR}$ was input into a novel translation gate to minimize the gap between the HR and LR domains. 
The output of the translation gate($T_{LR}(f_{HR})$) preserved LR information contained in the $f_{HR}$ and the preserving result was monitored by a discriminator, which was used to distinguish $T_{LR}(f_{HR})$ from the synthesized LR embedding. 
The final LR representations $f_{LR}^{Translate}$ combined $T_{LR}(f_{HR})$ and $f_{HR}$. 
$L_1$ loss and KL loss were applied on $f_{LR}^{Translate}$ and the real LR image representations to further ensure the quality of $f_{LR}^{Translate}$.

Surveillance cameras often capture near infrared (NIR) images in low-light environments. 
FR systems trained by the visible light spectrum (VIS) face images can not work effectively in this situation, due to the domain gap.
Lezama \emph{et al.} \cite{lezama2017not} proposed a NIR-VIS FR system, which can perform a cross-spectral FR and match NIR to VIS face images.
NIR-VIS contains two main components: cross-spectral hallucination and low-rank embedding.
The cross-spectral hallucination learns a NIR-VIS mapping on a patch-to-patch basis. 
After hallucination, the CNN output of luminance channel is blended with the original NIR image to avoid losing the information contained in the NIR image. 
The blending formula is shown as follows:
\begin{equation}
Y = \hat{Y} - \alpha\cdot G_{\sigma}^{2}*(N_{ir}-\hat{Y})
\end{equation}
where $Y$ is the output after combining the hallucinated result with the NIR image, $\hat{Y}$ is the hallucinated result, $N_{ir}$ is the NIR image, $G_{\sigma}$ is a Gaussian filter with $\sigma = 1$, and $*$ denotes convolution.
The second component is low-rank embedding, which performs a low-rank transform \cite{qiu2015learning} to embed the output of the VIS model. 

\begin{figure}[htp]
    \centering
    \includegraphics[width=15cm]{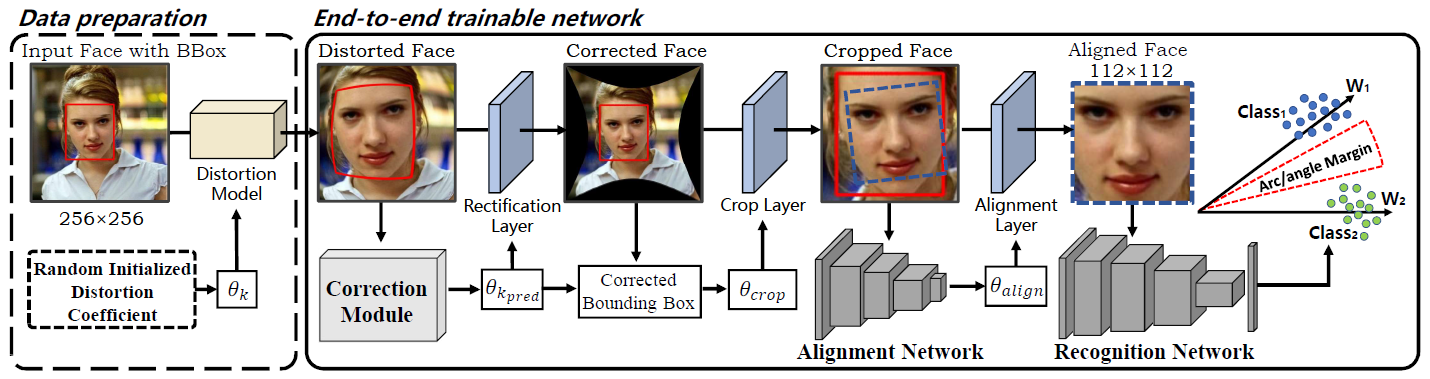}
    \caption{The pipeline of the RDCFace.}
    \label{fig:rdc2020}
\end{figure}

To alleviate the effects of radial lens distortion on face image, 
a distortion-invariant FR method called RDCFace \cite{zhao2020rdcface} was proposed for wide-angle cameras of surveillance and safeguard systems.
Inspired by STN \cite{jaderberg2015spatial}, RDCFace can learn rectification, alignment parameters and face embedding in a end-to-end way, therefore it did not require supervision of facial landmarks and distortion parameters.
The pipeline of RDCFace is shown in Fig. \ref{fig:rdc2020}.
In training, the data preparation module generates radial lens distortion on common face images with random parameters. 
Then the cascaded network of RDCFace sequentially rectifies distortion by correction network, aligns faces by alignment network, and extracts features by recognition network for FR.
The correction network predicts the distortion coefficient $k$ to rectify the radial
distortion based on the inverse transformation of the division model:
\begin{equation}
r_d = \frac{1- \sqrt{1-4kr_u^2}}{2kr_u}
\end{equation}
where $r_u$ and $r_d$ represent the Euclidean distance from an arbitrary pixel to the image center of original and distorted
images.
Taking a distorted face image $I^d$ as input,
a well-trained correction network $f_{correct}$ should predict the distortion
parameter $k$ accurately. 
Then the rectification layer $L_R$ then use parameter $k$ to eliminate the distortion and generates corrected
image $I^c$.
After that, $I^c$ is sent into $f_{correct}$ again. This time, the output parameter is
expected to be zero since the input $I^c$ is expected to be corrected perfectly without distortion.
Therefore, the re-correction loss is designed to 
1, encourage the correction network to better eliminate the distortion and 
2, suppress the excessive deformation on the distortion free image. 
\begin{equation}
L_{correct} = E\| f_{correct}(I^c)\|^2_2 = E\|f_{correct}(L_R( f_{correct}(I^d), I^d))\|^2_2
\end{equation}
The alignment networks predicts projective transformation parameters, which is similar with STN.
The final training loss in RDCFace is sum of ArcFace and $L_{correct}$.

Some previous works utilized synthetic faces to train face recognition model to solve the problems caused by real data, such as label noise, unbalanced data distribution, and privacy. While, the model trained on synthetic images can not perform as well as the model trained on real images. SynFace \cite{qiu2021synface} found one of the reason is the limitation of the intra-class variations within the synthetic data, therefore it proposed identity mixup (IM) for the input parameters of the generator DiscoFace-GAN \cite{deng2020disentangled}. The other technique named domain mixup, by adding a small quantity of real data into the training, the performance of model was improved greatly.

\subsection{FR pipeline acceleration}\label{subsec4.5}

Liu \emph{et al.} \cite{liu2017learning} proposed a novel technique named network slimming. The core idea is that each scaling factor in the batch normalization (BN) layer can indicate the importance of the corresponding channel. The smaller the scaling factor is, the more insignificant this channel will be. Based on this idea, the author performed L1 regularization on the scaling factor in the BN layer during training to make it sparse, and then cut out unimportant channels according to the size of the scaling factor.

Then, by experiments, Liu \emph{et al.} \cite{liu2018rethinking} further proved that the structure of the network is more important than network weights and pruning can be regarded as the process of network architecture search. 
After the pruning, retraining the network with randomly initialized parameters allows the network achieve a better result.

In face identification, we need to match the query face feature with all features in gallery. In order to accelerate this matching process, different searching technologies were adopted.
The K-D tree\cite{bentley1975multidimensional} is a common tree-based search algorithm.
The K-D tree continuously divides the whole feature space into two parts by a hyperplane to form a binary tree structure. 
Each point in the feature space corresponds to a node in the tree, and then searches the tree to find the nearest point.
However, the K-D tree degenerates into linear search in higher dimensions due to very sparse data distribution in high dimensional space. 
The efficiency of feature space segmentation is very low. Based on the segmentation, the search precision declines. 

Vector quantization is a process of encoding a feature space with a limited set of points.
After the encoding, the resulting set of limited points is called a codebook, and each point in the codebook can represent a region in the feature space.
Vector quantization can speed up the calculation of the distance between vectors. 
When m vectors in a feature space are encoded by a codebook of size N (m$>>$N), 
the number of calculations for comparison is reduced from m times to N times.
Product quantization\cite{jegou2010product} is an ANN algorithm based on vector quantization. Vectors in the feature space are first divided into m sub-vectors, and m groups of codebooks are used for quantization. Each codebook has a size of k. Then, the author proposed a novel method combining an inverted file system with the asymmetric distance computation (IVFADC). 

The core idea of graph-based search algorithm is to build points in the feature space into a graph, and start from a point $node_{curr}$ randomly during each retrieval. 
Then it calculate the distance between the neighbor of the point and the vector to be queried,
and select the neighbor with the smallest distance as the next retrieval point $node_{curr}$. 
Until the distance between $node_{curr}$ and query is less than the distance between any of $node_{curr}$'s neighbors and query. 
The $node_{curr}$ is considered as the closest point to query in the graph.

Navigable small world (NSW) algorithm \cite{malkov2014approximate} is a graph-based search algorithm. Delauenian triangulation has a high time complexity in constructing graph, so NSW algorithm does not adopt this method. During the graph construction process of NSW algorithm, each point is inserted in sequence, and after insertion, neighbors are found and joined in the current graph. In a graph constructed this way, there are two kinds of edges: short-range links to approximate the delaunay graph, and long-range links to reduce the number of searches on a logarithmic scale.

During the insertion point process, the early short-range links may become long-range links at a later stage, and these long-range links make the whole graph navigable Small World.

The core idea of the locality sensitive hashing search algorithm is to construct a hash function $h$ so that $h(p)=h(q)$ has a high probability when vectors $p$ and $q$ are close enough.
During search, locality sensitive hashing is performed on all features in the data set to obtain hash values. Then, only vectors with hash values similar to query vectors need to be compared to reduce computation.

Knowledge distillation is an effective tool to compress large pre-trained CNNs into models applicable to mobile and embedded devices, which is also a major method for model acceleration.
Wang \emph{et al.} \cite{wang2020exclusivity} put forward a knowledge distillation method specially for FR, which aims to improve the capability of the target FR student network.
They first reshape all filters from a convolutional layer $W$ from $\mathbb{R}^{N \times M \times K_1 \times K_2}$ to $\mathbb{R}^{N \times D}$, where $N$ and $M$ are numbers of filters and input channels; $K_1$ and $K_2$ are the spatial height and width of filters; $D=M \times K_1 \times K_2$.
Then they define weight exclusivity for weights $W$ as :
\begin{equation}
L_{WE}(W) = \sum_{1 \leq j \neq i \leq N} \sum_{k=1}^D \lvert w_i(k) \rvert \cdot \lvert w_j(k) \rvert
\end{equation}
where $w_i \in \mathbb{R}^{1 \times D}$ is a filter in $W$ with index $i$.
It can be seen that, $L_{WE}(W)$ encourages each of two filter vectors in $W$ to be as diverse as possible. Thus applying weight exclusivity on student network will force to enlarge its capability.
Finally, the proposed exclusivity-consistency regularized knowledge distillation becomes:
\begin{equation}
L = L_{HFC} + \lambda_1 L_{WD} + \lambda_2 L_{WE}
\end{equation}
where $L_{HFC}$ is L2 based hardness-aware feature consistency loss, which encourages the face features from teacher and student as similar as possible. $L_{WD}$ is L2 weight decay.

The knowledge distillation methods mostly regarded the relationship between samples as knowledge to force the student model learn the correlations rather than embedding features from the teacher model. Huang et al. \cite{huang2022evaluation} found that allowing the student study all relationships was inflexible and proposed an evaluation-oriented technique. The relationships that obtained different evaluation result from teacher and student models were defined as crucial relationships. Through a novel rank-based loss function, the student model can focus on these crucial relationships in training.

\subsection{Closed-set Training}\label{subsec4.6}
\begin{figure}[htp]
    \centering
    \includegraphics[width=15cm]{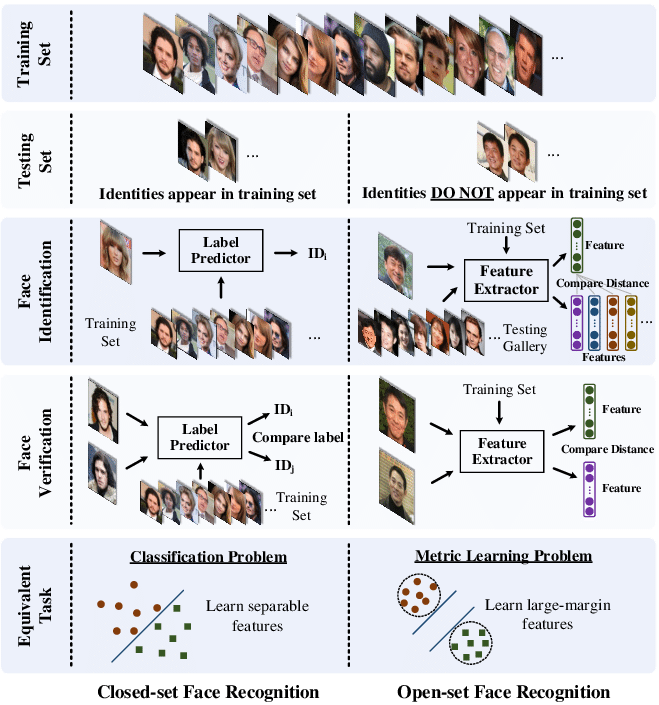}
    \caption{Closed-set and open-set FR systems.}
    \label{fig:closeset}
\end{figure}

For some face verification and identification projects in industry, FR problem can be treated as a closed-set classification problem. 
In the case of ``face identification of politicians in news'' or ``face identification of Chinese entertainment stars'', the business side which applies the FR requirements usually has a list of target people IDs. 
In this situation, the gallery of this identification work is given, and we can train with those IDs to achieve a higher accuracy.
As a result, FR model training becomes a closed-set problem.
In summary, based on whether all testing identities are predefined in the training set, FR systems can
be further categorized into closed-set systems and open-set systems, as illustrated in Fig. \ref{fig:closeset}.
A close-set FR task is equivalent to a multi-class classification problem by using the standard softmax loss function in the training phase \cite{sun2015deep,sun2014deep,taigman2014deepface}.

Tong \emph{et al.} \cite{tong2021facesec} proposed a framework for fine-grained robustness evaluation of both closed-set and open-set FR systems. Experimental results shows that, open-set FR systems are more vulnerable than closed-set systems under different types of attacks (digital attack, pixel-level physically realizable attack, and grid-level physically realizable attack).
As a result, we can conclude that, closed-set FR problem is easier than open-set FR.
Thus we can adopt more classification with delicate design to acquire a better performance in FR business. 
Techniques such as fine-grained recognition, attention based classification, etc. can be employed in closed-set FR.

\subsection{Mask face recognition}\label{subsec4.7}
Face recognition has achieved remarkable progress in the past few years. However, when applying those face recognition models to unconstrained scenarios, face recognition performance drops sharply, particularly when faces are occluded. The COVID-19 pandemic makes people have to  wear masks on daily trips, which makes the face recognition performance of occlusion need improvement. Current methods for occluded face recognition are usually the variants of two sets of approaches, one is recovering occluded facial parts \cite{zhao2017robust,wright2008robust,deng2011graph,zhang2017demeshnet}, the other is removing features that
are corrupted by occlusions \cite{yuan2022msml,feng2021towards,shao2021biased}. The pioneering works usually remove features that are corrupted.

\begin{figure}[htp]
    \centering
    \includegraphics[width=15cm]{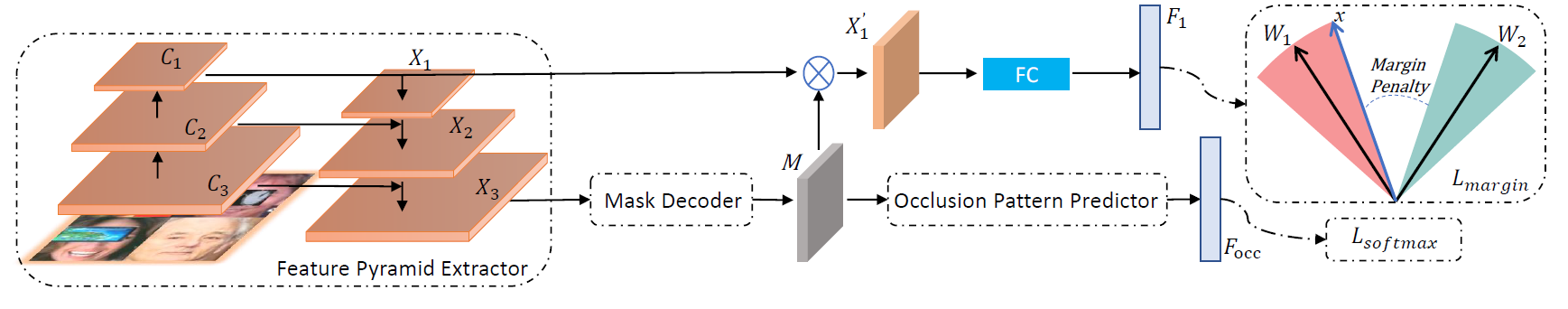}
    \caption{The architectures of FROM.}
    \label{fig:qiu2021end2end}
\end{figure}

FROM\cite{qiu2021end2end} proposed  Mask Decoder and Occlusion Pattern Predictor networks to predict the occlusion patterns. The structure of FROM is shown in Fig.
\ref{fig:qiu2021end2end}.
The structure first took a mini-batch images as input, through a Feature Pyramid Extractor got three different scale feature maps(including $X_{1}$,$X_{2}$,$X_{3}$). Then $X_{3}$ was used to decode the Mask, which contains the occlusion's location information. Mask applied to $X_{1}$ to mask out the corrupted feature elements and get the pure feature $X_{1}'$ for the final recognition. Finally, Occlusion Pattern Predictor predicted occlusion patterns as the extra supervision. The whole network was trained end-to-end.

In order to encourage the network to recognize diverse occluded face, random occlusion was dynamically generated by sunglasses, scarf, face mask, hand, eye mask, eyeglasses etc. The overall loss was a combination of the face recognition loss and the occlusion pattern prediction loss. Mathematically, they defined total loss as follows:
$$  L_{total} = L_{margin} + \lambda L_{pred} $$

$L_{margin}$ is the cosFace loss, $L_{pred}$ is MSE loss or Cross entropy loss.

\begin{figure}[htp]
    \centering
    \includegraphics[width=15cm]{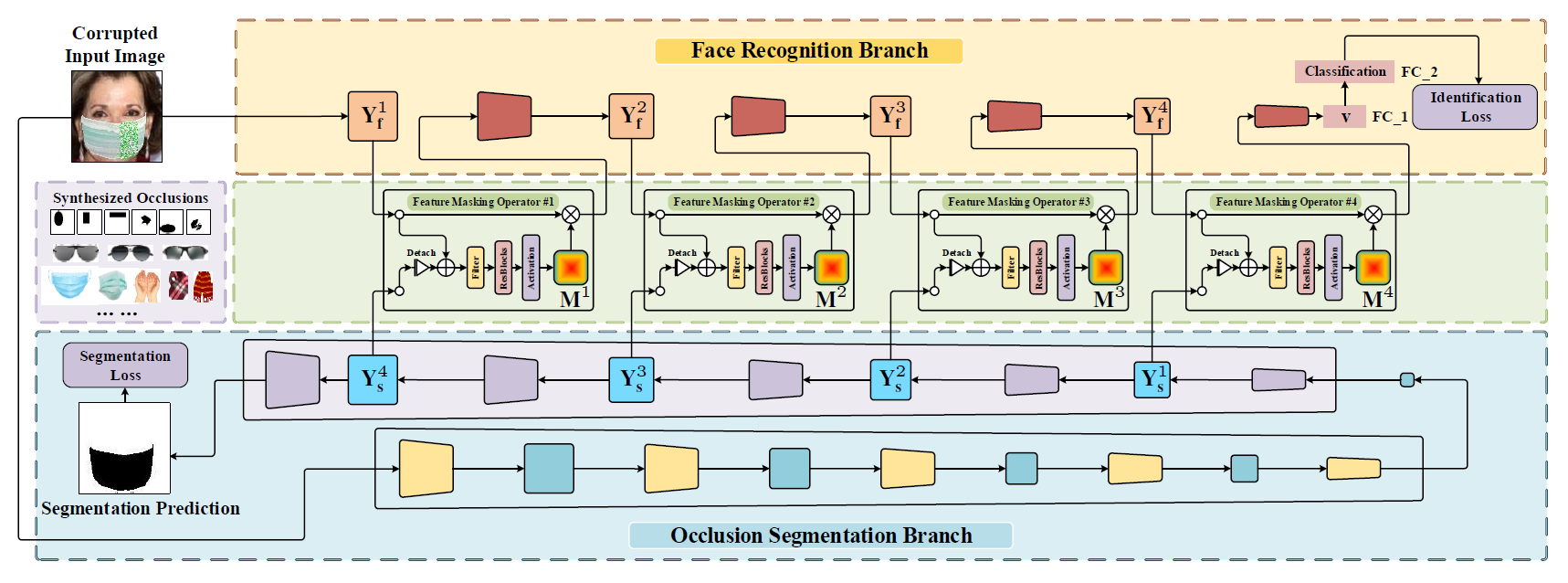}
    \caption{The architectures of MSML.}
    \label{fig:yuan2022msml}
\end{figure}

Different from FROM, \cite{yuan2022msml} adopted a multi-scale segmentation based mask learning (MSML) face recognition network, which alleviate the different scales of occlusion information and purify different scales of face information. The proposed MSML consisted of a face recognition branch (FRB), an occlusion segmentation branch (OSB), and hierarchical feature masking (FM) operators, as shown in Fig.
\ref{fig:yuan2022msml}. 

Using scarves and glasses to randomly generate various occlusions at any position of the original images, meanwhile obtained binary segmentation labels. Different scales characteristics of occlusion was get through the OSB branch and the binary segmentation map was generated by the decoder. In the training stage, segmentation map was constrained by the segmentation loss. Occlusion features extracted by OSB and original face features extracted by FB were fused at FM module to get pure face feature. The whole network was trained through a joint optimization. Total loss can be formulated as:
$$  L_{total} = L_{cls} + \lambda L_{occ} $$
$L_{cls}$ is Cross-entropy loss or other SOTA face recognition losses, $L_{occ}$ is a consensus segmentation loss.

\begin{figure}[htp]
    \centering
    \includegraphics[width=15cm]{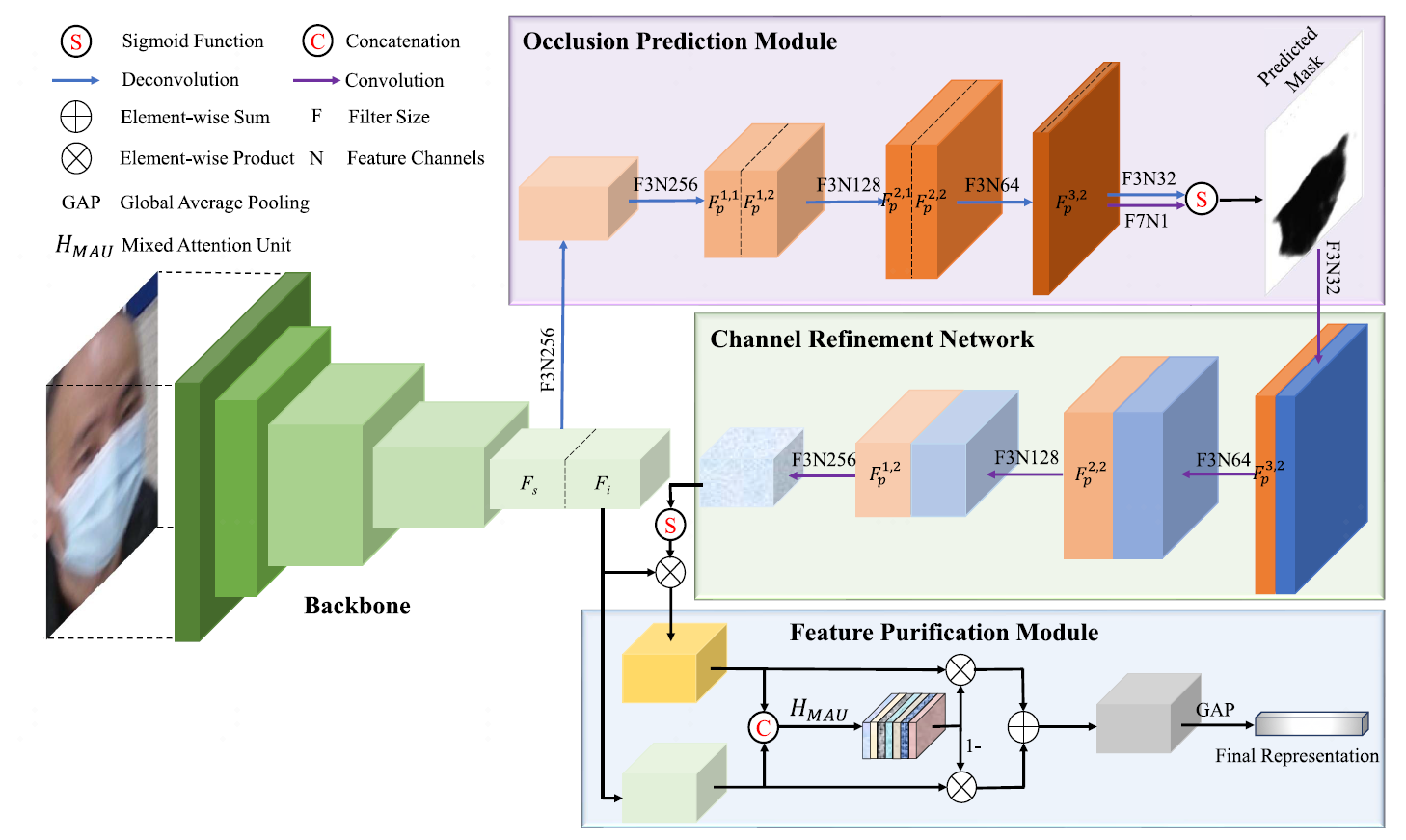}
    \caption{Outline of they proposed Model.}
    \label{fig:huang2022joint}
\end{figure}

Similar to MSML, \cite{huang2022joint} also integrated segmentation tasks to assist mask face recognition as shown in Fig.
\ref{fig:huang2022joint}.
X denoted a masked face which through the whole backbone get a feature map $F=Dconv(X)$, F can entirely or split into two subfeature maps in the channel dimension, one for occlusion prediction(OP) module and the other for identity embedding. The predict mask segmentation result was obtained through OP module. The segmentation task was constrained by segmentation loss. They proposed that the actual 2-D mask should be transformed into a 3-D mask which is more suitable for feature maps. Therefore, they proposed a channel refinement(CR) network for the transformation, the CR network can be defined as follows:
\begin{equation}
F_{r}^k=H(Concat(F_{r}^{k-1},F_{p}^{(4-k),2}))
\end{equation}
where $Concat(.,.)$ refers to the concatenation operation in channel dimension and $k\epsilon \{1, 2, 3\}$. $F_{r}^k$ represents the CR feature of the $k^{th}$ layer. H(·) denotes the downsampling process.
The final goal was to produce discriminative facial features free from occlusion. They multiplied the occlusion mask map with original face features to filter corrupted feature elements for recognition, formulated as:
\begin{equation}
F_{n}= F_{i} \otimes \sigma (F_{r}^3)
\end{equation}
where $F_{i}$ denotes the identity feature extracted by the backbone, $F_{r}^3$ represents the output of the CR network, and $\sigma$ is the sigmoid activation function. They used the loss function from CosFace to optimize the recognition network.

For face OP model, they introduced an occlusion-aware loss to guide the training of mask prediction, which can be formulated as:
\begin{equation}
L_{mask}=\frac{1}{n_{batch}}\sqrt{\frac{\sum_{i=1}^h\sum_{j=1}^w\|M_{i,j}^{gt}-M_{i,j}^{norm} \|^2_2}{hw}}
\end{equation}
where $M_{gt}$ refers to the occlusion mask labels in the training dataset, and $n_{batch}$ is the batch size during the training stage. $M^{norm}$ is the normalized mask map. The final loss function for the end-to-end training was defined as:
\begin{equation}
L =L_{id}+\alpha \times L_{mask}
\end{equation}
where $\alpha$ is a hyperparameter to trade off the classification and segmentation losses.

\begin{figure}[htp]
    \centering
    \includegraphics[width=15cm]{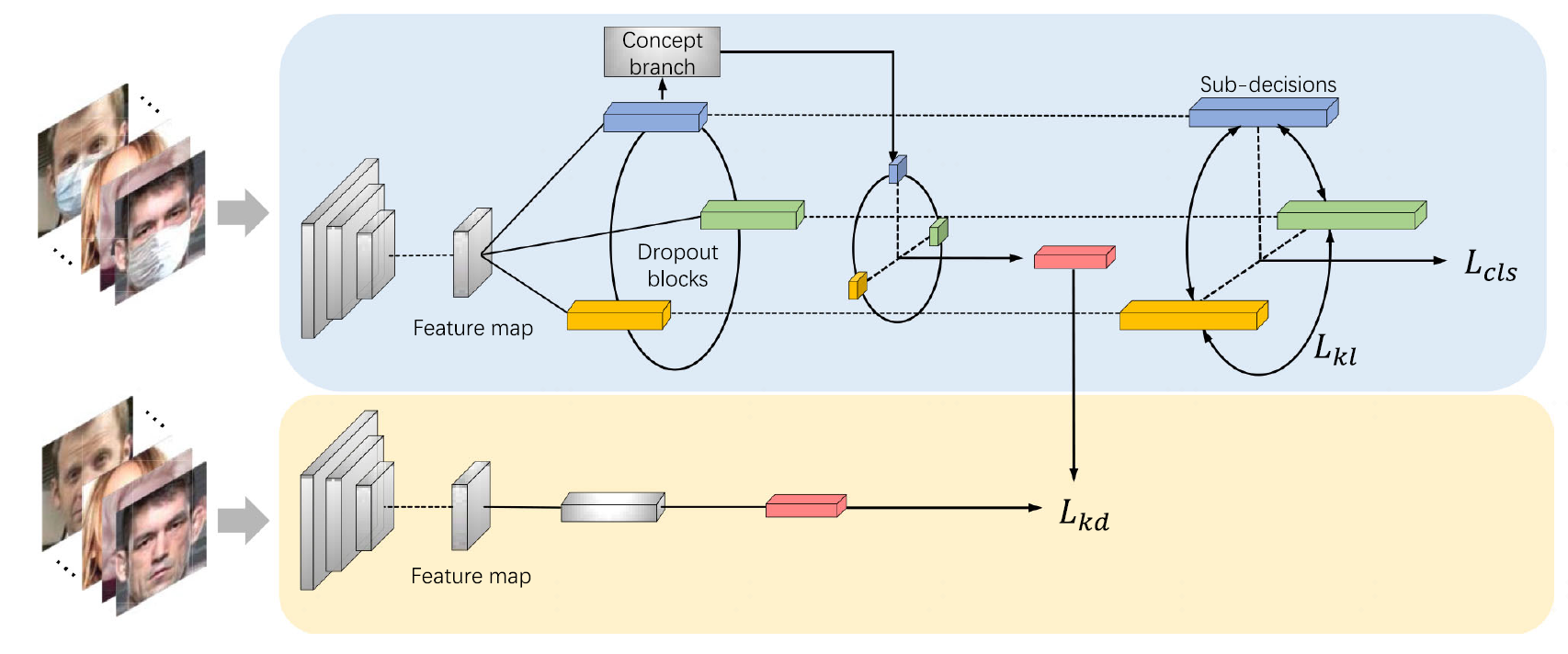}
    \caption{Outline of the Consistent Sub-decision Network.}
    \label{fig:zhao2022consistent}
\end{figure}

Consistent Sub-decision Network \cite{zhao2022consistent} proposed to obtain sub-decisions that correspond to different facial regions and constrain sub-decisions by weighted bidirectional KL divergence to make the network concentrate on the upper faces without occlusion. The whole network is shown as Fig.
\ref{fig:zhao2022consistent}.

The core of occluded face recognition is to lean a masked face embeddings which approximated normal face embeddings. So they proposed different dropout modules to obtain multiple sub-decisions. Every sub-decision uses a concept branch to get a face embeddings information degree $\omega$. \cite{zhao2022consistent} adopted simulation-based methods to generate masked faces from unmasked faces. However, among simulated faces, there were low-quality samples, which leads to ambiguous or absent facial features. The low sub-decision consistency values $\omega$ correspond to low-quality samples in simulated face images. They applied the $\omega$ as weights in the bidirectional KL divergence constraints.
\begin{equation}
L_{kl} = \sum_{i<j}(w_{i} \times KL(s_{i},s_{j})+w_{j} \times KL(s_{j},s_{i}))
\end{equation}
where $s_{i}$,$s_{j}$ indicates different sub-decisions that correspond to diverse face regions of each image.
The normal face contains more discriminative identity information than the masked face, so they adopt knowledge distillation to drive the masked face embeddings towards an approximation of the normal face embeddings to mitigate the information loss, which can be formulated as:
\begin{equation}
f^N =  \mathcal{M}_{teacher}(X^N)
\end{equation}
\begin{equation}
f^M=\sum_{i=1}^{i=3} \omega_{i} \times \mathcal{D}_i( \mathcal{g} (X^M))
\end{equation}
where $X^N$ is normal face, $X^M$ is masked face, $\mathcal{M}_{teacher}$ is the embedding encoder of pretrained model, $\mathcal{g}$ is the feature map generator, $\mathcal{D}_{i}$ is the i-th dropout block,
and $w_{i}$ indicates the output of concept branch. The overall loss function is formulated as follows:
\begin{equation}
L=L_{cls}+\lambda_{1}L_{kl}+\lambda_{2}L_{kd}
\end{equation}
$L_{cls}$ is the CosFace loss, $L_{kl}$ is the KL divergence constraints, $L_{kd}$ is a cosine distance to perform knowledge distillation.

The champion of ICCV 2021-Masked Face Recognition (MFR) Challenge \cite{feng2021towards} proposed some contributions in industrial. They adopt the mask-to-face texture mapping approach and then generated the masked face images with rendering. They constructed a self-learning based cleaning framework which utilized the DBSCAN to realize Inter-ID Cleaning and Intra-ID Cleaning. In the cleaning procedure, they performed self-learning by initializing the i-th model as the (i+1)-th model. Besides they proposed a Balanced Curricular Loss. The loss adaptively adjusted the relative importance of easy and hard samples during different training stages. The Balanced Curricular Loss can be formulated as:
\begin{equation}
\mathcal{L}=-log\frac{n_{y_i}e^{scos(\theta_{y_i}+m)}}{n_{y_i}e^{scos(\theta_{y_i}+m)}+\sum_{j=1,j\neq y_i}^nn_je^{sN(t^{(k),cos\theta_j})}}
\end{equation}
where $cos(\theta_{y_i}+m)$ and $N(t^{(k),cos\theta_j})$ are the cosine similarity function of positive and negative \cite{huang2020curricularface}.

\subsection{Privacy-Preserving FR}\label{subsec4.8}

Face recognition technology has brought many conveniences to people's daily life, such as public security, access control, railway station gate, etc.
However, misuse of this technique brings people hidden worries about personal data security.
Interview and investigation of some TV programs have  called out several well-known brands for illegal face collection without explicit user consent.
Conversely, On the premise of ensuring user data privacy, personal client's private data or public client's data are beneficial to the training of existing models.
Federated Learning (FL) is a technique to address the privacy issue, which can collaboratively optimize the model without sharing the data between clients.

\cite{mcmahan2017communication} is the pioneering work of Federated Learning. They proposed the problem of training on decentralized data from mobile devices as an important research direction.
Experiments demonstrated that the selection of a straightforward and practical algorithm can be applied to this setting.
They also proposed Federated Averaging algorithm, which is the basic framework of Federal Learning.

\cite{mcmahan2017communication} defined the ideal problems for federated learning have three properties.
They also fingered out the federated optimization has several key properties which are differentiate  from a typical distributed optimization.
The unique attributes of federated optimization are the Non-IID, unbalanced similarly, massively distributed, and limited communication.

They proposed  a synchronous update scheme that proceeds in rounds of communication.
There is a fixed set of K clients, every client has a fixed local dataset. And there is T round communication.
The Federated Averaging algorithm is shown as below.

1)At the beginning of i-th round, a random fraction C of clients is selected, and the server
sends the current global mdoel parameters to each of these clients.

2) In every client compute a update through every local data and sent the local model parameters to the server.

3) The server collect and aggregate the local parameters and get a new global parameters.

4) Repeat the above three steps until the end of round T communication.

Their experiments showed diminishing returns for adding more clients beyond a certain point, so they only selected a fraction of clients for efficiency.
Their experiments on MNIST training set demonstrated that common initialization can produces a significant reduction than independent initialization, so they conducted independent initialization on the client.

Federated learning can alleviate the public's concern about privacy data leakage to a certain extent. Because the base network parameters are updated by the client network parameters jointly, the performance of federated learning is not as good as the  ordinary face recognition.
Therefore, the PrivacyFace \cite{meng2022improving} proposed a method to improve the performance of federated face recognition by using differential privacy clustering, desensitizing the local facial feature space information and sharing it among clients, thus greatly improving the performance of federated face recognition model.
PrivacyFace was mainly composed of two components: the first is the Differently Private Local Clustering (DPLC) algorithm based on differential privacy, which desensitizes privacy independent group features from many average face features of the client; the second is the consensus-aware face recognition loss, which makes use of the desensitization group features of each client to promote the global feature space distribution.

\begin{figure}[htp]
    \centering
    \includegraphics[width=15cm]{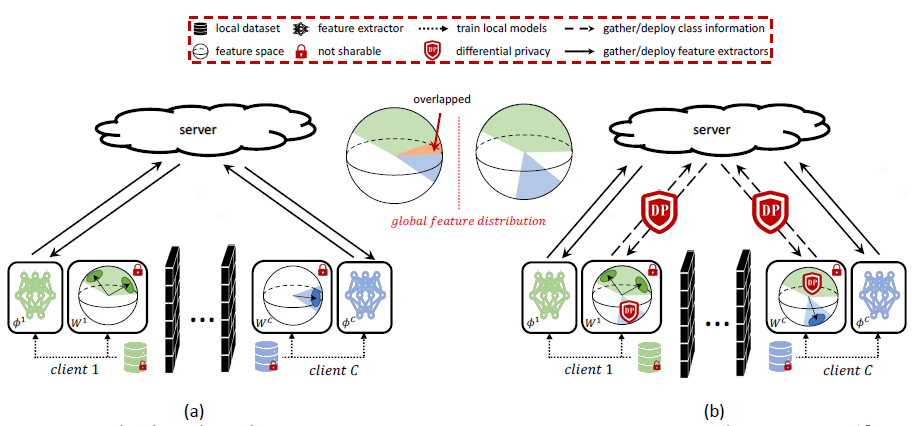}
    \caption{Framework of classical FL and the FL framework with DPLC.}
    \label{fig:dplcpng}
\end{figure}

As show in Fig. \ref{fig:dplcpng}, in the general FL framework, average face space $W^{c}$  cannot be updated during communication, the global optimization of average face space cannot be carried out, resulting in the problem of face space overlap between clients.

The core of their DPLC approximation algorithm was to find the most dense location centered on each sample, and then add noise processing to meet differential privacy.

After the local client has calculated the desensitized cluster center, the server will share the information of each client. Then the local client uses the local data and the group information of other clients to perform the global optimization of the space according to the newly designed consciousness loss.

\begin{equation}
L^{c}(\phi^{c},W^{c})=-\sum_{i=1}^{N_{c}}\frac{e^{u(w_{y_{i}^{c}},f_{i}^{c})}}{e^{u(w_{y_{i}^{c}},f_{i}^{c})}+\sum_{j=1,j\neq y_{i}^{c}}^{n_{c}}e^{v(w_{j},f_{i}^{c})}+\sum_{k=1,k\neq c}^{C}\sum_{l=1}^{Q_{k}}e^{\mu(\hat{p}_{l}^{k},f_{i}^{c},\rho)}}
\end{equation}
There are C clients,where $f_{i}^{c}$ is the feature extracted by i-th model instance in client c.
C owns a training dataset $D_{c}$ with $N_{c}$ images from $n_{c}$ identities.
For client c, its backbone is parameterized by $\phi^{c}$ and the last classifier layer encodes class centers in $W^{c}$=$[w_{1}^{c},w_{2}^{c}...w_{n_{c}}^{c}]$.
$\mu(\hat{p},f,\rho)$= $s \times cos(max(\theta -\rho; 0))$ computes the similarity between f and the cluster centered at $\hat{p}$ with margin $\rho$.
$Q_{c}$ clusters defined by the centers $P^{c}$=$[\hat{p}_{1}...\hat{p}_{Q_{c}}]$ with margin $\rho$, which is generated in each client.

FedGC\cite{niu2022federated} explored a new idea to modify the gradient from the perspective of back propagation, and proposed a regularizer based on softmax, which corrects the gradient of class embeddings by accurately injecting cross client gradient terms. In theory, they proved that FedGC constitutes an effective loss function similar to the standard softmax.
Class embeddings W was upgraded as:

\begin{equation}
W^{t+1}=\tilde{W}^{t+1}-\lambda \epsilon \nabla_{\tilde{W}^{t+1}} Reg(\tilde{W}^{t+1})
\end{equation}

\begin{figure}[htp]
    \centering
    \includegraphics[width=10cm]{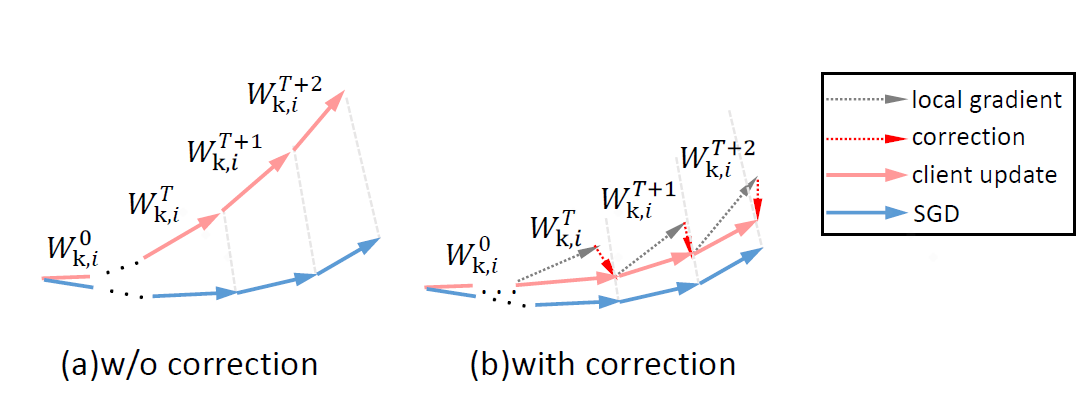}
    \caption{(a) is the divergence between FedPE and SGD, (b) is gradient correction term.}
    \label{fig:dcpn2}
\end{figure}

As we can see in the Fig.\ref{fig:dcpn2}, gradient correction term ensures the update moves towards the true optimum.

\begin{figure}[htp]
    \centering
    \includegraphics[width=15cm]{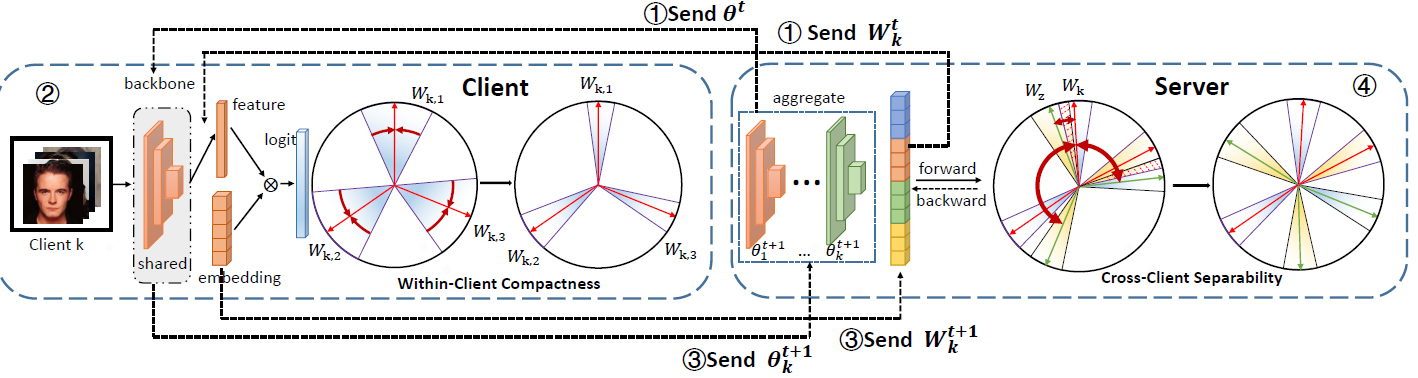}
    \caption{Framework of classical FL and the FL framework with DPLC.}
    \label{fig:dcpng1}
\end{figure}

The whole training schedule is shown in Fig. \ref{fig:dcpng1}.
In the t-th round communication, the server broadcast model parameters $(\theta^{t},W_{k}^{t})$ to the selected clients.
The clients compute a update  model parameters with the local data  asynchornously, and then sent the new  model parameters to the server.
Finally, The server conduct cross-client optimization by collecting and aggregating  the client updates.
The process of server optimization can make correct gradients and make cross-client embeddings spreadout.

\begin{figure}[htp]
    \centering
    \includegraphics[width=10cm]{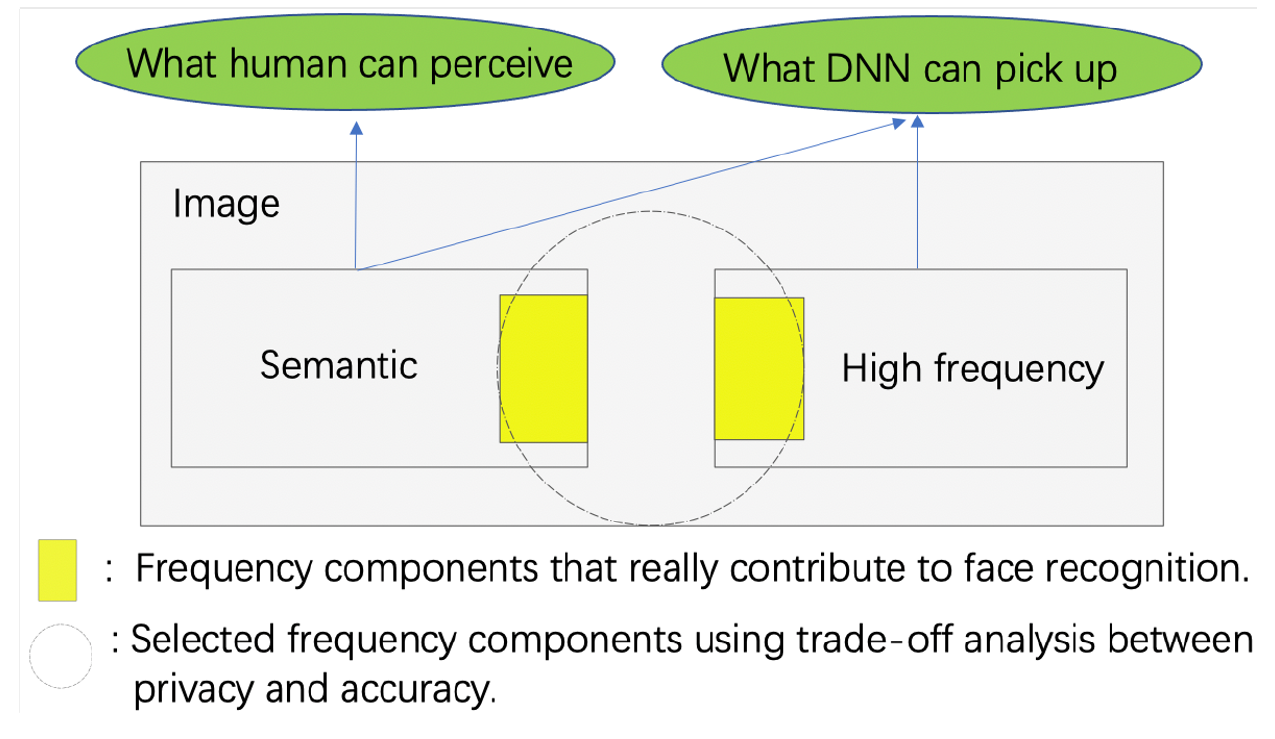}
    \caption{Image perception in the frequency domain.}
    \label{fig:df1png}
\end{figure}

\cite{wang2022privacy} proposed a frequency domain method to achieve Pricacy-Preserving.
As show in Fig. \ref{fig:df1png}, the low frequency of the picture directly affects the human visual perception of the picture, and accounts for most of the semantic. \cite{wang2020high} proposed existing deep neural network based FR systems rely on both low- and high-frequency components.
Yellow rectangles represent frequency components that really contribute to face recognition.
We need a trade-off analysis network to get the corresponding frequency components.

\begin{figure}[htp]
    \centering
    \includegraphics[width=11cm]{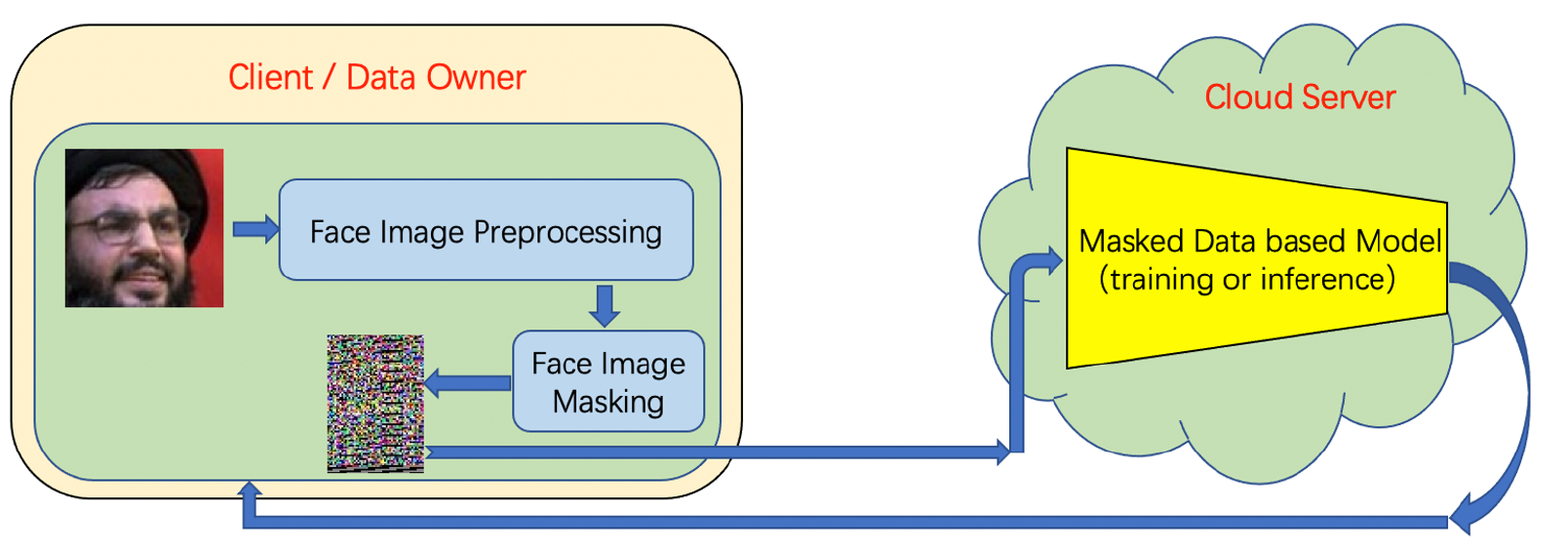}
    \caption{Framework of privacy-preserving FR.}
    \label{fig:df2png}
\end{figure}

Different from some cryptography based method\cite{mohassel2017secureml,makriepic,ma2019lightweight,wagh2020falcon}, \cite{wang2022privacy} proposed a frequency-domain
privacy-preserving FR scheme, which integrated an analysis network to collect the components with
the same frequency from different blocks and a fast masking method to further protect the remaining frequency components.
As show in Fig. \ref{fig:df2png}, the proposed privacy-preserving method includes client data processing part and cloud server training part.

\begin{figure}[htp]
    \centering
    \includegraphics[width=15cm]{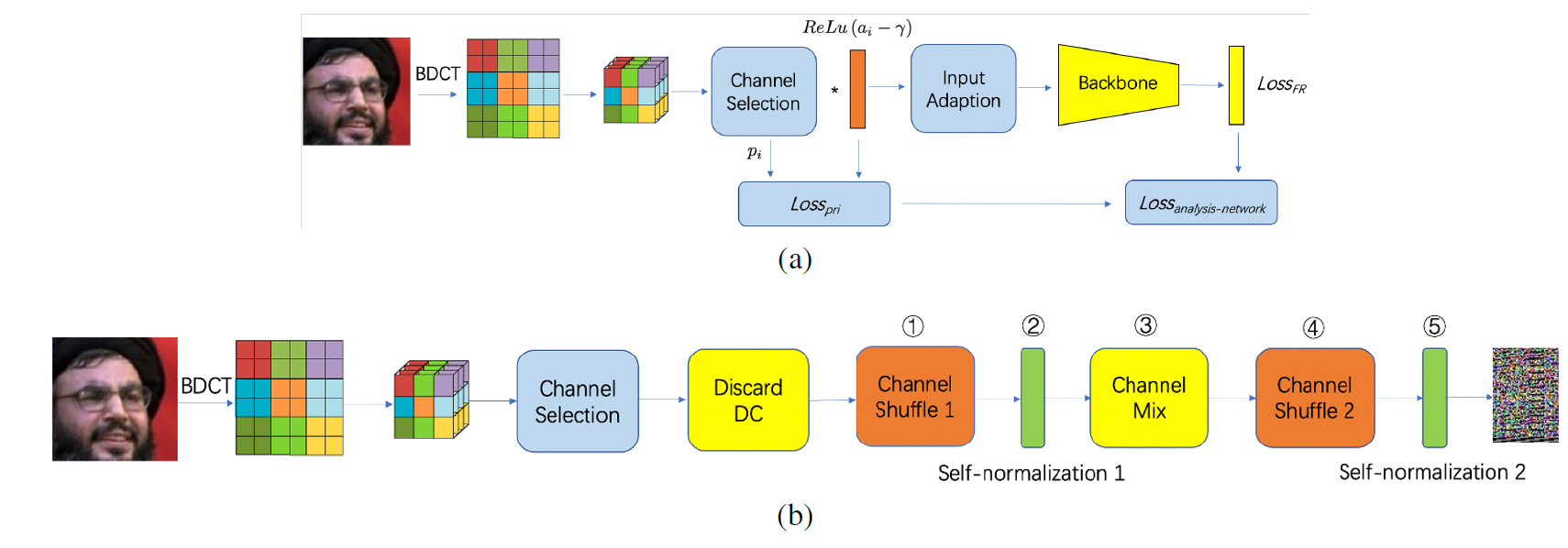}
    \caption{Schematic diagrams of (a) the proposed analysis network and (b) the proposed fast face image masking method.}
    \label{fig:df3png}
\end{figure}

The analysis network is show in Fig. \ref{fig:df3png}(a), the first step is block discrete cosine transform (BDCT), which is carried out on the face image obtained after converting it from a color image to a gray one. Then send the BDCT frequency component channels into the channel selection module.
In this work, for simplicity, they chose to remove a pre-fixed subset of frequency component channels that span low to relatively high frequencies.
The whole network was constrained by the following Loss function.
\begin{equation}
Loss_{analysis-network}=Loss_{FR}+\lambda Loss_{pri}
\end{equation}
$Loss_{FR}$ is the loss function of FR, such as arcface, CosFace, and $\lambda$ is a hyper
parameter. $Loss_{pri}$ is a loss function  to quantify face images’ privacy protection level.

\begin{equation}
    Loss_{pri}=\sum_{n=1}^M ReLu(a_{i}- \gamma) p
\end{equation}

where $a_{i}$ is the trainable weight coefficient for the $i_{th}$ channel,
p is the energy of channel i, denotes a threshold, and
M is the number of considered frequency channels. Clearly, if $a_{i}$  is less than  $\gamma$
, the corresponding channel is considered unimportant in terms of its contribution to the loss function $Loss_{pri}$. This is realized by the use of ReLu(y) function, which becomes zero when y is negative.

The whole diagram of the face image masking method is shown in Fig. \ref{fig:df3png}(b). It performs the BDCT and selects channels according to the analysis network.
Next, the remaining channels are shuffled two times with a channel mixing in between.
After each shuffling operation, channel self-normalization is performed.
The result of the second channel self-normalization is the masked face image that will be transmitted to third-party servers for face recognition.
The proposed face masking method aims at further increasing the difficulty in recovering the original face image from its masked version. Face fast masking feature then feed into the cloud server to finetune the model, as show in Fig. \ref{fig:df2png}.

\begin{figure}[htp]
    \centering
    \includegraphics[width=10cm]{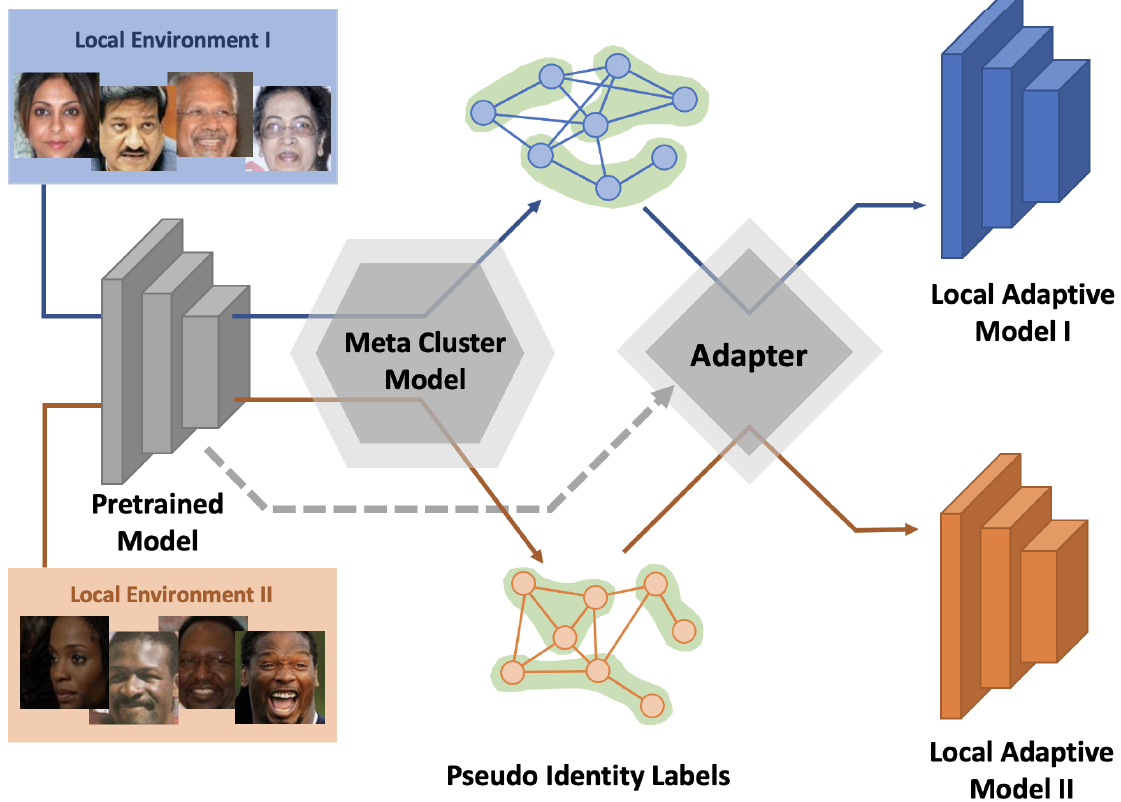}
    \caption{Local-Adaptive Face Recognition (LaFR).}
    \label{fig:cluster2}
\end{figure}

As we know that a model is predominantly trained on RGB images, it generalizes poorly for images captured by infrared cameras.
Likewise, for a model pre-trained on Caucasian only, it performs substantially worse for African and Indian.
\cite{zhu2022local} proposed a Local-Adaptive face recognition method to properly adapt the pre-trained model to a 'specialized' one that  tailors for the specific environment in an automatic and unsupervised manner.
As show in  Fig. \ref{fig:cluster2}, 
it starts from an imperfect pre-trained global model deployed to a specific environment.
The first step of \cite{zhu2022local} was  to train a graph-based meta-clustering model, which refer to \cite{yang2020learning}.
The conventional GCN training is via the following equation:
\begin{equation}
    \phi ^{'} =\phi - \alpha \nabla_{\phi} L_{mtr}^g (\phi) 
\end{equation}
They changed the usually GCN training strategy to a meta-learning, with the objective loss function:
\begin{equation}
    \phi  =\phi -\beta (\nabla_{\phi} L_{mtr}^g (\phi) + \xi \nabla_{\phi}L_{mte}^g(\phi ^{'}))
\end{equation}
The client private data label are  obtained by running the meta-clustering model.
The pseudo 'identity' labels  as well as their corresponding images are used to train the adapted model $\theta_{A}$ from  an imperfect model $\theta_{0}$.
They employed AM-softmax as the training objective to optimization initial imperfect model, the loss function is as below.
\begin{equation}
L(\theta) = - \frac{1}{N} \sum_{i=1}^{N}log \frac{e^{\gamma(cos(\theta_{y_{i}}-m))}}{e^{\gamma(cos(\theta_{y_{i}}-m))}+\sum_{j \neq- y_{i}}^{C}e^{\gamma cos(\theta_{y_{i}})}}
\end{equation}
where C is the total number of classes, and m is the margin that needs empirically determined.

They thinked that the initial mdoel $\theta_{0}$ is already has  strong discriminative power,
so they transfered this pre-trained class center as prior knowledge to the adapted model $\theta_{A}$.
They wanted to have $C_{y_{i}^{\theta_{A}}}$ to be as close to $C_{y_{i}^{\theta_{0}}}$ as possible during the adaptation.
The  $C_{y_{i}^{\theta_{0}}}$ denotes as the center of face face embedding for $y_{i}$ on the pre-trained model.
\begin{equation}
   C_{y_{i}^{\theta_{0}}} =  \frac{1}{M_{i} 1(y_{k}=y_{i})f_{k}^{\theta_{0}}}
\end{equation}
To further reduce the overfitting risk when adapting to a small dataset, they added another model regularization term to let $\theta_{A}$ not deviate too much from $\theta_{0}$.
They final loss function was defined as follow:
\begin{equation}
  L(\theta) = - \frac{1}{N} \sum_{i=1}^{N}log \frac{e^{\gamma(cos(\theta_{y_{i}}-m))}}{e^{\gamma(cos(\theta_{y_{i}}-m))}+\sum_{j \neq- y_{i}}^{C}e^{\gamma cos(\theta_{y_{i}})}}+ \lambda \lvert \theta_{A}-\theta_{0} \rvert_{2}^{2}
\end{equation}
subject to
\begin{equation}
  W=W^{*}/||W^{*}||,
\end{equation}
\begin{equation}
  f=f^{*}/||f^{*}||,
\end{equation}
\begin{equation}
  cos\theta_{j}=W_{j}^{T}f_{i},
\end{equation}
\begin{equation}
  W_{y_{i}}=C_{y_{i}}^{\theta_{0}}
\end{equation}

where W is the normalized clssifier matrix, $\lambda$  is a hyperparameter to trade-off the face loss and model regularization term.
They initialize each $W_{y_{i}}$ with $C_{y_{i}^{\theta_{0}}}$.

\begin{figure}[htp]
    \centering
    \includegraphics[width=15cm]{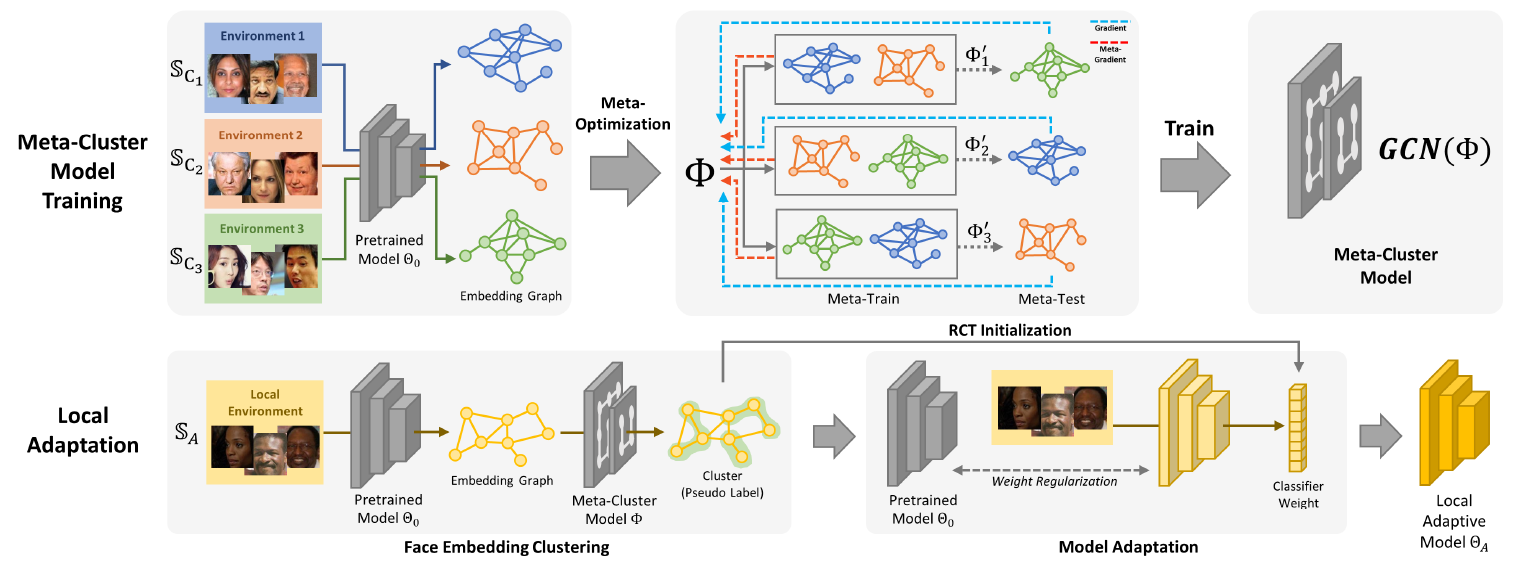}
    \caption{Overview of our Local-Adaptive face
    recognition (LaFR) framework.}
    \label{fig:cluster3}
\end{figure}

As show in Fig. \ref{fig:cluster3},
it shows the training process of the agent's adaptive model.
First, GCN model is obtained through a meta-cluster model training, which used to get the client privacy-data's  pseudo label.
Second, the regularized center transfer (RCT) initialization method is used to optimize the pretrained model to obtain a local adaptive model.
Third, the locally client are aggregated via federated learning \cite{2016Federated} in a secure way.

FedFR \cite{liu2022fedfr} proposed a new FL joint optimization  framework for generic and personalized face recognition.
As we all konw, the universal FedAvg framework usually adopt  public face recognition datasets used for the training of server model.
The server model performance may degenerated during  the communication with local agent.

\begin{figure}[htp]
    \centering
    \includegraphics[width=15cm]{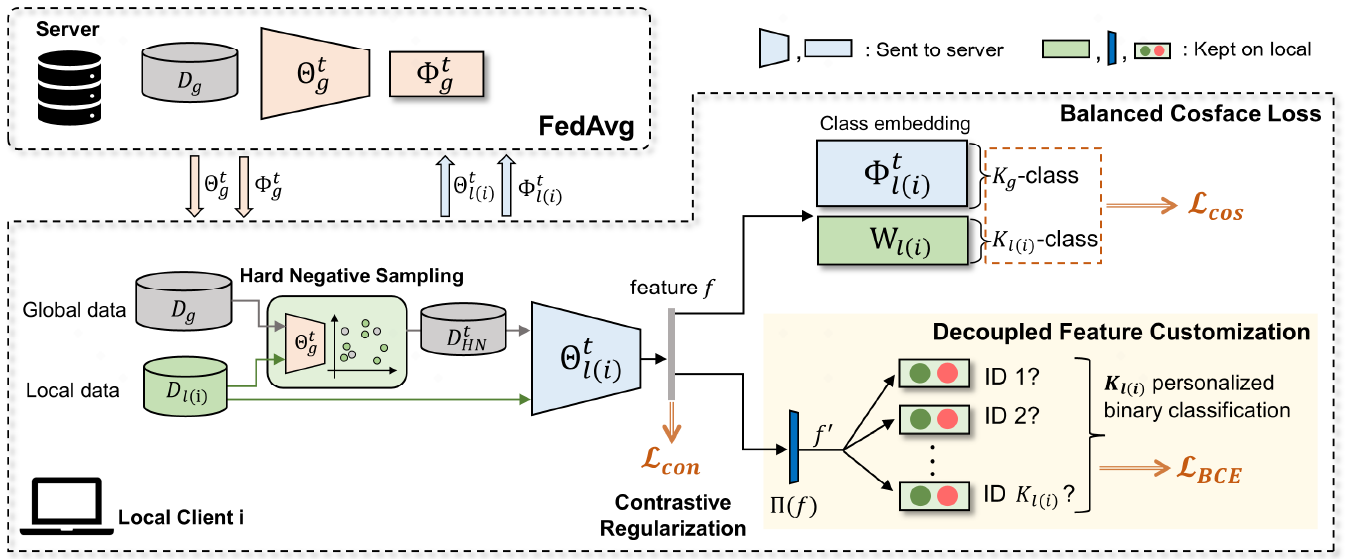}
    \caption{Overview of the FedFR framework.}
    \label{fig:FedFr}
\end{figure}

FedFR supposed to  simultaneously improve the general face representation at the center server, and generate an optimal personalized model for each client without transmitting private identities’ images or features out of the local devices.
As show in Fig. \ref{fig:FedFr}, they add some novel ideas on the classical FL framework.

First, they proposed a hard negative sampling which can auxiliary local agent training.
The $D_{HN}^{t}$ represent the hard negative samples selected from the global shared public dataset.
They calculated the pair-wise cosine
similarity between the proxy and the global data and local data features. Then selected the similarity score large than a threshold.
$D_{HN}^{t}$ and the local data were used for training the local client, which can prevent $\theta_{l}^{t}$ overfitting on local data.

Second, they proposed a contrastive regularization on local clients, which can be used for optimizing the server's general recognition ability.

The contrastive regularization formula is shown as below:

\begin{equation}
L_{con}=-log\frac{exp(sim(f,f_{glob})/\phi)}{exp(sim(f,f_{glob})/\phi+exp(sim(f,f_{prev})/\phi}
\end{equation}

$f_{glob}$ means the global
model ($f_{glob} = \theta_{g}^{t}(x)$), and the $f_{prev}$ represent the face representation learned by the local model at time t-1.
Namely, the expected to decrease the distance
between the face representation learned by the local model
at time $t$ and the one learned by the global model, and increase the distance between the face representation learned by the local model at time $t$ and time $t-1$.

Third, they added a new decoupled feature customization module to achieve the personalization, which can improve the local user experience.
They adopt a transformation with a fully-connected
layer to map the global feature to a client-specific feature space, which can recognize the $k_{l}$ identities well.
Given the transformed feature f, they feed it into $k_{l}$ binary classification branches. Every branch module contains learnable parameters which only target on classifying the positive samples from the k-th class and the negative samples from “any other” classes. 


\section{Backbone size and data distribution}\label{sec5_tmp}
Previous parts provide a comprehensive survey of face recognition algorithms, but rarely mentioned the effect caused by the backbone size and the distribution of training data, which is as important as the former. Different from algorithms which are mostly designed for a specific hypothesis, backbone size and data distribution can affect all scenarios. In this section, we will discuss them from 
three aspects: backbone size, data depth and breadth, and long tail distribution.

\subsection{Backbone size}
It is widely known that training on a large dataset can improve algorithm performance. 
However, for a specific backbone, when the training data size achieves a certain amount, its performance is no longer able to be significantly enhanced by adding data. Meanwhile, the training cost is dramatically increased. Therefore, we aimed at figuring out the effect on the performance of different algorithms caused by increasing training data amount.


We chose Iresnet50, Iresnet100 and Mobilefacenet as the backbone
and selected 10\%, 40\%, 70\% and 100\% ids respectively from Webface42m as training data.
We adopted Arcface loss and PartialFC operation to achieve convergence.
The network is trained on 8 Tesla V100 GPUs with a
batch size of 512 and a momentum of 0.9. We employed
SGD as the optimizer. The weight decay is set to $5e-4$.
We evaluated the model on four test datasets the LFW, the AgeDB, the CFP-FP, and the IJB-C.

\begin{figure}[htbp]
\centering
\subfigure[Mobilefacenet]{
\includegraphics[width=4.2cm]{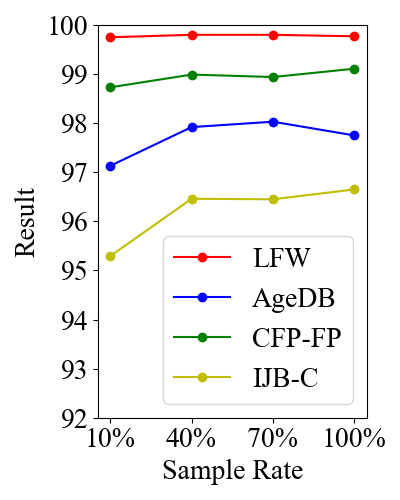}
}
\quad
\subfigure[Iresnet50]{
\includegraphics[width=4.2cm]{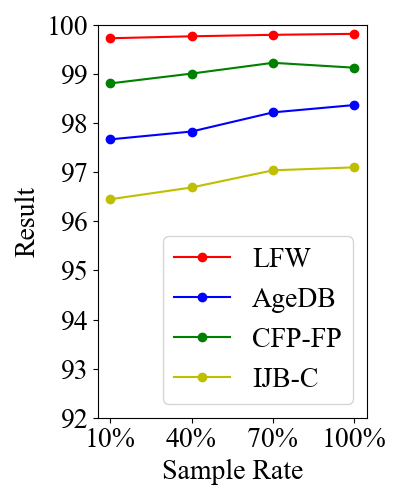}
}
\quad
\subfigure[Iresnet100]{
\includegraphics[width=4.2cm]{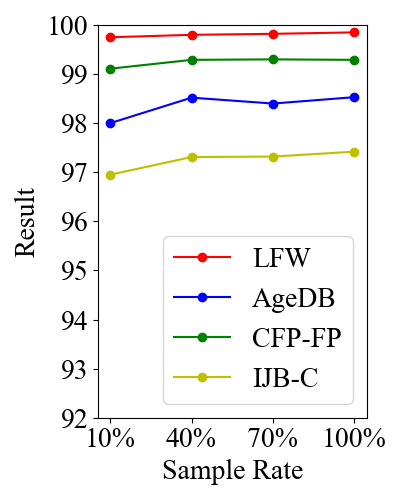}
}
\caption{Results of models with different backbone sizes on LFW,  AgeDB,  CFP-FP, and IJB-C test datasets.}
\label{fig:backboneresult}
\end{figure}

The results are shown in Fig. \ref{fig:backboneresult}. For the Mobilefacenet, as the sample rate increases from 10\% to 40\%, model performance obviously enhances on four test datasets, from 99.75\% to 99.80\% on LFW, from 97.13\% to 97.92\% on AgeDB, from 98.73\% to 98.99\% on CFP-FP, and from 95.29\% to 96.46\% on IJB-C. When the sample rate is over 40\%, the performance of Mobilefacenet remains stable. For the Iresnet50, the turning point is 70\% sample rate. While, the performance of the Iresnet100 improves slightly and continually with the increase in training data. For three different backbones, it is clear that the model performance improves as the amount of training data increases.

\subsection{Data depth and breadth}

For data collection, we can collect images from a limited number of subjects, but each subject contains lots of images, which improves the depth of a dataset and ensures the intra-class variations of a dataset. We also can gather images from lots of subjects, but only collect a limited number of images for each subject, which improves the breadth of a dataset and allows the algorithm to be trained by sufficient different identities. Previous works \cite{wang2021deep} mentioned the two kinds of data collection methods, however, did not discuss their influence. In industry, improving the breadth is easier than improving the depth. Therefore, in this part, we aimed at comparing the importance of data breadth and data depth when the number of training data is fixed, which can guide the formulation of a data acquisition strategy. 

We chose Iresnet100 as the backbone, which had enough ability to express the  distribution of training data.
We had four settings, the product of person numbers and images of each person is fixed to 80k in every setting.
So the four settings can be expressed as 1w\_80, 2w\_40, 4w\_20 and 8w\_10.
For example, 1w\_80 meaned that 10k person and each id contained 80 images.
The training details were the same as section 5.1.

\begin{figure}[htp]
    \centering
    \includegraphics[width=7cm]{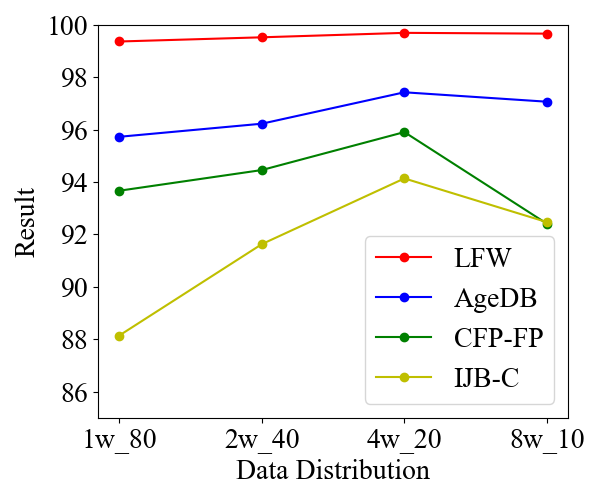}
    \caption{Results on LFW,  AgeDB,  CFP-FP, and IJB-C. N\_S implies that the corresponding dataset has N identities with S samples per identity.}
    \label{fig:d_w}
\end{figure}

Fig. \ref{fig:d_w} compares the model performance with different training data distributions on four test sets. From ``1w\_80" to ``4w\_20", we can see that the model performance rises dramatically, from 99.37\% to 99.70\% on LFW, from 95.73\% to 97.43\% on AgeDB, from 93.67\% to 95.91\% on CFP-FP, and from 88.14\% to 94.14\% on IJB-C. But when the data width is 8w and the depth is 10, the performance declines sharply on CFP-FP (92.41\%) and IJB-C (92.47\%). Based on the results, there is a balance point between the depth of training data and the width of training data.

\subsection{Long tail distribution}
The purity and long-tail distribution  of training data were essential factors affecting the performance of state-of-the-art face recognition models. In order to explore the impact of these two factors on the performance of face recognition models, we conducted self-learning based cleaning similar with \cite{feng2021towards} to get a pure dataset WebFace35M.
The details are introduced as follows:
(1) An initial model is first trained with the WebFace42M
to clean the original dataset, which mainly consist of DBSCAN-cluster,
and Intra-ID cleaning;
(2) Then the the i-th model is trained on the cleaned datasets from (1);
(3) We iterate this process by initializing the i-th model as the
(i+1)-th model. 
Different from \cite{feng2021towards}, we only conducted Intra-ID cleaning.
Because WebFace42M is derived from WebFace260M, and through our manual observation, we found that WebFace42M contained intra-id noisy.
We conducted DBSCAN-cluster on each folder and eliminated data that didn't  belong to this category to achieve intra-id cleaning. 
Through two rounds of data cleaning, we got the WebFace35M, which further filtered out the ID with images numberes less than 10 as the long tail data.
The following long tail distribution and noisy experiment were based on the WebFace35M.

In the previous chapters, we have discussed long-tail distribution and how to alleviate its effect. It is a typical unbalanced distribution and is defined as a dataset containing a large number of identities that have a small number of face images. Increasing the number of identities can increase training costs due to the last fully connected Layer. In this part, we carefully designed a set of experiments to visually display the effect of the distribution.

We chose Iresnet100 as the backbone and then we added 0\%, 25\%, 50\%, 75\% and 100\% long tail data from id dimension.
The training details were the same as section 5.1.

\begin{figure}[htp]
    \centering
    \includegraphics[width=7cm]{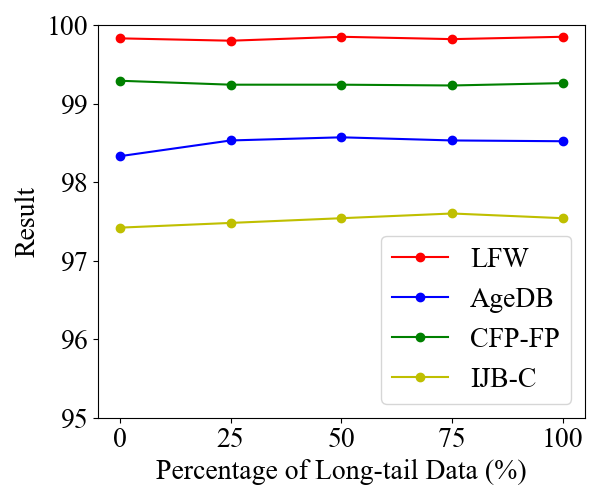}
    \caption{Results on LFW,  AgeDB,  CFP-FP, and IJB-C with different percentage of long-tail data.
    }
    \label{fig:longtail}
\end{figure}

The results are shown in Fig. \ref{fig:longtail}. With the addition of long tail data, model performance keeps steady. The evaluation results on LFW,  AgeDB,  CFP-FP, and the IJB-C are roughly 99.8\%, 98.5\%, 99.3\%, and 97.5\%, respectively. 

\section{Datasets and Comparison Results}\label{sec5}

\subsection{Training datasets}\label{subsec5.1}
In this part, we list all the major training datasets for FR with their details.
All information can be checked in Table \ref{table:train}, including the number of images and identities in these training sets.

\begin{table}[]
\centering
\begin{tabular}{@{}llll@{}}
\toprule
Training Dataset   & IDs     & Images     & Details       \\ 
\midrule
CASIA-Webface\cite{yi2014learning}            & 10,575       & 494,414    & 
    \parbox[c]{5cm}{A semi-automatical way is used to collect face images from Internet} \\
    \hdashline[0.5pt/1pt]
CelebA\cite{liu2015deep}                      & 10k celebs   & 0.2M    \\
    \hdashline[0.5pt/1pt]
UMDFaces\cite{bansal2017umdfaces}             & 8,277        & 367,888    & 
    \parbox[c]{5cm}{A semi-automatical annotation procedure is used to the images crawled from Yahoo, Yandex, Google and Bing. The annotation of this set includes face ID, face bounding box, 21 face landmarks, face pose and gender.} \\
    \hdashline[0.5pt/1pt]
vggface\cite{parkhi2015deep}                  & 2,622        & 2.6M    \\
    \hdashline[0.5pt/1pt]
vggface2(VGG2)\cite{cao2018vggface2}          & 9,131        & 3.31M  &
    \parbox[c]{5cm}{VGG2 consists of a training set with 8,631 identities (3,141,890 images) and a test set with 500 identities (169,396 images). The annotation of this set includes face ID, face bounding box, 5 face landmarks, predicted face pose and age. } \\
    \hdashline[0.5pt/1pt]
MS-Celeb-1m(MS1M)\cite{guo2016ms}             & 100k celebs  & 10M     \\
    \hdashline[0.5pt/1pt]
MS1M-ibug\cite{deng2017marginal}              & 85k celebs   & 3.8M  & 
    \parbox[c]{5cm}{This dataset is also named as MS1MV1, which is obtained by cleansing on MS1M.} \\
    \hdashline[0.5pt/1pt]
MS1M-ArcFace\cite{deng2019arcface}            & 85k celebs   & 5.8M  & 
    \parbox[c]{5cm}{This dataset is also named as MS1MV2, which is obtained by cleansing on MS1M.} \\
    \hdashline[0.5pt/1pt]
Asian-Celeb                                   & 94k celebs   & 2.8M   & 
    \parbox[c]{5cm}{This dataset has been excluded from both LFW and MS-Celeb-1M-v1c.} \\
    \hdashline[0.5pt/1pt]
Glint360K\cite{an2020partial}                 & 360k         & 17M     \\
    \hdashline[0.5pt/1pt]
DeepGlint                                     & 181k         & 6.75M  &
    \parbox[c]{5cm}{This dataset is obtained by merging MS1M and Asian-Celeb with data cleansing.} \\
    \hdashline[0.5pt/1pt]
IMDB-Face\cite{wang2018devil}                 & 59k          & 1.7M   &
    \parbox[c]{5cm}{The images in this set is collected from IMDb website.}\\
    \hdashline[0.5pt/1pt]
Celeb500k\cite{cao2018celeb}                  & 500k         & 50M     \\
    \hdashline[0.5pt/1pt]
WebFace260M\cite{zhu2021webface260m}          & 4M           & 260M    \\
    \hdashline[0.5pt/1pt]
MegaFace(train)\cite{kemelmacher2016megaface}            & 672k         & 4.7M  & 
    \parbox[c]{5cm}{This No longer being distributed from official website} \\ 
\bottomrule
\hline
\end{tabular}
\label{table:train}
\caption{Information about training datasets in FR}
\end{table}

\subsection{Testing datasets and Metrics}\label{subsec5.2}

In this part, we first list all commonly used test sets in table \ref{table:test}.
Then we will introduce the evaluation metrics/protocols on these set.
And we will provide some evaluation results of each state-of-the-art on these metrics.

\begin{table}[]
\centering
\begin{tabular}{@{}llll@{}}
\toprule
Test Dataset    & IDs     & Images    & Details      \\ 
\midrule

LFW\cite{huang2008labeled}            &   5749           & 13,233 \\
    \hdashline[0.5pt/1pt]
CPLFW\cite{zheng2018cross} 
    & -  & - &
    \parbox[c]{4.5cm}{For cross-pose challenge in FR.}   \\
    \hdashline[0.5pt/1pt]
CALFW\cite{zheng2017cross}                          & -  & - &
    \parbox[c]{4.5cm}{Both CPLFW and CALFW are derivative datasets of LFW, addressing cross-pose and cross-age challenge in FR. They contains different 6k positive and negative pairs.}   \\
    \hdashline[0.5pt/1pt]
YTF\cite{wolf2011face}                & 1,595            & 3,424 videos \\
    \hdashline[0.5pt/1pt]
MegaFace gallery                      & 0.69M            & more than 1M   \\
    \hdashline[0.5pt/1pt]
Facescrub\cite{ng2014data} 
                                      & 530 celebs       & 106,863 &
    \parbox[c]{4.5cm}{The images were retrieved from the Internet and are taken under real-world situations (uncontrolled conditions). Name and gender annotations of the faces are included.} \\
    \hdashline[0.5pt/1pt]
FGNet\cite{lanitis2002toward}         & 82               & 1002   &
    \parbox[c]{4.5cm}{The dataset includes lots of face images at the age phase of the child and the elderly,with age from 0 to 69.} \\
    \hdashline[0.5pt/1pt]
CACD\cite{chen2014cross}              & 2,000 celebs     & 163,446  &
    \parbox[c]{4.5cm}{CACD-VS is a subset of CACD, which consists of 4000 face image pairs with different ages for age-invariant face verification, and the face pairs are divided into 2,000 positive pairs and 2,000 negative pairs.} \\
    \hdashline[0.5pt/1pt]
MORPH Album 2\cite{ricanek2006morph}  & 20,000           & 78,000 &
    \parbox[c]{4.5cm}{Containing individuals across different ages.} \\
    \hdashline[0.5pt/1pt]
CFP\cite{sengupta2016frontal}         & 500              & 7,000 &
    \parbox[c]{4.5cm}{Each ID contains 10 frontal and 4 profile images.} \\
    \hdashline[0.5pt/1pt]
IJB-A\cite{klare2015pushing}          & 500              &  \tabincell{l}{5,396 images \\ 20,412 frames} &
    \parbox[c]{4.5cm}{Proposed by NIST(National Institute of Standards and Technology)
    }
    \\
    \hdashline[0.5pt/1pt]
IJB-B                                 &   1,845          & \tabincell{l}{21,798 images \\ 55K frames } & 
    \parbox[c]{4.5cm}{IJB-B is an extension of the IJB-A.} \\
    \hdashline[0.5pt/1pt]    
IJB-C                                 &   3,531          & \tabincell{l}{31,334 images \\ 117,542 frames \\ 11779 videos} & 
    \parbox[c]{4.5cm}{IJB-C is derived from IJB-A, and it includes 10040 non human face images.} \\
    \hdashline[0.5pt/1pt]
Multi-PIE\cite{gross2010multi}        & 337              & 754,204    &
    \parbox[c]{4.5cm}{Identities in this set is from 15 view points and 20 illumination conditions for evaluating pose invariant FR.} \\
    \hdashline[0.5pt/1pt]    
AGE-DB\cite{moschoglou2017agedb}   & 568 celebs       & 16,488   &
    \parbox[c]{4.5cm}{The annotation contains age information.} \\

\bottomrule
\hline
\end{tabular}
\caption{Information about test datasets in FR}
\label{table:test}
\end{table}

Here, we give some metrics used by latest FR papers.

1, verification accuracy from unrestricted with labeled outside data protocol \cite{huang2014labeled}.
In this metrics, a list of face pairs is be provided, with binary labels which shows each pair sharing same ID or not. And the accuracy is the proportion of right prediction by evaluating algorithm. 
LFW is a typical set for testing verification accuracy. 6000 face pairs from LFW/CALFW/CPLFW are provided for evaluation.
The metrics of verification accuracy has also been applied in other test datasets, such as CFP-FP and AgeDB. 
CFP-FP is a verification experiment from dataset CFP, which contains face frontal-profile pairs.
YTF provides 5000 video pairs for evaluation on this metrics.
Here we list some comparison results of state-of-the-arts in verification accuracy in Table \ref{table:veri-acc}.
We will not further list evaluation results of all methods in FR on verification accuracy, because different methods trained on different training sets with different backbones.  

\begin{table}[]
\centering
\begin{tabular}{@{}llllll@{}}
\toprule
Method    & LFW     & AgeDB    & CFP-FP    & CALFW    & CPLFW      \\ 
\midrule

Softmax                                      & 99.45   & 96.58  & 92.67  & 93.52  & 86.27      \\
Center loss\cite{wen2016discriminative}      & 99.65   & 96.83  & 93.37  & 94.23  & 86.58      \\
Triplet loss\cite{schroff2015facenet}        & 99.58   & 96.27  & 92.30  & 93.27  & 85.07      \\
UniformFace\cite{duan2019uniformface}        & 99.70   & 96.90  & 94.34  & 94.40  & 87.45      \\
SphereFace\cite{liu2017sphereface}           & 99.70   & 96.43  & 93.86  & 94.17  & 87.81      \\
CosFace\cite{wang2018cosface}                & 99.73   & 97.53  & 94.83  & 95.07  & 88.63      \\
ArcFace\cite{deng2019arcface}                & 99.75   & 97.68  & 94.27  & 95.12  & 88.53      \\
AdaCos\cite{zhang2019adacos}                 & 99.68   & 97.15  & 94.03  & 94.38  & 87.03      \\
AdaM-Softmax\cite{liu2019adaptiveface}       & 99.74   & 97.68  & 94.96  & 95.05  & 88.80      \\
MV-softmax\cite{wang2020mis}                 & 99.72   & 97.73  & 93.77  & 95.23  & 88.65      \\
ArcNegFace\cite{liu2019towards}              & 99.73   & 97.37  & 93.64  & 95.15  & 87.87      \\
CurricularFace\cite{huang2020curricularface} & 99.72   & 97.43  & 93.73  & 94.98  & 87.62      \\
CircleLoss\cite{sun2020circle}               & 99.73   & -      & 96.02  & -      & -          \\
NPCFace\cite{zeng2020npcface}                & 99.77   & 97.77  & 95.09  & 95.60  & 89.42      \\
MagFace\cite{meng2021magface}                & 99.83   & 98.17  & 98.46  & 96.15  & 92.87      \\
AdaFace\cite{kim2022adaface}                 & 99.80   & 97.90  & 99.17  & 96.05  & 94.63      \\

\bottomrule
\hline
\end{tabular}
\caption{Verification accuracy ($\%$) on easy benchmarks}
\label{table:veri-acc}
\end{table}


2, testing benchmark of MegaFace dataset.
MegaFace test dataset contains a gallery set and a probe set. The
gallery set contains more than 1 million images from 690K
different individuals. 
The probe set consists of two existing datasets: Facescrub \cite{ng2014data} and FGNet. 
MegaFace has several testing scenarios including identification, verification and
pose invariance under two protocols (large or small training set). 
(The training set is viewed as small if it is less than 0.5M.)
For face identification protocol, CMC (cumulative match characteristic) and ROC (receiver operating characteristics) curves will be evaluated.
For face verification protocol, ``Rank-1 Acc.'' and ``Ver.'' will be evaluated. ``Rank-1 Acc.'' indicates rank-1 identification accuracy with 1M distractors.
``Ver.'' indicates verification TAR for 10-6 FAR. TAR and FAR denote True Accept Rate and False Accept Rate respectively.

3, IJB-A testing protocol.
This protocol includes `compare' protocol for 1:1 face verification and the `search' protocol for 1:N face identification. 
For verification, the true accept rates (TAR) vs. false positive rates (FAR) are reported. 
For identification, the true positive identification rate (TPIR) vs. false positive identification rate (TPIR) and the Rank-N accuracy are reported.
Recently, less papers published their results on IJB-A, since they have reached a high performance on it. 
The protocol on IJB-A can also be applied on IJB-B and IJB-C. 
The comparison results on IJB-B and IJB-C (TAR@FAR=1e-4, FAR=1e-5) are shown in Table .

\begin{table}[]
\centering
\begin{tabular}{@{}llllll@{}}
\toprule
Method    & IJB-B 1e-4     & IJB-B 1e-5    & IJB-C 1e-4    & IJB-C 1e-5     \\ 
\midrule

Softmax                                      & 85.66   & 73.63   & 86.62   & 76.48      \\
Center loss\cite{wen2016discriminative}      & 86.43   & 74.16   & 86.87   & 76.64      \\
Triplet loss\cite{schroff2015facenet}        & 73.21   & 40.37   & 78.12   & 48.07      \\
UniformFace\cite{duan2019uniformface}        & 87.22   & 75.01   & 88.87   & 79.64      \\
SphereFace\cite{liu2017sphereface}           & 86.67   & 74.75   & 87.92   & 78.77      \\
CosFace\cite{wang2018cosface}                & 90.60   & 82.28   & 91.72   & 86.68      \\
ArcFace\cite{deng2019arcface}                & 90.83   & 82.68   & 91.82   & 85.75      \\
AdaCos\cite{zhang2019adacos}                 & 86.04   & 73.34   & 87.53   & 78.91      \\
AdaM-Softmax\cite{liu2019adaptiveface}       & 90.54   & 82.70   & 91.64   & 86.84      \\
MV-softmax\cite{wang2020mis}                 & 90.67   & 83.17   & 92.03   & 87.52      \\
ArcNegFace\cite{liu2019towards}              & 90.62   & 81.59   & 90.91   & 85.64      \\
CurricularFace\cite{huang2020curricularface} & 90.04   & 81.15   & 90.95   & 84.63      \\
NPCFace\cite{zeng2020npcface}                & 92.02   & 85.59   & 92.90   & 88.08      \\
MagFace\cite{meng2021magface}                & 94.51   & 90.36   & 95.97   & 94.08      \\
AdaFace\cite{kim2022adaface}                 & 96.03   & -       & 97.39   & -          \\

\bottomrule
\hline
\end{tabular}
\caption{Verification accuracy ($\%$) on IJB-B and IJB-C.}
\label{table:veri-acc-ijb}
\end{table}

The aforementioned protocols are used to evaluating FR methods without any restriction.  
On the contrary, the Face Recognition Under Inference Time conStraint (FRUITS) protocol \cite{zhu2021webface260m} is designed to comprehensively evaluate FR systems (face matchers in verification way) with time limitation.
In particular, the protocol FRUITS-x (x can be 100, 500 and 100) evaluates a whole FR system which must distinguish
image pairs within x milliseconds, including preprocessing (detection and alignment), feature embedding, and matching.
FRUITS-100 targets on evaluating lightweight FR system which can be deployed on mobile devices.
FRUITS-500 aims to evaluate modern and popular networks deployed in the local surveillance system.
FRUITS-1000 aims to compare capable recognition models performed on clouds.

More specific metrics in the other competitions in the field of FR will be concluded in section \ref{sec7}.

\section{Applications}\label{sec6}
In this section, we will introduce applications by using face embeddings.
Most important applications will be face verification and identification.
We will not give more information about them in this section, because section \ref{subsec3.3} has presented the details.
Using the face features extracted from FR system, we can implement a lot of other applications, such as face clustering, attribute recognition and face generation, which will be elaborated in the following subsections. 

\subsection{Face clustering}\label{subsec6.1}
Given a collection of unseen face images, face clustering groups images from the same person together. 
This application can be adopted in many areas of industry, such as face clustering in photo album, characters summarizing of videos.
Face clustering usually uses face features from a well trained FR system as input. 
Therefore high quality face embeddings have positive effect on face clustering.
In this section, we do not consider the generation of face embedding. Instead, we give the clustering procedure afterwards.

There are two types of face clustering methods. 
The first type treats each face embedding as a point in a feature space, and utilizes unsupervised clustering algorithms. 
Unsupervised methods usually achieve great clustering results in the condition of data distribution conforming to certain assumptions.
For instance, K-Means \cite{lloyd1982least} requires the clusters to be convex-shaped,
Spectral Clustering \cite{shi2000normalized} needs different clusters to be balanced in the number of instances, 
and DBSCAN \cite{ester1996density} assumes different clusters to be in the same density.
The second one adopts GCN (Graph Convolutional Network) to group features. GCN based cluster methods are supervised, thus they can generally outperform unsupervised clustering algorithms. 

In this subsection, we introduce some recently published GCN based cluster methods. First, we set some annotations here.
Given a face dataset, features of all face images will be extracted by a trained CNN, forming a set of features
$X = [x_1,x_2, \dots , x_N]^T \in \mathbb{R}^{n \times d}$, where $n$ is the number of images, and $d$ is the dimension of features.
Then each feature is regarded as a vertex and cosine similarity is used to find $K$ nearest neighbors for
each sample. 
By connecting between neighbors, an affinity graph $G = (V, E)$ is obtained.
A symmetric adjacent matrix $A \in \mathbb{R}^{n \times n}$ will be calculated, where the element $a_{i,j}$ is the cosine similarity between $x_i$ and $x_j$ if two vertices are connected, or zero otherwise.

Yang \emph{et al.} \cite{yang2019learning} proposed a GCN based clustering framework which consists of three modules, 
namely proposal generator, GCN-D, and GCN-S. 
The first module generates cluster proposals (i.e. sub-graphs likely to be clusters), from the affinity graph $A$.
To do so, they remove edges with affinity values below a threshold, and constrain the size of sub-graphs below a maximum number.
GCN-D performs cluster detection. Taking a cluster proposal $P$ as input, it evaluates how likely the proposal
constitutes a desired cluster by two metrics, namely IoU and IoP scores, 
which are defined as:
\begin{equation}
IoU(P) = \frac{ \lvert P \cap  \hat P  \lvert}{  \lvert P \cup  \hat P \lvert} \ \ , \ \ 
IoP(P) = \frac{ \lvert P \cap  \hat P  \lvert}{  \lvert P \lvert}
\end{equation}
where $\hat P$ is the ground-truth set comprised all the vertices with label $l(P)$, and $l(P)$ is the majority label of the cluster
P.
IoU reflects how close P is to the desired ground-truth $\hat P$, while IoP reflects the purity.
GCN-D is used to predict both the IoU and IoP scores, which consists of $L$ layers. The computation of each layer in GCN-D can be formulated as:
\begin{equation}
F_{l+1} = \sigma( D^{-1} (A+I) F_l W_l)
\label{eq:yang19}
\end{equation}
where $D= \sum_j A_{i,j}$ is a diagonal degree matrix.
$F_l$ contains the embeddings of the $l$-th layer.
$W_l$ is a learnable parameter matrix which transforms the embeddings.
$\sigma$ is the ReLU activation function.
While training, the objective is to minimize the mean square error(MSE) between ground-truth and predicted  IoU and IoP scores.
Then GCN-S performs the segmentation to refine the selected proposals, which has similar structure with GCN-D.

Different from \cite{yang2019learning}, Wang \emph{et al.} \cite{wang2019linkage} adopted GCN to predict the similarity between two features.
In detail, Wang \emph{et al.} first proposed the Instance Pivot Subgraphs (IPS). For each instance $p$ in the graph $G$, an IPS is a subgraph centered at a pivot instance $p$, which is comprised of nodes including the KNNs of $p$ and the high-order neighbors up to 2-hop of $p$.
Each layer of the proposed GCN is formulated as:
\begin{equation}
F_{l+1} = \sigma( [ F_l \| (D^{-\frac{1}{2}}AD^{-\frac{1}{2}}) F_l] W_l)
\end{equation}
where operator $\|$ represents matrix concatenation along the feature dimension. 
The annotations in this equation is as the same as eq.(\ref{eq:yang19}).
The input $F_0$ of this GCN is not feature matrix of an IPS, instead, 
$F_{0} = [ \dots , x_q - x_p, \dots ]^T \ , \ q \in V_p$,
where $V_p$ is the node set of IPS with pivot $p$.
While training, the supervised ground-truth is a binary vector whose value in index $q$ is 1 if node $q$ shares a same label with pivot $p$, and 0 if not.
In inference, this GCN predicts the likelihood of linkage between each node with the pivot.
To get clustering result, IPS with each instance as the pivot will be built, linkages in the whole face graph will be obtained.
At last, a pseudo label propagation strategy \cite{zhan2018consensus} will be adopted to cut the graph to get final cluster result.

Yang \emph{et al.} \cite{yang2020learning} proposed a concept of confidence for each face vertex in a graph.
The confidence is the probability of a vertex belonging to a specific cluster.
For a face with high confidence, its neighboring faces tend to belong to the same class 
while a face with low confidence is usually adjacent to the faces from the other classes.
As a result, the confidence $c_i$ of a vertex $i$ can be calculated as:
\begin{equation}
c_i = \frac{1}{\lvert N_i \rvert} \sum_{v_j \in N_i} ( 1_{y_j = y_i} - 1_{y_j \neq y_i} ) a_{i,j}
\end{equation}
where $N_i$ is the neighbour set of node $i$ according to KNN results.
So, they proposed GCN-V with $L$ layers to predict the confidence of each node. And computation of each layer can be formulated as:
\begin{equation}
F_{l+1} = \sigma( [ F_l \| D^{-1}(A+I) F_l] W_l)
\label{eq:yang2020}
\end{equation}
While training GCN-V, the objective is to minimize the mean square error (MSE) between ground truth and predicted confidence scores.
Then they built GCN-E to calculate connectivity of edges, which has similar structure with GCN-V. 
The edge with high connectivity indicates the two connected samples tend to belong to the same class.
The input of GCN-E is a candidate set $S$ for each vertex: 
$S_i = \{ v_j \lvert c_j > c_i , v_j \in N_i  \}$.
Set $S_i$ only contains the vertices with higher confidence than the confidence $c_i$.
In training, ground-truth for input $S_i$ is a binary vector, where for a vertex $v_i$, the GT connectivity (index $j$) is set to 1 if a neighbor $v_j$ shares the same label with the $v_i$, otherwise it is 0.
MSE loss is also used to trained GCN-E.

The aforementioned GCN methods can be roughly divided into global-based (such as \cite{yang2020learning}) and local-based ones (such as \cite{wang2019linkage,yang2019learning}) according to whether their GCN inputs are the whole graph or not.
Global-based methods suffer from the limitation of training data scale, while local-based ones are difficult to grasp the whole graph structure information and usually take a long time for inference. 
To address the dilemma of large-scale training and efficient inference, a STructure-AwaRe Face Clustering (STAR-FC) method \cite{shen2021structure} was proposed.
The proposed GCN consists of 2-layer of MLP, and each layer has similar structure with GCN-V in \cite{yang2020learning} (eq. (\ref{eq:yang2020})).
However, this GCN takes pair features as input and predicts the two dimension edge confidence corresponding to these two nodes, which are connected in the affinity graph $G$.
For inference, a single threshold $\tau_1$ is used to eliminate most of the wrong edges with smaller confidence.
After that, all subgraphs can be treated as clusters, which form a cluster set $C$.
Then the concept of node intimacy ($NI$) between two nodes $v_1$ and $v_2$ is defined as 
$ NI = \max( \frac{k}{n_1} , \frac{k}{n_2} ) $,
where $n_1$ and $n_2$ are the numbers of edges connected to node $v_1$ and $v_2$; $k$ is the number of their common neighbor nodes.
Node intimacy will further purify the clusters.
In detail, a smaller set of clusters will be sampled from $C$ firstly, and all nodes in this set build a new subgraph $S$.
Then, the NI values in $S$ will be calculated, and the edges with lower $NI$ value than $\tau_2$ will be cut. 
Finally, the newly obtained clusters with less false positive edges become final clustering results.

Despite of GCN, transformer can also be used in face clustering.
\cite{guo2020density} abstracted face clustering problem as forming a face chain.
First, data density $\rho$ of a face node $vi$ is proposed as:
\begin{equation}
\rho(v_i) = \sum_{v_j \in N(v_i)} x_i \cdot x_j
\end{equation}
where $\cdot$ is inner product, which calculates the cosine similarity of two normalized features. $N(v_i)$ is as set of neighbours of $v_i$ on affinity graph $G$.
The node with high data density tend to have a high probability to be a certain person.
Given a node $v_k$, a node chain $C(v_k) = \{ v_k = c_k^1 , c_k^2 , \dots , c_k^N  \}$ can be generated by gradually finding its nearest neighbors with higher density:
\begin{equation}
c_k^{i+1} = \arg \max_v \{  x_{c_k^i} \cdot x_v \} \ , \ v \in \{ u \lvert \rho(u) > \rho(c_k^i) , u \in N(c_k^i) \}
\end{equation}
Then, a transformer architecture is designed to further update the node feature. 
For a node chain $C(v_k)$ with $N$ nodes, this transformer predicts a weight set $\{  w(c_k^i) \lvert i=1, \dots ,N \}$ for each of its node.
The final density-aware embedding $\psi (v_k)$ for node $v_k$ is:
\begin{equation}
\psi (v_k) = \sum_{i=1}^N w(c_k^i) \cdot x_{c_k^i}
\end{equation}
At last, these feature is compatible with all kinds of clustering methods, e.g., merging nodes with high similarity, K-means, DBSCAN.

\subsection{Face attribute recognition}\label{subsec6.2}

Predicting face attributes is another widely used application for face embedding. By extracting features from face images, the network could estimate the age, gender, expression, hairstyle, and other attributes of this face. Mostly, the attributes prediction is performed based on the localization results, which have been summarized in Section \ref{subsec3.1}. 

For prediction, multi-task learning was widely utilized to recognize a cluster of attributes at the same time. Liu \emph{et al.} \cite{liu2015deep} proposed the ANet model to extract face features and used multiple support vector machine (SVM) classifiers to predict 40 face attributes. The ANet was pre-trained by the identity recognition task, and then fine-tuned by attributes tags. In the fine-tuned stage, multiple patches of the face region were generated from each face image, and a fast feature extraction method named interweaved operation was proposed to analyze these patches. The outputs of ANet were feature vectors and were utilized to train the SVM classifiers.

ANet utilized the same features to predict all the attributes, however, attributes heterogeneity had not been considered. To address this limitation, some works adjusted the network structure and allowed its last few layers to be shared among a specific category of attributes through a multi-branch structure \cite{han2017heterogeneous,wang2017deep,hand2017attributes,savchenko2021facial}. The attributes were grouped following different grouping strategies. Han \emph{et al.} \cite{han2017heterogeneous} proposed a joint estimation model. Data type, data scale, and semantic meaning were utilized to build the grouping strategy; based on that, four types of attributes were defined, i.e., holistic-nominal, holistic-ordinal, local-nominal, and local-ordinal. The loss function is formulated as:
\begin{equation}
\arg \min_{W_c,\lbrace W^j\rbrace^M_{j=1}}\sum_{g=1}^G\sum_{j=1}^{M^g}\sum_{i=1}^N\lambda^g\mathcal{L}^g(y_i^j,\mathcal{F}(X_i,W^g\circ W_c))+\gamma_1\phi(W_c)+\gamma_2\phi(W_g)
\end{equation}
where G and $M^g$ represent the number of heterogeneous attribute categories and attributes within each attribute category, separately; $\lambda^g$ is an adjustment factor to adjust the weight of each category; $\mathcal{F}(., .)$ denotes the predicted result based on weight vectors $W_c$ and $W^g$; $W_c$ and $W^g$ control shared features among all the face attributes and shared features among the attributes within each category, separately; $y_i^j$ is the ground truth; $X_i$ is the input; $\mathcal{L}$ is the loss function; $\phi$ is regularization function, and $\gamma_1$ and $\gamma_2$ are regularization parameters.

\begin{figure}[htp]
    \centering
    \includegraphics[width=0.95\textwidth]{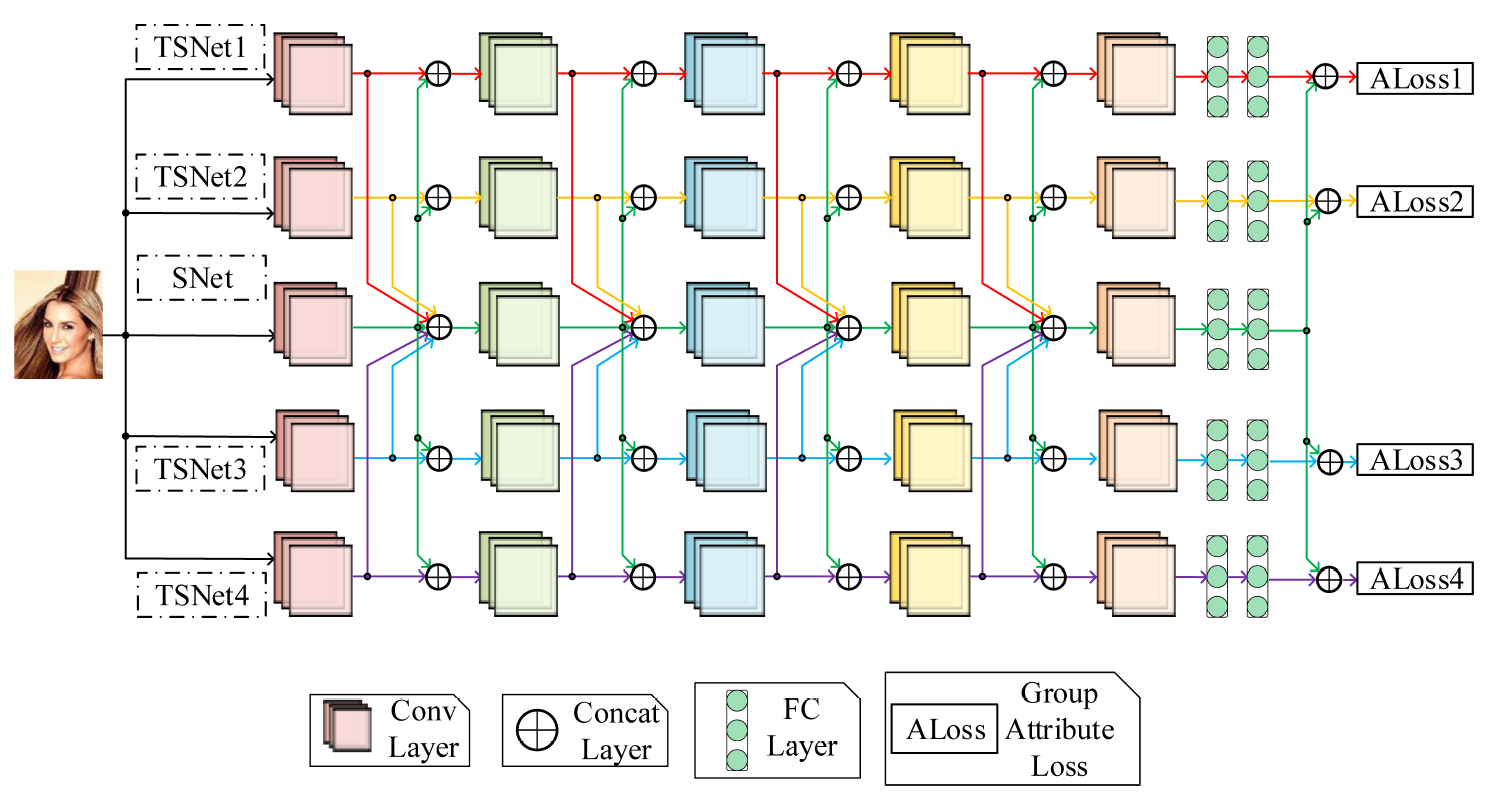}
    \caption{The pipeline of PS-MCNN. \cite{cao2018partially}.}
    \label{cao2018}
\end{figure}

For the multi-branch structure, each branch is trained separately and can not be affected by other branches. However, some researchers thought that the task relation was conducive to the attributes prediction, thus, they added connections between the branches. Cao \emph{et al.} \cite{cao2018partially} proposed a Partially Shared Multi-task Convolutional Neural Network (PS-MCNN) which consisted of the Shared Network (SNet) and the Task Specific Network (TSNet)  (Fig. \ref{cao2018}). SNet extracted the task relation and shared informative representations. TSNet learned the specific information corresponding to each task. The number of TSNet was consistent with the number of face attribute groups. Face attributes were split into four categories based on their locations, i.e. upper, middle, lower, and whole group.

Based on the PS-MCNN, Cao \emph{et al.} further developed a model named Partially Shared Network with Local Constraint (PS-MCNN-LC) to utilize identity information \cite{cao2018partially} because a high degree of similarity existed among the face attribute tags from the same identity. A novel loss function named LCLoss was proposed to add this constraint into the network training process and is formulated as:
\begin{equation}
LCLoss = \frac{1}{N(N-1)}\sum_{i=1}^N\sum_{j=i+1}^N
w_{i,j}\|feat^t_{si}-feat^t_{sj}\|^2_2\end{equation}
where $w_{i,j}=1$ if sample $i$ and $j$ have the same identity, otherwise $w_{i,j}=0$; $feat^t_s$ means features that are extracted from the $t^{th}$ layer of the SNet.

Besides multi-task learning, previous works also focused on improving model performance on some hard attribute prediction tasks using single-task learning. Age estimation \cite{chen2017using, pan2018mean, zhang2019c3ae, zeng2020soft} and expression recognition \cite{yang2018facial, zhang2021learning, antoniadis2021exploiting, xue2021transfer} are representative. For age estimation, traditionally, it was treated as an over-simplified linear regression task or a multi-class classification task, which ignored the ordinal information, the semantic information, and the nonlinear aging pattern. To overcome it, Chen \emph{et al.} \cite{chen2017using} proposed a ranking-CNN and treated it as multiple binary classification problems. Each binary classifier predicted whether the age of the input face was greater than a certain value $k$, where $k \in \{1,...,K\}$ and the value of $K$ depends on the ordinal labels. Then a set of binary classification results is aggregated by the following formula:
\begin{equation}
r(x_i) = 1 + \sum_{k=1}^{K-1}[f_k(x_i)>0]
\end{equation}
where $f_k(x_i)$ is the binary classification result, if $f > k$, $f=1$, otherwise, $f=-1$; $[.]$ is an operator, if the inner condition is true, it is 1, otherwise, it is 0.

\begin{figure}[htp]
    \centering
    \includegraphics[width=0.95\textwidth]{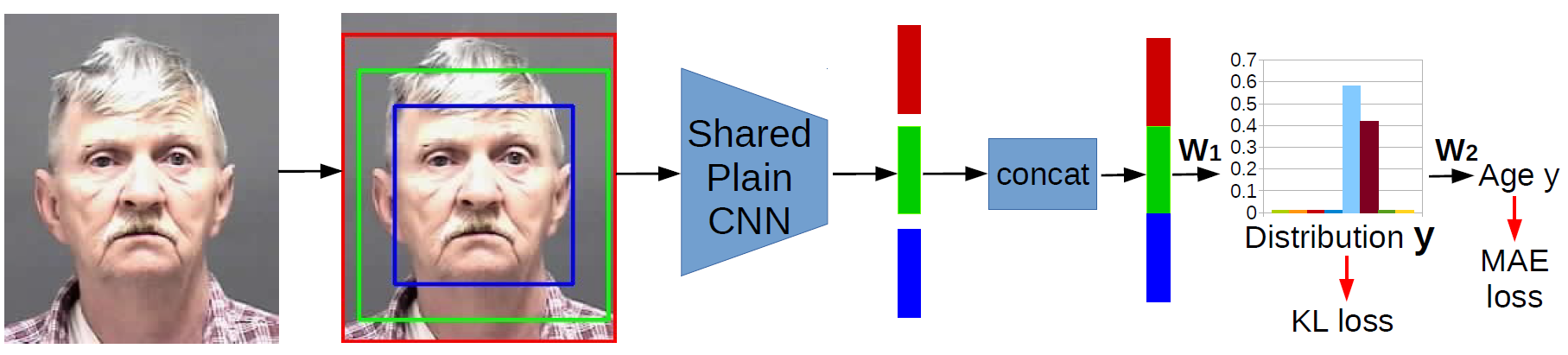}
    \caption{The pipeline of the C3AE model which was developed for age estimation \cite{zhang2019c3ae}.}
    \label{2019c3ae}
\end{figure}

Zhang \emph{et al.} \cite{zhang2019c3ae} proposed an efficient and effective age estimation model named C3AE (Fig. \ref{2019c3ae}). Instead of using a large and deep network, C3AE utilized only five convolution layers and two dense layers. The inputs of the model were three scales of cropped face images. Their results were concatenated at the last convolution layer and analyzed by the first dense layer, which output an age distribution utilizing a novel age encoding method named the two-points representation. The second dense layer output the final prediction of age. Cascade training was utilized.

For expression recognition, due to the huge variance within the emotional classes caused by different demographic characteristics, previous work focused on utilizing embedding methods to extract the expression features from images and made predictions based on them. Zhang \emph{et al.} \cite{zhang2021learning} proposed the Deviation Learning Network (DLN) to explicitly remove identity attributes from input face. An identity model and a face model were contained in the DLN and both were pre-trained Inception-Reasnet FaceNet \cite{schroff2015facenet} models. The difference was that the parameters of the identity model were fixed, but those of the face model were trainable during the training process. The outputs of the face model ($V_{face}$) and the identity model ($V_{id}$) were 512-dimensional vectors. The expression vector was given by ($V_{face} - V_{id}$), and then converted to a 16-dimensional feature space through a proposed high-order module. The final prediction was made based on the 16-dimensional features by a crowd layer \cite{rodrigues2018deep}, which was used to eliminate the annotation bias.

\subsection{Face generation}\label{subsec6.3}
The last application of FR we introduce here is face generation, especially for the ones with ID preserving.
We divide face generation methods into three types: GAN, 3D, and residual learning based face generation.
As we introduced before, algorithms \cite{tran2017disentangled,liu2018exploring,chen2019r3,huang2021age} in subsection \ref{subsubsec4.2.1} have merged face generation and recognition together. 
A lot of applications, such as face editing (age/expression changing, glasses/bread removing), face swapping use GAN to generate synthetic face images. 
3D based methods usually generate faces with different angle, namely face frontalization and rotation.
Residual learning based methods usually focus on generating faces without much content changing, such as face denoising, deblurring, super resolution.
However, many face denoising and super resolution problems adopt GAN to model the problem.

\section{Competitions and Open Source Programs}\label{sec7}

The first FR competition introduced in this section is Face Recognition Vendor Test (FRVT) \cite{phillips2003face}. 
FRVT is regularly held by the National Institute of Standards and Technology (NIST) to evaluate FR algorithms of state-of-the-art. 
It is the most authoritative and largest FR testing competition recently. 
Nearly 100 companies and research institutions have participated in this test to date. 
The FRVT does not restrict the face training set.
After participants provide the algorithm SDK, FRVT tests the performance of these algorithms directly. 
FRVT has strict restrictions on the submitted algorithms.
In specific, all submissions can only use no more than 1 second of computational resources in a single CPU thread to handle a whole FR pipeline of a single image, from face detection and alignment to feature extraction and recognition. 
FRVT is divided into four tracks, which are FRVT 1:1, FRVT 1:N, FRVT MORPH and FRVT Quality.

FRVT 1:1 evaluates algorithms with the metric of FNMR at FMR.
FNMR is the proportion of mated comparisons below a threshold set to achieve the false match rate (FMR) specified. 
FMR is the proportion of impostor comparisons at or above that threshold. 
FRVT 1:1 tests the algorithm on multiple datasets (scenes) with and without constraint environment respectively. 
The former contains visa photos, mugshot photos, mugshot photos 12+years, visaborder photos and border photos; the latter contains child photos and child exp photos.

FRVT 1:N mainly tests the identification performance and investigation performance of FR algorithms. The evaluation metrics are FNIR at FPIR, and matching accuracy.
FNIR is the proportion of mated searches failing to return the mate above threshold. 
FPIR is the proportion of non-mated searches producing one or more candidates above threshold.
Matching accuracy evaluates whether the probe image matches rank1's with a threshold of 0.

FRVT MORPH measures the performance of face forgery, whose evaluation metric is the APCER corresponding to BPCER at 0.1 and 0.01. 
APCER, or morph miss rate, is the proportion of morphs that are incorrectly classified as bona fides (nonmorphs). 
BPCER, or false detection rate, is the proportion of bona fides falsely classified as morphs. FRVT MORPH is divided into three tier, which are low quality morphs, automated morphs and high quality morphs.

FRVT Quality evaluates face quality assessment algorithms (QAAs). 
In face identification, the quality of face images in gallery is crucial to identification performance. 
As a result, the measurement of face identification is used in FRVT Quality metrics.
In detail, given a gallery set with high and low quality face images, FNMR of a FR system will be calculated first (FNMR-1).
Then the part of faces in gallery with lowest quality are discarded, and FNMR is calculated again (FNMR-2).
A smaller value of FNMR-2 indicates a better quality model performance.
Theoretically, when FNMR-1 = 0.01, after discarding the lowest quality 1\% of the images, the FNMR-2 will become 0\%.
To find the images with lowest quality in gallery, this track contains two metrics: a quality scalar and a quality vector. 
The quality scalar directly evaluates the quality of an input image by a scalar score. 
The quality vector scores multiple attributes of the input face image, such as focus, lighting, pose, sharpness, etc. 
This quality vector result can provide more precise feedback to contestants for a specific attribute which may affect image quality.

Besides FRVT from NIST, there are some famous FR competitions, such as MegaFace challenge \cite{kemelmacher2016megaface}, and MS-Celeb-1M challenge \cite{guo2016ms}.
These two competitions are no longer updating nowadays, since their goals have been met with high evaluation performance.
MegaFace competition has two challenges.
In challenge 1, contestants can use any face images to train the model. In evaluation, face verification and verification task are performed under up to 1 million distractors. 
Performance is measured using probe and gallery images from FaceScrub and FGNet.
In challenge 2, contestants need to train on a provided set with 672K identities, and then test recognition and verification performance under 1 million distractors. 
Probe and gallery images are used from FaceScrub and FGNet.
As we mention before, FaceScrub dataset is used to test FR on celebrity photos, and FGNet is to test age invariance FR.
MS1M challenge was proposed in 2016, based on real world large scale dataset on celebrities, and open evaluation system.
This challenge has provided the training datasets to recognize 1M celebrities from their face images.
The 1M celebrities are obtained from Freebase based on their occurrence frequencies (popularities) on the web, thus this dataset contains heavy noise and needs cleaning.
In evaluation, the measurement set consists of 1000 celebrities sampled from the 1M celebrities (which is not disclosed). For each celebrity, up to 20 images are manually labeled for evaluation. 
To obtain high recognition recall and precision rates, the contestants should develop a recognizer to cover as many as possible celebrities.

\section{Conclusion}\label{sec9}

In this paper, we introduce about 100 algorithms in face recognition (FR), including every sides of FR, such as its history, pipeline, algorithms, training and evaluation datasets and related applications.

\bibliographystyle{ieeetr}
\bibliography{bib.bib}

\end{document}